%% file: arxiv.tex
\documentclass[11pt, a4paper, logo, copyright, nonumbering]{mll_style}

\input{math_commands.tex}

\PassOptionsToPackage{numbers,square,sort&compress}{natbib}
\usepackage{natbib}
\usepackage[most]{tcolorbox}
\usepackage{nicefrac}
\usepackage{multirow}
\usepackage{subcaption}
\usepackage{wrapfig}
\usepackage{placeins}

\usepackage{enumitem}
\usepackage{algorithm}
\usepackage{algpseudocode}
\usepackage{booktabs}
\usepackage{tabularx}
\usepackage{makecell}
\usepackage[capitalise,noabbrev]{cleveref}

\usepackage[toc,page,header]{appendix}
\hypersetup{linktoc=all}

\makeatletter
\newif\ifefeappendixentries
\newcommand{\efeappendixcontents}{%
  \begingroup
  \small
  \setlength{\parskip}{0pt}
  \hypersetup{linkcolor=citecolor}
  \renewcommand{\contentsname}{\textcolor{citecolor}{Contents}}%
  \efeappendixentriesfalse
  \def\efefirstappendix{appendix.A}%
  \let\efecontentsline\contentsline
  \renewcommand*\contentsline[4]{%
    \def\efeentryanchor{##4}%
    \ifx\efeentryanchor\efefirstappendix
      \efeappendixentriestrue
    \fi
    \ifefeappendixentries
      \efecontentsline{##1}{##2}{##3}{page.##3}%
    \fi}%
  \let\l@part\@gobbletwo
  \let\l@xpart\@gobbletwo
  \renewcommand*\l@section[2]{%
    \addpenalty{-\@highpenalty}%
    \vskip 1.5em
    \@dottedtocline{1}{0pt}{1.8em}{\bfseries ##1}{\bfseries ##2}}%
  \let\efesubsectionentry\l@subsection
  \renewcommand*\l@subsection[2]{%
    \vskip 0.4em
    \efesubsectionentry{##1}{##2}}%
  \tableofcontents
  \endgroup
}
\makeatother

\newcommand{\efe}{Embodied First Exam}
\newcommand{\efeshort}{EFE}
\newcommand{\evalsetting}[1]{\texttt{#1}}
\usepackage{fontawesome5}
\usepackage{simpleicons}
\definecolor{findingaccent}{HTML}{5E77D6}
\definecolor{findingback}{HTML}{F1F5FE}
\definecolor{findingframe}{HTML}{CFD9F3}
\newtcolorbox{researchquestion}{enhanced,
  width=0.94\linewidth, center,
  colback=abstractbgcolor, colframe=abstractbgcolor, boxrule=0pt, arc=5pt,
  left=9pt, right=9pt, top=7pt, bottom=7pt,
  before skip=9pt, after skip=10pt, fontupper=\normalsize}
\newtcolorbox{findingbox}[1]{enhanced, colback=findingback, colframe=findingframe, boxrule=0.4pt, arc=5pt,
  sharp corners=west, borderline west={3.5pt}{0pt}{findingaccent},
  title={\faLightbulb\enspace Key Takeaways: #1}, fonttitle=\bfseries\small, coltitle=white,
  attach boxed title to top left={xshift=10pt, yshift*=-\tcboxedtitleheight/2},
  boxed title style={colback=findingaccent, colframe=findingaccent, boxrule=0pt, arc=3pt, left=5pt, right=6pt, top=2pt, bottom=2pt},
  left=8pt, right=8pt, top=5pt, bottom=5pt, before skip=10pt, after skip=10pt, fontupper=\small,
  before upper={\setlist[itemize]{leftmargin=1.1em, topsep=0pt, partopsep=0pt, itemsep=2pt, parsep=0pt}}}

\providecommand{\efelayout}{3}
\ifcase\efelayout\or
  \newcommand{\efetitleshift}{-10pt}\newcommand{\efeauthorshift}{-10pt}\newcommand{\efeauthorend}{-20pt}\newcommand{\efeaffilfont}{\small}\newcommand{\efeauthorfont}{}\newcommand{\efeauthorbalance}{}\newcommand{\efeauthorunbalance}{}
  \newcommand{\efelinks}{\\{\small\normalfont\href{https://www.embodied-first-exam.ai}{\nolinkurl{https://www.embodied-first-exam.ai}}} }
\or
  \newcommand{\efetitleshift}{-22pt}\newcommand{\efeauthorshift}{-18pt}\newcommand{\efeauthorend}{-20pt}\newcommand{\efeaffilfont}{\small}\newcommand{\efeauthorfont}{}\newcommand{\efeauthorbalance}{}\newcommand{\efeauthorunbalance}{}
  \newcommand{\efelinks}{\\{\small\normalfont
    \href{https://www.embodied-first-exam.ai}{\faGlobe~embodied-first-exam.ai}\quad
    \href{https://github.com/mll-lab-nu/embodied-first-exam}{\faGithub~github.com/mll-lab-nu/embodied-first-exam}\quad
    \href{https://huggingface.co/datasets/MLL-Lab/embodied-exam-data}{\simpleicon{huggingface}~huggingface.co/datasets/MLL-Lab/embodied-exam-data}} }
\or
  \newcommand{\efetitleshift}{0pt}\newcommand{\efeauthorshift}{-10pt}\newcommand{\efeauthorend}{0pt}\newcommand{\efeaffilfont}{\footnotesize}\newcommand{\efeauthorfont}{\fontsize{9}{12}\selectfont}%
  \newcommand{\efeauthorbalance}{\leftskip=0pt plus 4em\rightskip=0pt plus 4em\parfillskip=0pt\relax}%
  \newcommand{\efeauthorunbalance}{\leftskip=0pt plus 1fil\rightskip=0pt plus 1fil\parfillskip=0pt\relax}
  \newcommand{\efelinks}{}
\fi

\title{\centering
\vspace{\efetitleshift}
Embodied First Exam
}

\date{}

\author[*]{
\vspace{\efeauthorshift}
\efeauthorbalance\efeauthorfont{\mbox{Kangrui Wang$^{1*\S}$}, \mbox{Qineng Wang$^{1*\S}$}, \mbox{Changhe Chen$^{1*\S}$}, \mbox{Pingyue Zhang$^{1*\S}$}, \mbox{Chi Gui$^{1*\S}$}, \mbox{Feishi Wang$^{1*\S}$}, \mbox{Canyu Chen$^{1*\S}$}, \mbox{Wenli Xiao$^{2\S}$}, \mbox{Xiaoyu Huang$^{3\S}$}, \mbox{Tonghe Zhang$^{2\S}$}, \mbox{Linjie Li$^{4\S}$}, \mbox{Zihan Wang$^{1}$}, \mbox{Minyeong Kim$^{5}$}, \mbox{Yiyou Sun$^{3}$}, \mbox{Mingkai Deng$^{2}$}, \mbox{Jinyu Hou$^{2}$}, \mbox{Qi Zhu$^{1}$}, \mbox{Han Liu$^{1}$}, \mbox{Eric P. Xing$^{2}$}, \mbox{Huaxiu Yao$^{6}$}, \mbox{Zhiwen Fan$^{7}$}, \mbox{Yuejiang Liu$^{5}$}, \mbox{Dawn Song$^{3}$}, \mbox{Pieter Abbeel$^{3}$}, \mbox{Li Fei-Fei$^{5}$}, \mbox{Jiajun Wu$^{5}$}, \mbox{Yuke Zhu$^{8}$}, \mbox{Guanya Shi$^{2\dagger}$}, \mbox{Ruohan Zhang$^{1\dagger}$}, \mbox{Manling Li$^{1\dagger}$}}
\\
\efeauthorunbalance\efeaffilfont $^{1}$Northwestern University~~~$^{2}$Carnegie Mellon University~~~$^{3}$UC Berkeley~~~$^{4}$University of Washington\\
\efeaffilfont $^{5}$Stanford University~~~$^{6}$UNC-Chapel Hill~~~$^{7}$Texas A\&M University~~~$^{8}$UT Austin\\
\efeaffilfont $^*$Equal contribution~~~$^\S$Core contributor~~~$^\dagger$Equal advising\efelinks
\vspace{\efeauthorend}
}

\makeatletter

\ifcase\efelayout\or
  \newcommand{\efeintroteaser}{%
    \begin{figure}[t]
      \centering
      \includegraphics[width=\linewidth]{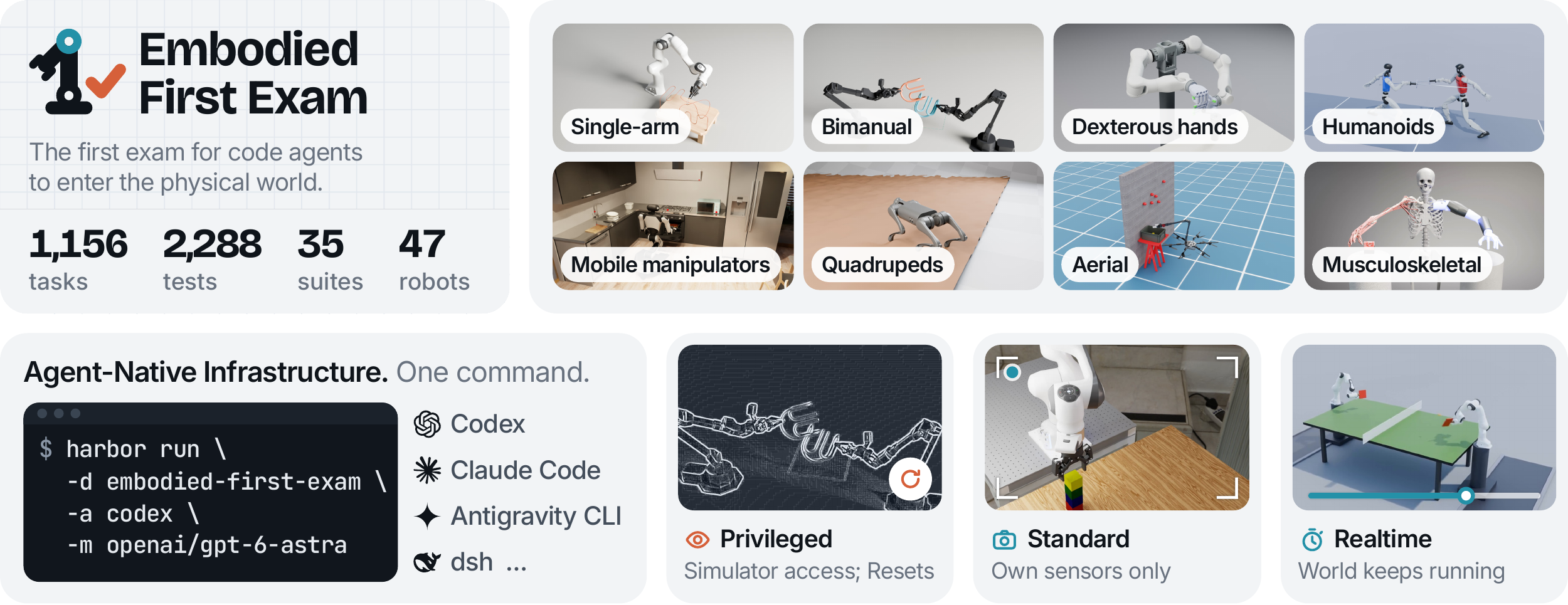}
      \caption{\efeteasercaption}
      \label{fig:teaser}
    \end{figure}}
  \AtBeginDocument{\g@addto@macro\theabstract{\par\vspace{6pt}%
    {\normalfont\small
    \begin{tabular}{@{}ll@{}}
    \textbf{Code:} & \href{https://github.com/mll-lab-nu/embodied-first-exam}{\nolinkurl{https://github.com/mll-lab-nu/embodied-first-exam}} \\
    \textbf{Trajectory Data:} & \href{https://huggingface.co/datasets/MLL-Lab/embodied-exam-data}{\nolinkurl{https://huggingface.co/datasets/MLL-Lab/embodied-exam-data}} \\
    \end{tabular}}}}
\or
  \newcommand{\efeintroteaser}{}
  \newcommand{\efeteaser}{%
    \vspace*{-14pt}\noindent\begin{minipage}{\textwidth}
      \centering
      \includegraphics[width=0.8\linewidth]{figures/teaser.pdf}
      \captionof{figure}{Overview of the \efe{}, a comprehensive evaluation framework for coding agents on robotics tasks. It is large and diverse, spanning 1,156 tasks, 35 suites and 47 robots, and easy to run and evaluate: one command runs any coding agent, and each task can be graded in up to three evaluation settings.}
      \label{fig:teaser}
    \end{minipage}\par}
  \newcommand{\efe@abscontent}{%
    \noindent
    \begin{tcolorbox}[colframe=abstractbgcolor, colback=abstractbgcolor, boxrule=0pt, arc=2mm, enhanced,
        top=8pt, bottom=8pt, left=15pt, right=15pt, width=\textwidth]
      {\linespread{1.1}\fontsize{11}{12}\selectfont \theabstract}
    \end{tcolorbox}}
  \renewcommand{\abscontent}{\efeteaser\efe@abscontent}
\or
  \newcommand{\efeintroteaser}{}
  \fancypagestyle{efefirststyle}{%
    \fancyhead{}%
    \fancyfoot{}%
  }
  \renewcommand{\maketitle}{%
    \thispagestyle{efefirststyle}%
    \enlargethispage{44pt}%
    \vspace*{-42pt}%
    \noindent
    \begin{tcolorbox}[enhanced, colback=abstractbgcolor, colframe=abstractbgcolor, boxrule=0pt, arc=2mm,
        left=6pt, right=6pt, top=10pt, bottom=8pt, width=\textwidth,
        segmentation style={draw=black!30, dashed, line width=0.6pt}]
      {\centering\titlefont
        \raisebox{-0.25\height}{\includegraphics[height=22pt]{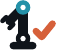}}\hspace{6pt}%
        \@title\par}
      \vskip 12pt
      {\centering \@author\par}
      \tcbline
      {\centering
        \includegraphics[width=\linewidth]{figures/teaser.pdf}\par
        \captionsetup{font=small}%
        \captionof{figure}{Overview of the \efe{}, a comprehensive evaluation framework for coding agents on robotics tasks. It is large and diverse, spanning 1,156 tasks, 35 suites and 47 robots, and easy to run and evaluate: one command runs any coding agent, and each task can be graded in up to three evaluation settings.}\label{fig:teaser}\par}
      \tcbline
      {\footnotesize\setlength{\parskip}{2pt}
        \textbf{Website:} \href{https://www.embodied-first-exam.ai}{\nolinkurl{embodied-first-exam.ai}}\par
        \textbf{Code:} \href{https://github.com/mll-lab-nu/embodied-first-exam}{\nolinkurl{github.com/mll-lab-nu/embodied-first-exam}}\par
        \textbf{Trajectory Data:} \href{https://huggingface.co/datasets/MLL-Lab/embodied-exam-data}{\nolinkurl{huggingface.co/datasets/MLL-Lab/embodied-exam-data}}\par}
    \end{tcolorbox}
    \vskip 4pt
    {\centering\bfseries Abstract\par}
    {\normalsize\noindent \theabstract\par}
  }
\fi
\makeatother

\input{sections/0_abstract/abstract}

\makeatletter
\providecommand\@LN[2]{}
\@ifundefined{ifmtcsecondpart}{\newif\ifmtcsecondpart}{}
\makeatother

\begin{document}

\maketitle
\clearpage

\input{sections/1_introduction/introduction}

\input{sections/2_benchmark/benchmark}

\input{sections/3_experiments/experiments}

\input{sections/4_findings/findings}

\input{sections/5_related_work/related_work}

\input{sections/6_conclusion/conclusion}

\subsubsection*{Acknowledgements}
We acknowledge and disclose the use of AI tools in coding and paper writing. Pieter Abbeel holds concurrent appointments as a Professor at UC Berkeley and as an Amazon Scholar. This paper describes work performed at UC Berkeley and is not associated with Amazon.

\input{appendix_body}

\clearpage
\pdfbookmark[0]{References}{efe-references}
{
\small
\bibliographystyle{plainnat}
\bibliography{references}
}

\end{document}

%% file: math_commands.tex
\usepackage{amsmath,amsfonts,bm}

\def\eqref#1{equation~\ref{#1}}

\def\1{\bm{1}}

\DeclareMathAlphabet{\mathsfit}{\encodingdefault}{\sfdefault}{m}{sl}
\SetMathAlphabet{\mathsfit}{bold}{\encodingdefault}{\sfdefault}{bx}{n}

%% file: sections/0_abstract/abstract.tex
\begin{abstract}
Early successes on agentic robotics point to a potential paradigm shift toward using coding agents to control robots. 
Existing robotics benchmarks are built around visuomotor policy learning, leaving it open whether coding agents can reliably control diverse robots under physical-world constraints. 
We therefore introduce the \textbf{\efe{}}, a unified, comprehensive, and agent-native evaluation framework where coding agents can choose their own control approach: direct control, code to control, or train a policy to control. We hope it  serves as a first exam for coding agents before they enter the physical world, with 1,156 tasks in 35 suites on 47 robots, including 9 new suites that go beyond pick-and-place to spatial puzzles, dexterous manipulation, multi-robot interaction, dynamic whole-body control, and sim-to-sim engineering. 
Each task can be run in up to three evaluation settings. The \evalsetting{Standard} Setting mimics the real world: the agent receives only sensor observations and, by default, cannot reset the environment. Its \evalsetting{Realtime} variant keeps the environment running while the agent reasons. The \evalsetting{Privileged} Setting gives the agent access to the full simulator state and unlimited resets, allowing it to search for an oracle solution for training purposes. Success rates in this setting serve as an approximate upper bound. On the 456 core tasks, the strongest of eight agents, GPT-6 Astra in Codex, achieves a success rate of 34\% in the \evalsetting{Standard} Setting, leaving much room for improvement. To see where agents fail, we label the stage at which failed attempts go wrong and compare failures across agents, settings, and stage categories.

\end{abstract}

%% file: sections/1_introduction/introduction.tex
\section{Introduction}
\label{sec:intro}

\efeintroteaser

Robots acquire new skills either by manual programming or by learning from data, as in recent vision-language-action models~\citep{brohan2023rt2,kim2024openvla,black20240,intelligence2025pi05,nvidia2025gr00t,team2025gemini} and world action models~\citep{ye2026world,wang2026world,shen2026world}. Coding agents powered by large language models suggest a third way. They already complete challenging software engineering tasks in the digital world~\citep{jimenez2023swebench,merrill2026terminalbench,miserendino2025swelancer,kwa2025measuring}. What makes these agents promising for robotic engineering is their capabilities to turn instructions into executable code and revise their plans through automated experimentation~\citep{liang2022code,singh2023progprompt,vemprala2023chatgpt}. Recent early success in agentic robots suggests such a potential paradigm shift, where agents directly control robots~\citep{hu2026via,zhang2026unexpected,zhou2026gpt6,sensenova2026astra}, write robot programs~\citep{jia2026agent,fu2026capx,merler2026coding,xie2026pusht,tan2026forca,hu2026simex}, or train policies~\citep{xiao2026enpire,li2026harbor}. This motivates a systematic assessment of the capability boundary of coding agents:
\begin{researchquestion}
\textbf{What robotic tasks can general-purpose coding agents solve, under what physical-world constraints, and where do they fail?}
\end{researchquestion}

Existing benchmarks do not yet fully address this question. The capability boundary cannot be established from a focused set of robotics tasks ~\citep{james2020rlbench,yu2020metaworld,liu2023libero,li2024behavior1k,nasiriany2024robocasa,chen2026robodojo}, nor from automated programming tasks without imposing physical-world constraints~\citep{jimenez2023swebench,merrill2026terminalbench,zhou2023webarena,xie2024osworld,chan2024mlebench,ouyang2025kernelbench}.
This requires defining which robotic capabilities to evaluate, measuring how physical-world constraints affect success, and supporting different models and harnesses through task interfaces.

To answer this question, we introduce the \textbf{\efe{} (\efeshort{})}, a unified evaluation framework that connects coding agents to simulated robots (\Cref{fig:teaser}), following three principles.

\textbf{First, include diverse tasks that go beyond pick-and-place.} 
Spatial puzzles test reasoning about spatial constraints, while sim-to-sim engineering tests how agents revise controllers from execution feedback under unfamiliar dynamics. Dynamic tasks test responses to a changing world. Multi-robot interaction and whole-body control test coordination. 
\efeshort{} therefore contains 1,156 tasks in 35 suites on 47 robots. It combines 26 suites adapted from existing robotics benchmarks with 9 new suites covering these capabilities and safety.

\textbf{Second, measure how success depends on physical-world constraints.} Its tasks should follow the constraints of a real robot as closely as possible. Success with full simulator state and resets does not establish what an agent can achieve through partial observations, with no resets, or while the dynamic world keeps running. We therefore define up to three evaluation settings per task. \evalsetting{Standard} Setting follows real-robot conditions: the agent observes the environment only through the robot's sensors and cannot reset it by default. The simulation waits while the agent reasons, and its \evalsetting{Realtime} variant keeps the environment running while the agent makes decisions. \evalsetting{Privileged} Setting provides an approximate upper bound and can be used to generate training data, where the agent has access to the full simulator state and unlimited resets, allowing it to search for an oracle solution by trial and error.  We call each <task--setting> pair a \emph{test}. Tests of the same task share the initial scene and success rule, allowing us to compare success as state access, reset limits, and timing change.

\begin{figure}[t]
  \centering
  \includegraphics[width=\linewidth]{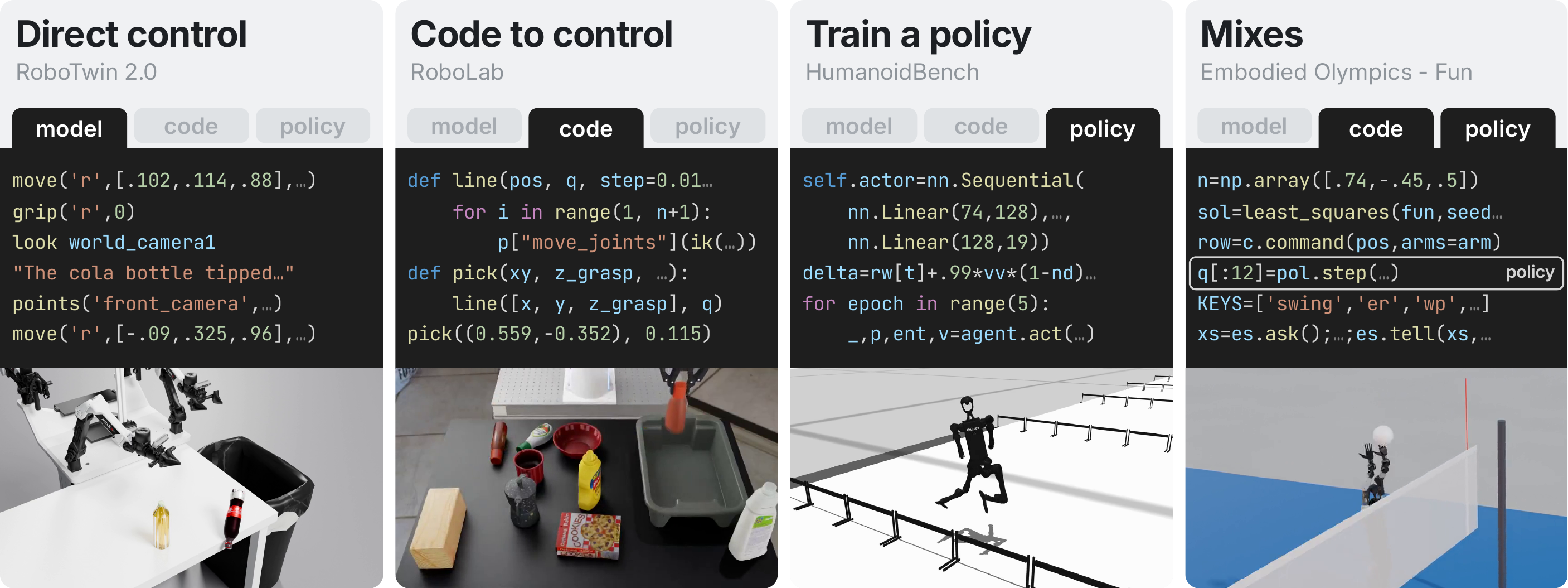}
  \caption{\looseness=-1 Examples of the control approaches, each with an excerpt of the agent's commands or code. Dark tabs mark the approaches used (\emph{model} is direct control), and the framed line is a policy.}
  \label{fig:approaches}
  \vspace{-6pt}
\end{figure}

\textbf{Third, provide unified infrastructure for different models, harnesses, and control approaches.} 
With a single command, \efeshort{} runs different combinations of models and harnesses across diverse robotics tasks. 
Given a natural-language task instruction, the agent chooses its control approach (\Cref{fig:approaches}). In \emph{Direct control}, it directly outputs actions after receiving observations~\citep{hu2026via,zhang2026unexpected,zhou2026gpt6}. In \emph{Code to control}, it outputs a planner, controller, or other program that generates the robot's actions~\citep{liang2022code}. In \emph{Train a policy}, it trains a policy in simulation and uses it to control the robot~\citep{ma2024eureka,li2026harbor}. The framework enforces the access permitted by each evaluation setting, and a verifier outside the agent's control grades every attempt~\citep{zhu2025establishing,zhong2025impossiblebench}.

We evaluate eight coding agents on a core set of 456 hard tasks. The strongest agent, GPT-6 Astra~\citep{openai2026gpt6astra} in Codex, solves 77\% of the \evalsetting{Privileged} tests but only 30\% of the \evalsetting{Standard} tests of the same 391 tasks. Apart from the three strongest agents, GPT-6 Astra, Claude Opus 5.5 and Gemini 4 Argon, the best result comes from Kimi 3.1 Pro, which solves 9\% of the \evalsetting{Standard} tests of the same 391 tasks (34 tests, \Cref{fig:transfer}) and 7\% of the \evalsetting{Realtime} tests (\Cref{fig:leaderboard}). The \evalsetting{Realtime} Setting is harder still: 63\% of its tests remain unsolved by any agent, compared with 50\% of the \evalsetting{Standard} tests of the same 54 tasks. To identify where failures occur within a task, we divide each task into stages. Each stage corresponds to a change in the world required to complete the task. For example, the task of putting a tomato into a closed fridge has three stages: the fridge is open, the tomato is inside, and the fridge is closed again. For each failed attempt of the eight agents on 427 tasks, a separate coding agent labels the stage at which the failure occurred. Our main findings are as follows.
\begin{itemize}[leftmargin=1.5em, itemsep=2pt, topsep=4pt]
  \item \textbf{\evalsetting{Privileged} versus \evalsetting{Standard} (\Cref{sec:f-settings}).} Resets help GPT-6 Astra more than the exact object state does, and with both, its success rate closes most of the gap to the \evalsetting{Privileged} Setting. Stages that need precision become much harder in the \evalsetting{Standard} Setting. At these stages, resets help, but the exact state alone does not.
  \item \looseness=-1 \textbf{Where agents fail (\Cref{sec:f-difficulty}).} Failures on a task tend to happen at the same stage, usually the first stage of the task. This stage also tends to be the same in the \evalsetting{Privileged} and \evalsetting{Standard} settings. Stages that change a material or move the robot's body have the highest failure rates.
  \item \textbf{Differences between agents (\Cref{sec:f-agents}).} Weaker agents are much more likely than stronger agents to fail before making the required contact. In the \evalsetting{Standard} Setting, the three strongest agents solve partly different sets of tests. Switching harnesses can change a model's success rate by up to 12 percentage points.
  \item \textbf{How agents control the robot (\Cref{sec:f-control}).} In the \evalsetting{Standard} and \evalsetting{Realtime} settings of a soccer task, agents trained a policy when given a GPU and a training library, and only trained policies won. Agents mostly control the robot directly in the \evalsetting{Standard} Setting and write programs to control it in the \evalsetting{Privileged} and \evalsetting{Realtime} settings.
\end{itemize}

%% file: sections/2_benchmark/benchmark.tex
\section{The Embodied First Exam: A Unified Evaluation Framework for Coding Agents in Robotics}\label{sec:benchmark}

In this section, we describe how we evaluate coding agents on robotics tasks (\Cref{sec:test}), the resource constraints of each task (\Cref{sec:limits}), and the tasks of the exam (\Cref{sec:tasks}).

\begin{figure}[t]
  \centering
  \setlength{\parskip}{0pt}
  \captionsetup[subfigure]{font={footnotesize,sf,bf},justification=raggedright,singlelinecheck=false,position=top,skip=2pt}
  \begin{subfigure}[t]{\linewidth}
    \caption{Unified Evaluation Framework for Coding Agents in Robotics}\label{fig:framework}
    \includegraphics[width=\linewidth]{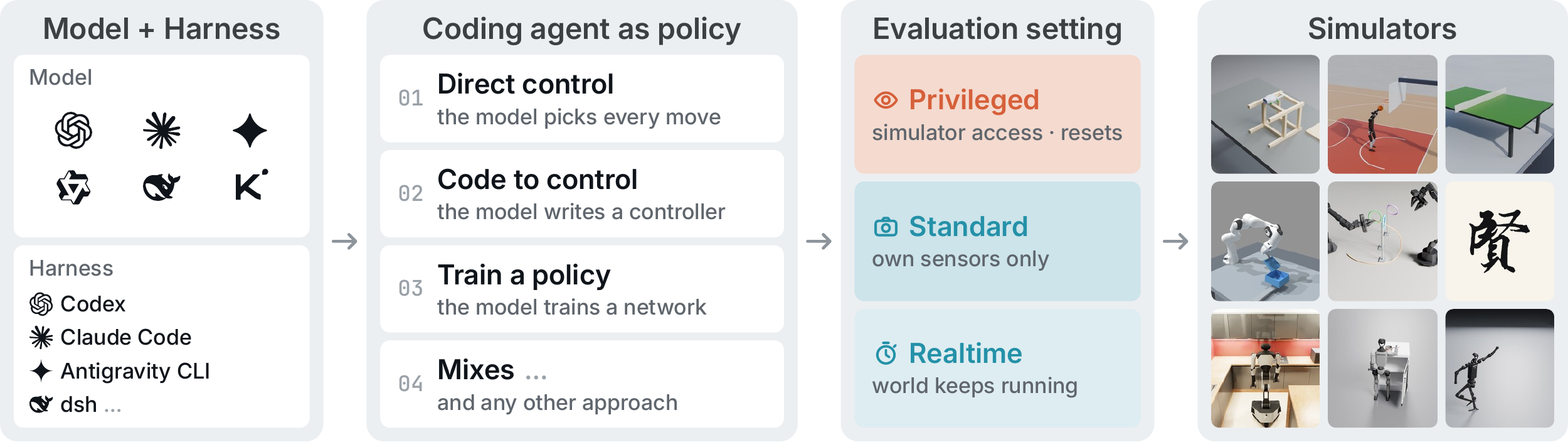}
  \end{subfigure}
  \par\nointerlineskip\vspace{3pt}
  \begin{subfigure}[t]{\linewidth}
    \caption{Evaluation Protocol: Privileged, Standard, and Realtime Settings}\label{fig:settings}
    \includegraphics[width=\linewidth,trim=0 0 0 31.5bp,clip]{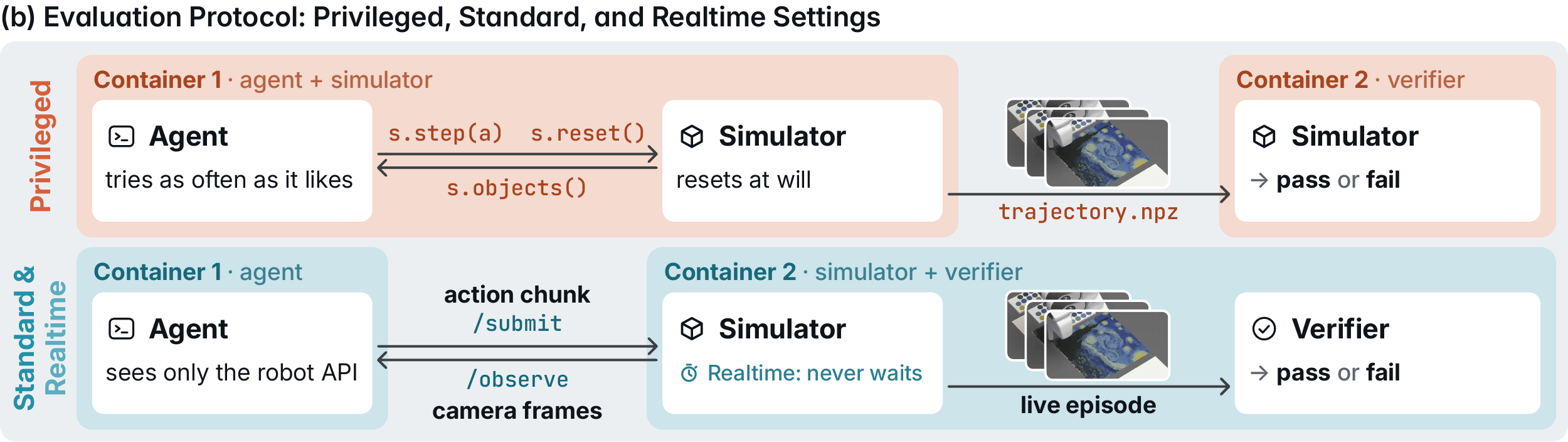}
  \end{subfigure}
  \caption{\looseness=-1 The \efeshort{} framework and evaluation protocol. (a) Our unified framework supports different models, harnesses, control approaches, and simulators. (b) In \evalsetting{Privileged}, the agent and simulator share a container, and the verifier replays the submitted answer separately. In \evalsetting{Standard} and \evalsetting{Realtime}, the simulator and verifier are separated from the agent and grade the live episode.}
  \label{fig:framework-settings}
\end{figure}

\subsection{Framework and Evaluation Protocol}\label{sec:test}\label{sec:settings}\label{sec:honest}

Existing robotics benchmarks focus on visuomotor policy learning with a relatively fixed input-output interface, whereas we evaluate coding agents that can choose their own control approach. Our design principle is therefore to give the agent broad freedom over how it acts and to constrain only the result: we fix where it starts, what resources it gets, and how success is checked. To achieve this, we package every robotics task in the format of Harbor~\citep{harbor2026}, the evaluation framework of Terminal-Bench~\citep{merrill2026terminalbench}, which runs the task in containers, as shown in \Cref{fig:framework}. The agent has its own workspace, where it can install packages, write and run programs, and train models. Each task starts from a fixed initial state; a \emph{verifier} outside the agent's container reports whether the agent's answer meets the task's success rule.

\looseness=-1 On top of this framework, we define a unified evaluation protocol with up to three settings per task (\Cref{fig:settings}). We design \evalsetting{Privileged} as an approximate upper bound on performance, with full access to the simulator for trial and error, while \evalsetting{Standard} and \evalsetting{Realtime} make the evaluation as close to a real robot as possible. The bound holds over a set of tasks, not for every single task, because an agent can solve a task in \evalsetting{Standard} that it fails in \evalsetting{Privileged} (\Cref{app:results}). For a given task, all settings share the same initial scene, success rule, and time limit, so their results are paired. They differ in what the agent observes, whether it can reset, whether the world waits, and where the simulator and the verifier run, never in the control approach the agent may choose. We ablate the first two differences in \Cref{sec:f-settings}.

\noindent
\textbf{Privileged.} In \evalsetting{Privileged}, the agent runs in the same container as the simulator, so it can read the full simulation state and reset without limit. The agent must submit a low-level robot trajectory (or a controller program for some tasks). The verifier replays the trajectory (or runs the controller) in a separate container to determine task success.

\noindent
\textbf{Standard.} In \evalsetting{Standard}, we run the simulator as a service in a separate container. Through predefined APIs, the agent has limited access to the simulator: it can query proprioceptive state (e.g., joint positions, velocities, and torques), and sensor data (e.g., calibrated RGB-D images), and send action chunks to control the robot. The simulation is paused between API requests, so the agent faces no real-time pressure. Resets are disallowed unless the task provides an explicit reset budget. The verifier grades task progress online as the agent controls the robot, and in most suites the interface reports a success to the agent and ends the episode.

\noindent
\textbf{Realtime.} \evalsetting{Realtime} is the same as \evalsetting{Standard}, except that the simulation does not pause between agent actions but keeps running in real time; for computationally expensive tasks, it runs at a fixed fraction of real time (e.g., 0.5$\times$). Due to compute constraints, we currently evaluate \evalsetting{Realtime} only on dynamic tasks. \Cref{app:framework} details each setting.

\subsection{Resource Constraints for Coding Agents}\label{sec:limits}

\begin{figure}[t]
  \centering
  \includegraphics[width=\linewidth]{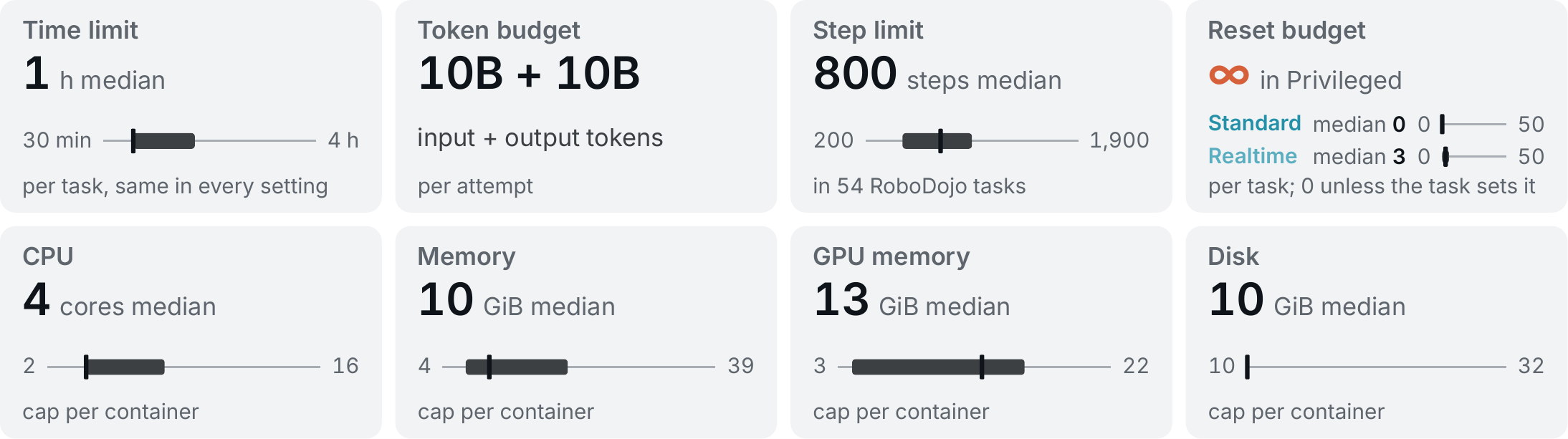}
  \caption{Resource constraints across the exam. The large number and the tick give the median over tests (for GPU memory, over the tests that reserve a GPU), the line spans the range, and the dark box spans the middle half of the tests.}
  \label{fig:limits}
\end{figure}

As shown in \Cref{fig:limits}, we set resource constraints for each task: a time limit and a token budget for the agent, a reset budget and, in some tasks, a step limit for the robot (RoboDojo's own limit in \Cref{fig:limits}, or a cap on the episode length that the instruction states), and caps on CPU, memory, GPU memory, and disk. The token budget is high enough that no attempt reaches it. The time limit is the same in every setting of a task, and when it runs out we stop the agent and grade the result it has at that point. Resets are unlimited in \evalsetting{Privileged}. In \evalsetting{Standard} and \evalsetting{Realtime}, there are no resets unless the task grants a budget, which nine suites do, with 3 to 50 resets. We set the compute caps per task from its measured needs and monitor usage throughout the run (\Cref{app:framework}).

We observe that coding agents may choose different control approaches under different resource constraints and in different settings (\Cref{fig:approaches,sec:f-control}). For example, if we provide a GPU, agents can train a policy to control the robot (\Cref{fig:approach-case}).

\subsection{Task Suites}\label{sec:tasks}

The exam contains 1,156 tasks in 35 suites. A suite is a set of related tasks on one simulator. Each task states a goal, usually in natural language, such as untying a knot or turning the page of a book with a dexterous hand, and the agent must control the robot to reach that goal. We built 9 suites with 411 tasks for the exam and adapted the other 26 suites with 745 tasks from existing robotics benchmarks. Together, the tasks use 47 robots of nine types
(\Cref{tab:suites} in \Cref{app:tasks}, which lists the simulator, the robots and the number of tasks of
every suite).

\begin{figure}[t]
  \centering
  \includegraphics[width=\linewidth]{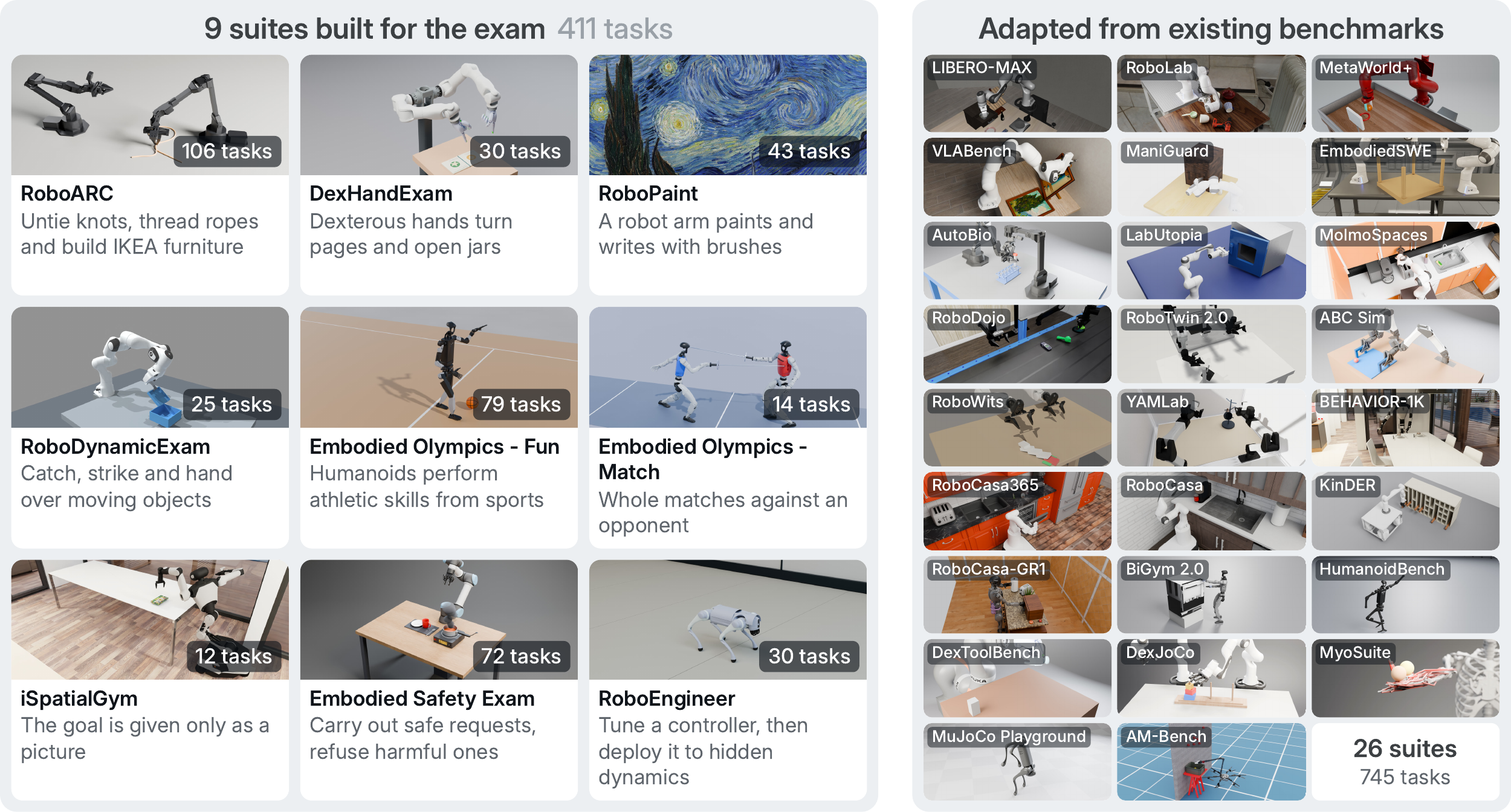}
  \caption{The tasks of the exam: the 9 suites built for the exam, each with a frame and its number of tasks,
  in the three groups of \Cref{sec:tasks}, and a frame from each of the 26 adapted suites, ordered by their
  main robot embodiment, from single arms to aerial robots.}
  \label{fig:tasks}
\end{figure}

\noindent
\textbf{New suites.} We built nine new suites because the adapted suites contain few tasks for several embodiments
and capabilities. Most of their original benchmarks focus on visuomotor policy learning (\Cref{sec:related}), and
618 of their 745 tasks use single-arm, bimanual or mobile manipulators. The new suites form three groups, which are the rows of
\Cref{fig:tasks}, and \Cref{app:tasks-new} describes each suite. RoboARC, DexHandExam and RoboPaint require
contact-rich manipulation of deformable objects, multi-part assemblies and tools. RoboDynamicExam and the two Embodied
Olympics suites involve moving objects and opponents, so the agent must act at the right time, and they contain most of
the humanoid tasks of the exam. iSpatialGym, Embodied Safety Exam and RoboEngineer test abilities beyond executing a
stated goal, since they give the goal only as an image, mix hazardous with benign requests, or require adapting a
controller to hidden dynamics.

\noindent
\textbf{Adapted suites.} The other 26 suites come from existing benchmarks, and their tasks range from
language-conditioned tabletop manipulation and long-horizon household activities to humanoid whole-body control,
dexterous tool use, musculoskeletal control, aerial manipulation and laboratory automation (\Cref{app:tasks-adapted}).
For each suite, we keep the scenes, robots and success rules of the original benchmark, in some suites with a stricter
success rule. We save the initial state of each task as a frozen scene, write an instruction that states every
convention that the agent cannot infer and connect the suite to the interfaces of the three settings. We also remove
the expert policies and demonstration data of the original benchmark, so that the agent cannot copy a known solution from its container.

\begin{figure}[t]
  \centering
  \includegraphics[width=\linewidth]{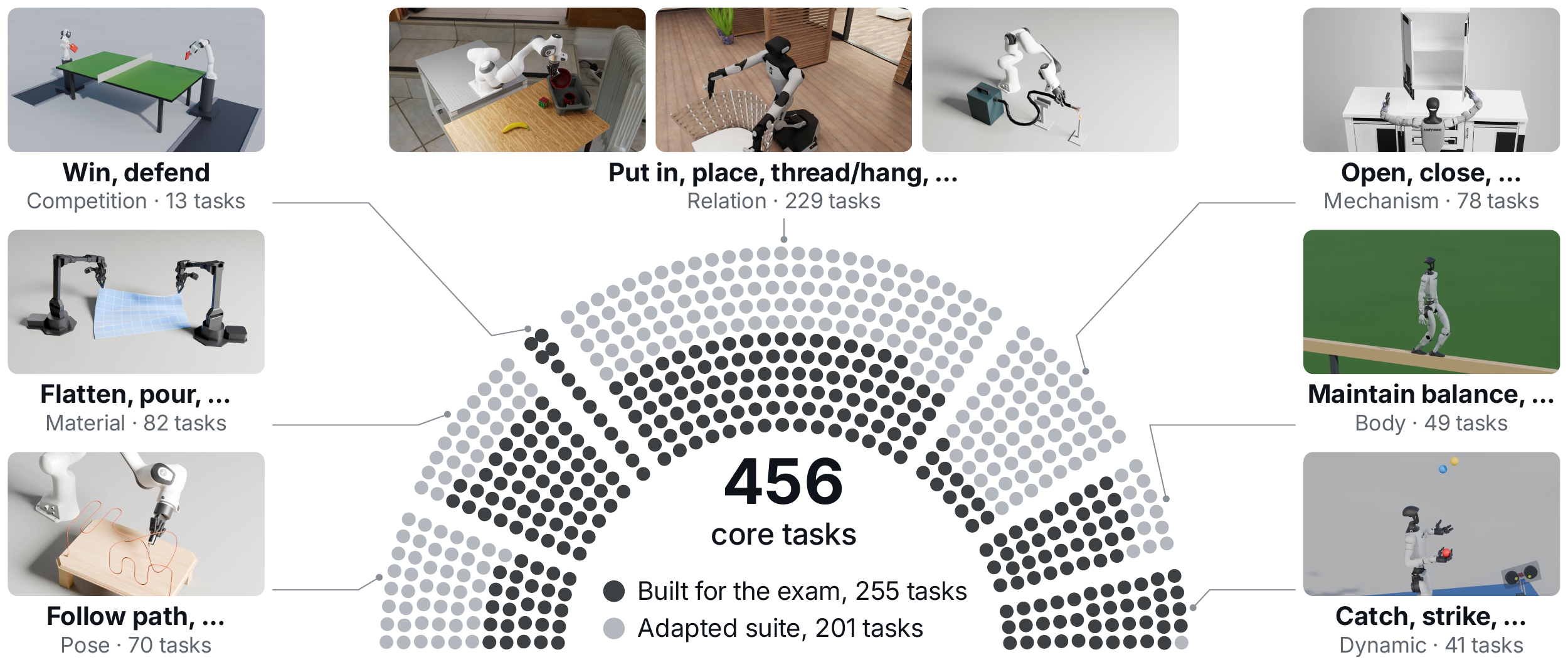}
  \caption{The core tasks by the stage categories of their task graphs, with an example task of each category. Each seat is one task in one category, and the angle of a category is proportional to its number of tasks. A task counts once in each category its task graph contains. The centre counts all 456 core tasks, and the seats cover the 427 with a task graph; no task graph has a Safety or Information stage, since the graphs of Embodied Safety Exam give its safety rules as conditions of their stages (\Cref{app:fa-procedure}).}
  \label{fig:task-dist}
\end{figure}

\noindent
\textbf{Core set.} We report results on a core set of 456 hard tasks (901 tests), because tasks within a
suite often require similar skills and current agents already solve many of the tasks. The core set excludes tasks
that we judged easy and tasks that an initial screening agent, GPT-6 Luna, solved, and it keeps every task that the
authors of a suite asked to keep. Because GPT-6 Luna belongs to the same model family as GPT-6 Astra, the screening
may bias the core set against tasks that this family solves. \Cref{tab:suites} gives the number of core tasks in each
suite. Each core task also has a default setting, and the controlled runs of \Cref{sec:f-settings} use the default test of each core task. The default is
\evalsetting{Standard} for 403 tasks and \evalsetting{Realtime} for 53 tasks in which timing is essential; one more core task also has a \evalsetting{Realtime} test.
The remaining 700 tasks stay in the exam and can be run in the same way.
\looseness=-1 We split 427 of the core tasks into stages, such as opening a fridge (\Cref{sec:labels}), and \Cref{fig:task-dist} groups the core tasks by the categories of these stages (\Cref{tab:fa-steps}). Relation stages, such as placing or assembling objects, are the most common, and most tasks with Dynamic, Body or Competition stages come from the new suites (86 of 99).

%% file: sections/3_experiments/experiments.tex
\section{Analysis}\label{sec:experiments}

We evaluate eight coding agents on the core set (\Cref{fig:leaderboard}). \Cref{sec:results} gives their success rates, and \Cref{sec:f-settings,sec:f-difficulty,sec:f-agents,sec:f-control} report four findings. An agent is a model in a harness, such as Codex~\citep{openai2025codex} or Claude Code~\citep{anthropic2025claudecode}, at high reasoning effort, and the harness is part of what we measure (\Cref{sec:f-agents}). Each agent makes one attempt per test, and a test is solved when the verifier passes it (\Cref{app:protocol}). Success rates are the results of the public leaderboard (\Cref{app:results}).

\begin{figure}[H]
  \centering
  \begin{subfigure}[t]{0.490\linewidth}
    \includegraphics[width=\linewidth]{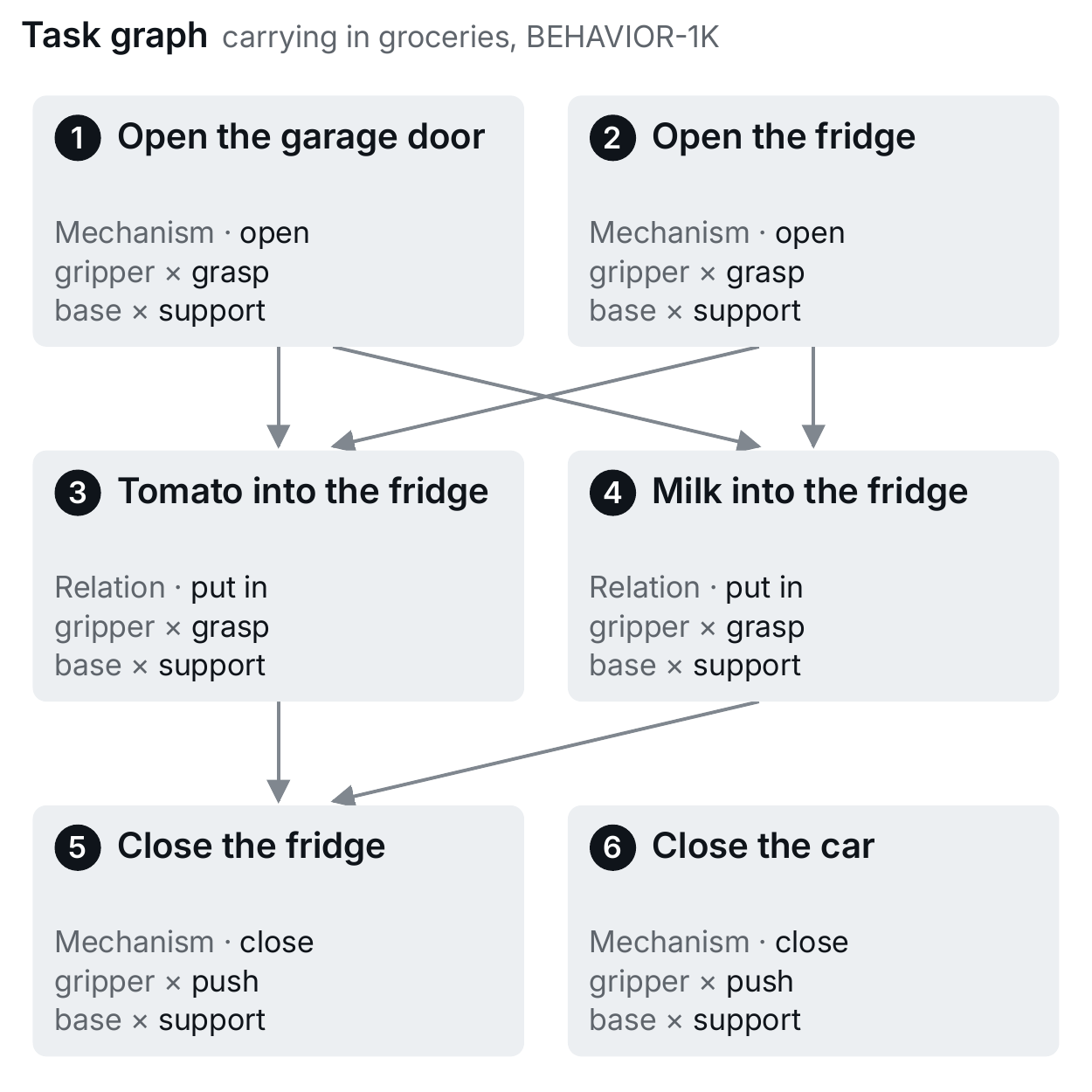}
    \caption{}
    \label{fig:fa-method-a}
  \end{subfigure}\hfill
  \begin{subfigure}[t]{0.490\linewidth}
    \includegraphics[width=\linewidth]{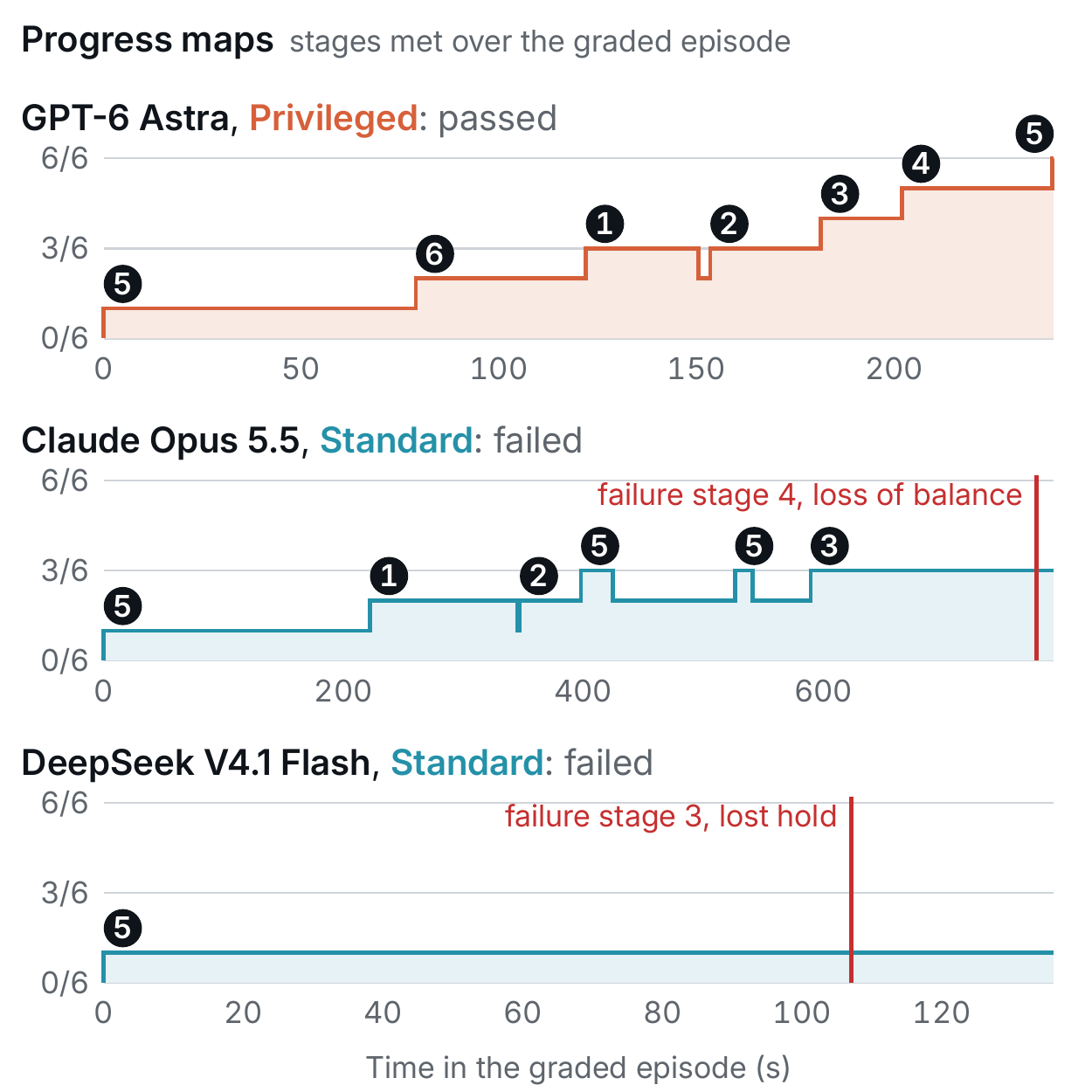}
    \caption{}
    \label{fig:fa-method-b}
  \end{subfigure}
  \caption{Task graph and progress maps of \emph{carrying in groceries} (BEHAVIOR-1K). Panel (a) gives the stages, the order in which they must be met, and the type and contact of each stage. Panel (b) gives the share of the stages whose conditions hold at each moment of the graded episode, for three attempts. Red lines mark failure stages and their failure phenomena (\Cref{app:fa-cases}).}
  \label{fig:fa-method}
\end{figure}

  \paragraph{Stage-level failure labels.}\label{sec:labels} \looseness=-1 The verifier decides whether an attempt passed, but not where it went wrong. We therefore split 427 of the 456 core tasks into stages in a task graph (\Cref{fig:fa-method-a}). Its edges give the order in which the stages must be met. Each stage has a type, such as \emph{open}, in one of nine categories (\Cref{tab:fa-steps}), and the contact it needs. \Cref{fig:fa-method-b} shows how many stages hold during three attempts. A failed attempt has at most one failure stage: the stage of an irreversible event, such as a fall, or else of the earliest mistake that the agent did not correct. This mistake can come before the stage's prerequisites are met, as when an agent drops the tomato before it has opened the fridge. It also has up to two failure phenomena. Claude Opus 5.5, for example, falls while carrying the milk, so it fails at stage 4 with a loss of balance. An attempt reaches a stage if it meets the stage's condition or fails there. We analyze the failed attempts of the eight agents on these 427 tasks. GPT-6.1 Sol in Codex wrote the labels of most attempts and Claude Sonnet 5.5 in Claude Code those of the remaining attempts; Claude Sonnet 5.5 reviewed every label file (\Cref{app:fa-procedure}).

\subsection{Overall Results}\label{sec:results}

\begin{figure}[H]
  \centering
  \includegraphics[width=\linewidth]{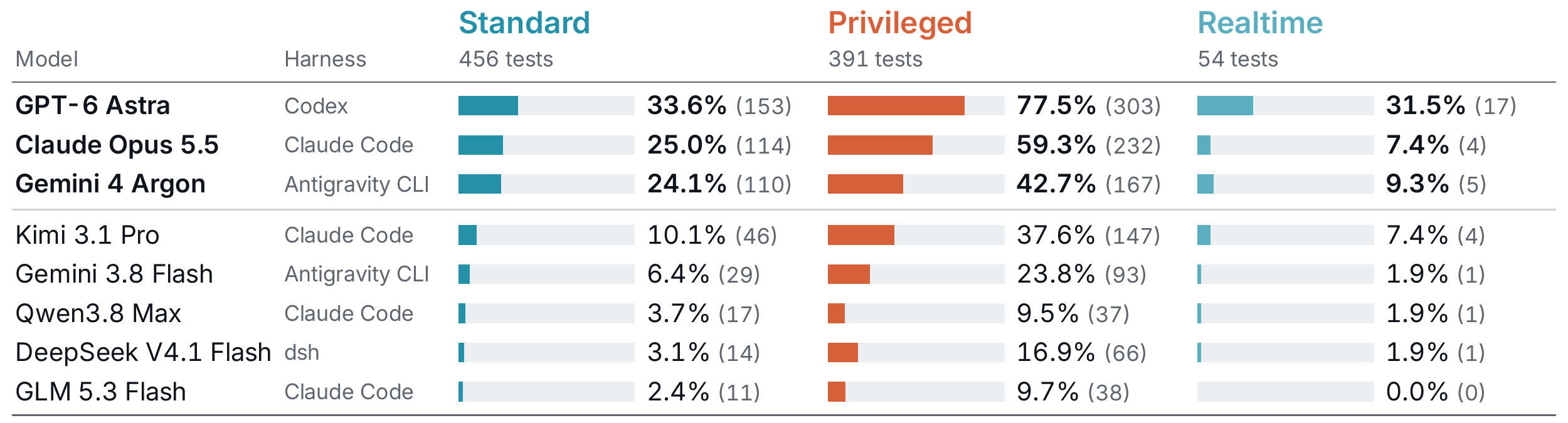}
  \caption{Success rate of each agent on the core set, with solved tests in parentheses. The harness dsh is DeepSeek Harness, DeepSeek's own harness.}
  \label{fig:leaderboard}
\end{figure}

\begin{figure}[!b]
  \centering
  \includegraphics[width=\linewidth]{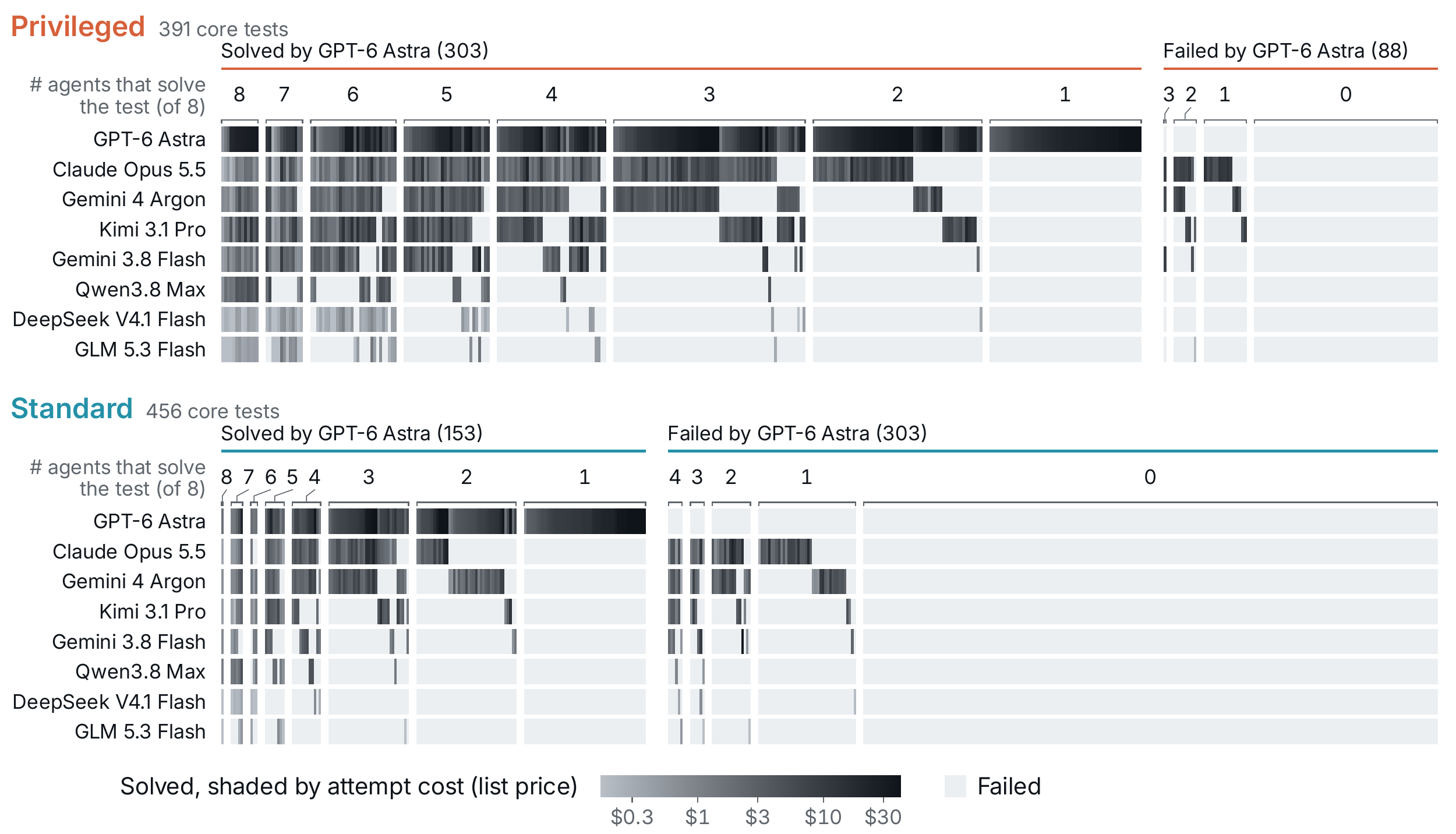}
  \caption{Every attempt on the core \evalsetting{Privileged} and \evalsetting{Standard} tests, with the agents in the order of \Cref{fig:leaderboard}.}
  \label{fig:task-agent-matrix}
\end{figure}

\textbf{Three agents each solve more than twice as many \evalsetting{Standard} tests as any of the other five.} GPT-6 Astra, Claude Opus 5.5 and Gemini 4 Argon solve 24.1\% to 33.6\% of the 456 \evalsetting{Standard} tests (\Cref{fig:leaderboard}). We call them the three strongest agents and the other five the weaker agents. The \evalsetting{Realtime} Setting is harder than the \evalsetting{Standard} Setting. On the 54 tasks with both tests, GPT-6 Astra solves 23 \evalsetting{Standard} and 17 \evalsetting{Realtime} tests, Claude Opus 5.5 solves 10 and 4, and Gemini 4 Argon solves 15 and 5. 63\% of these \evalsetting{Realtime} tests remain unsolved by any agent, compared with 50\% of the \evalsetting{Standard} tests of the same tasks.

\looseness=-1 \textbf{Together, the three strongest agents solve 48\% of the \evalsetting{Standard} tests.} \Cref{fig:task-agent-matrix} shows the outcome of every attempt. In the \evalsetting{Standard} Setting, Claude Opus 5.5 and Gemini 4 Argon solve many tests that GPT-6 Astra fails, whereas in the \evalsetting{Privileged} Setting GPT-6 Astra alone solves nearly every test that any agent solves. The five weaker agents add only a few tests in either setting. \Cref{sec:f-agents} examines which tests each agent solves.

\looseness=-1 \textbf{In most suites with at least five \evalsetting{Standard} tests, at least half of these tests are solved by none of the three strongest agents.} This holds in 14 of these 22 suites. Of the 20 suites with at least five \evalsetting{Privileged} tests, it holds only in HumanoidBench (\Cref{fig:benchmarks}). Averaged over the three strongest agents, the success rate is at least 70\% in 10 \evalsetting{Privileged} suites and in no \evalsetting{Standard} suite.

%% file: sections/4_findings/findings.tex
\subsection{What Causes the Drop from the \evalsetting{Privileged} to the \evalsetting{Standard} Setting?}
\label{sec:f-settings}

\begin{findingbox}{\evalsetting{Privileged} versus \evalsetting{Standard}}
\begin{itemize}
  \item \textbf{Resets help GPT-6 Astra more than the exact object state does.} Resets raise its success rate from 28\% to 53\%, the state to 45\%.
  \item \textbf{Stages that need precision become much harder in the \evalsetting{Standard} Setting.} Their failure rate rises by 24 to 46 percentage points, that of coarse stages by 1 to 20.
  \item \textbf{At stages that need precision, resets help GPT-6 Astra, but the exact state alone does not.} Of its failures there, a run without the success signal solves 19\%, a run with the state 21\% and a run with resets 36\%.
\end{itemize}
\end{findingbox}

\begin{figure}[t]
  \centering
  \includegraphics[width=\linewidth]{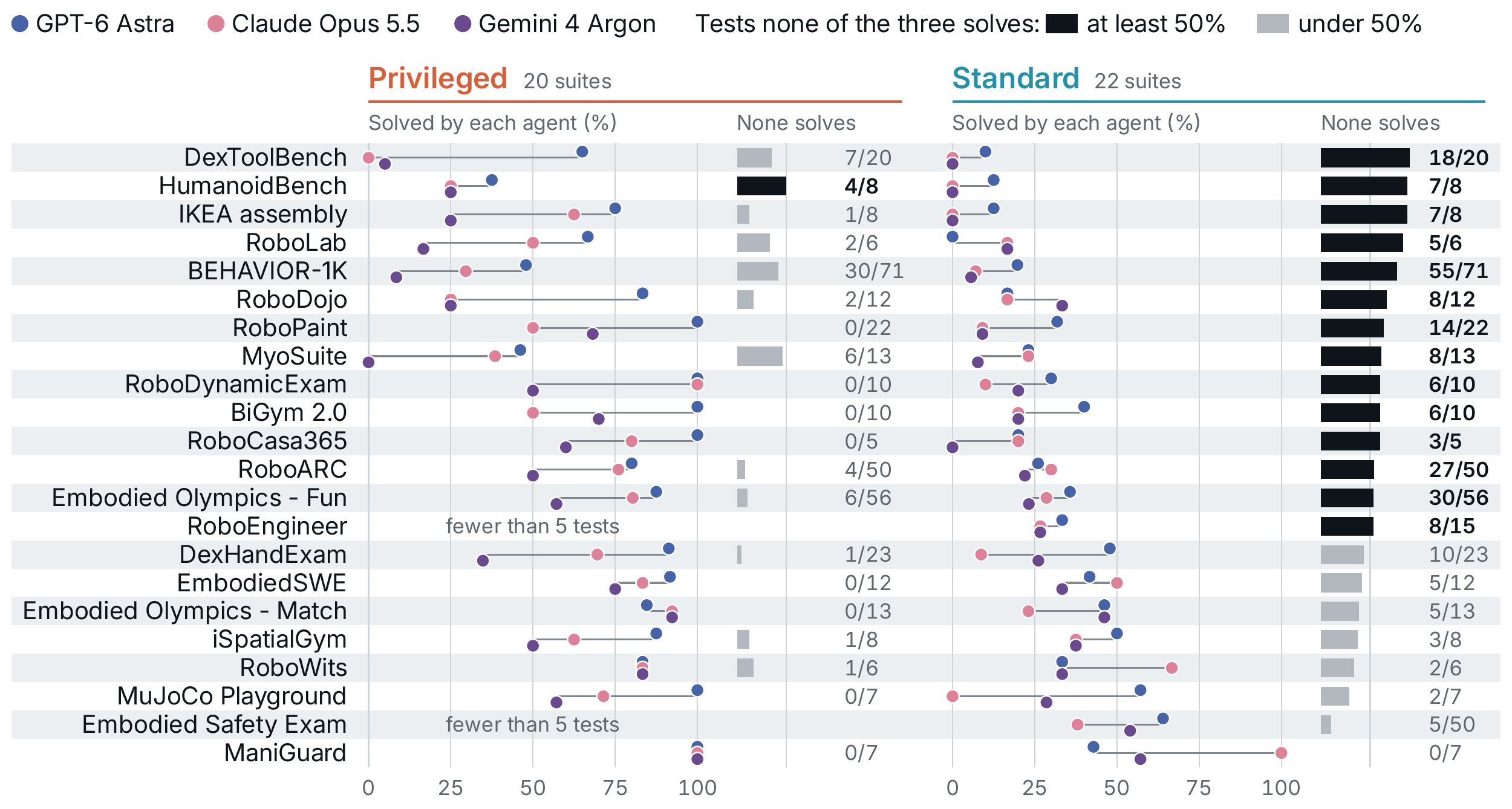}
  \caption{Success rates by suite, for suites with at least five core tests in the setting. Dots give the success rate of each of the three strongest agents, and bars the share of tests that none of them solves. The IKEA assembly tasks of RoboARC count as their own suite.}
  \label{fig:benchmarks}
\end{figure}

\begin{figure}[tb]
  \centering
  \includegraphics[width=\linewidth]{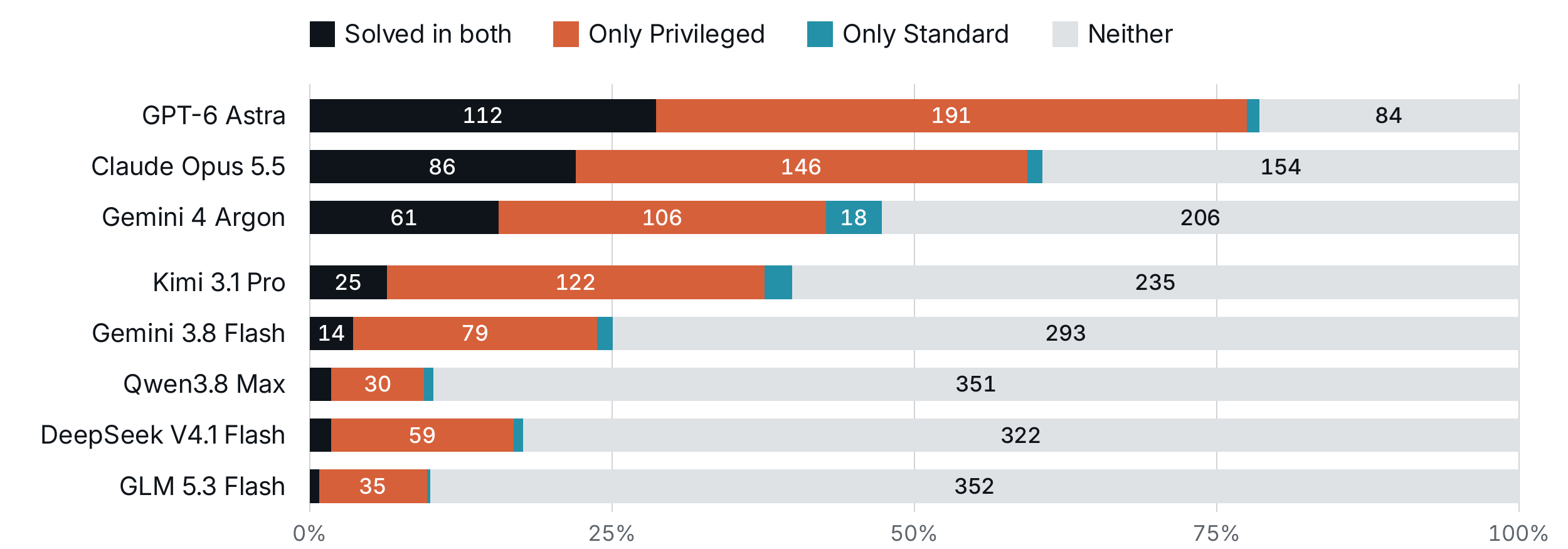}
  \caption{The 391 core tasks that have both a \evalsetting{Privileged} and a \evalsetting{Standard} test, split for each agent by the settings in which it solves them.}
  \label{fig:transfer}
\end{figure}

\Cref{fig:transfer} pairs the \evalsetting{Privileged} and \evalsetting{Standard} tests of the 391 core tasks that have both. Of the 303 tasks that GPT-6 Astra solves in the \evalsetting{Privileged} Setting, it fails 191 in the \evalsetting{Standard} Setting. It solves only 4 tasks in the \evalsetting{Standard} Setting alone, and 116 (30\%) in all. Its success rate drops in each of the 20 suites with at least five such tasks. Two of the features that the \evalsetting{Standard} Setting lacks are the exact state of every object and unlimited resets.

\begin{figure}[tb]
  \centering
  \begin{subfigure}[t]{0.3125\linewidth}
    \includegraphics[width=\linewidth]{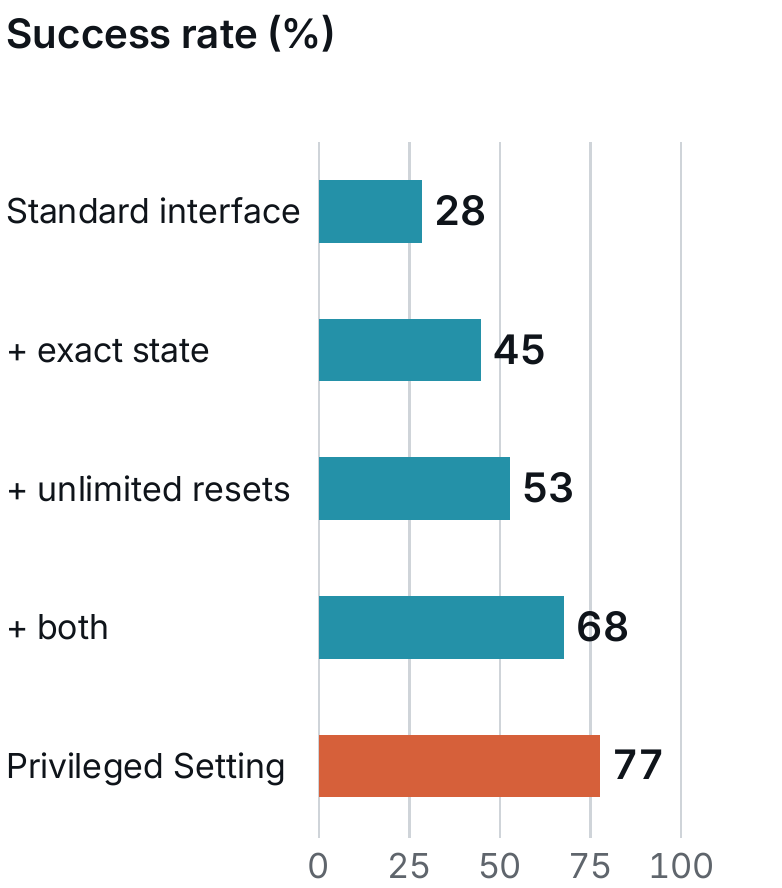}
    \caption{}
    \label{fig:ablation-ladder}
  \end{subfigure}\hfill
  \begin{subfigure}[t]{0.375\linewidth}
    \includegraphics[width=\linewidth]{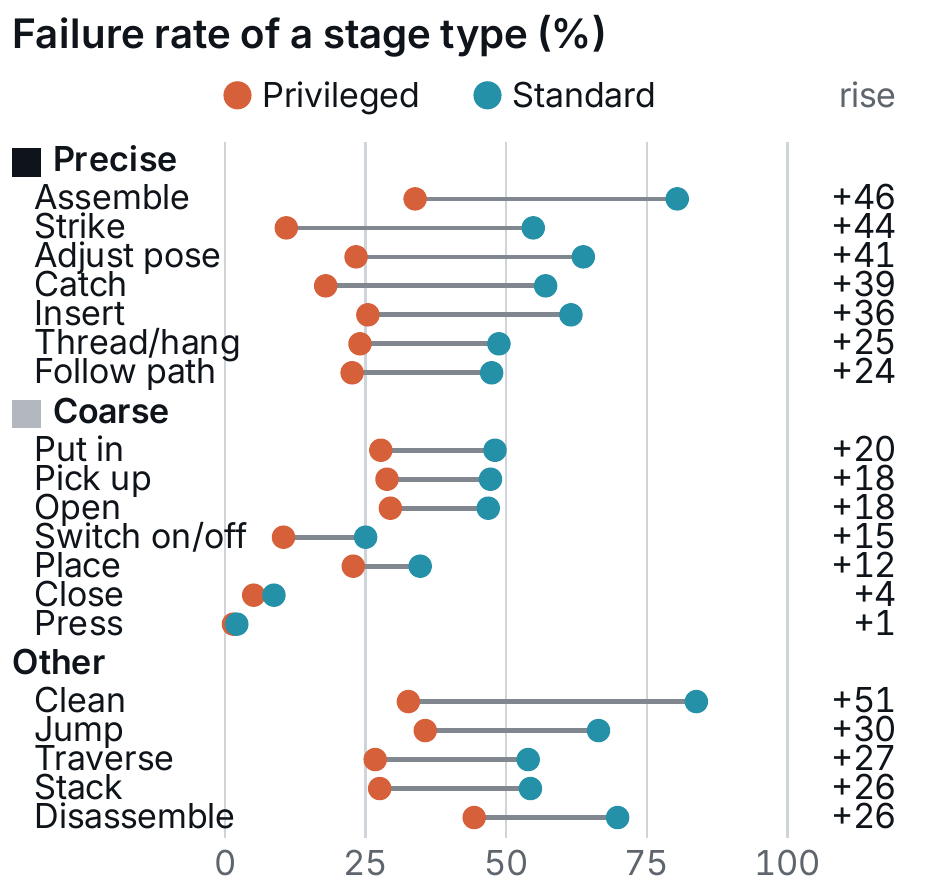}
    \caption{}
    \label{fig:fa-transfer-a}
  \end{subfigure}\hfill
  \begin{subfigure}[t]{0.3125\linewidth}
    \includegraphics[width=\linewidth]{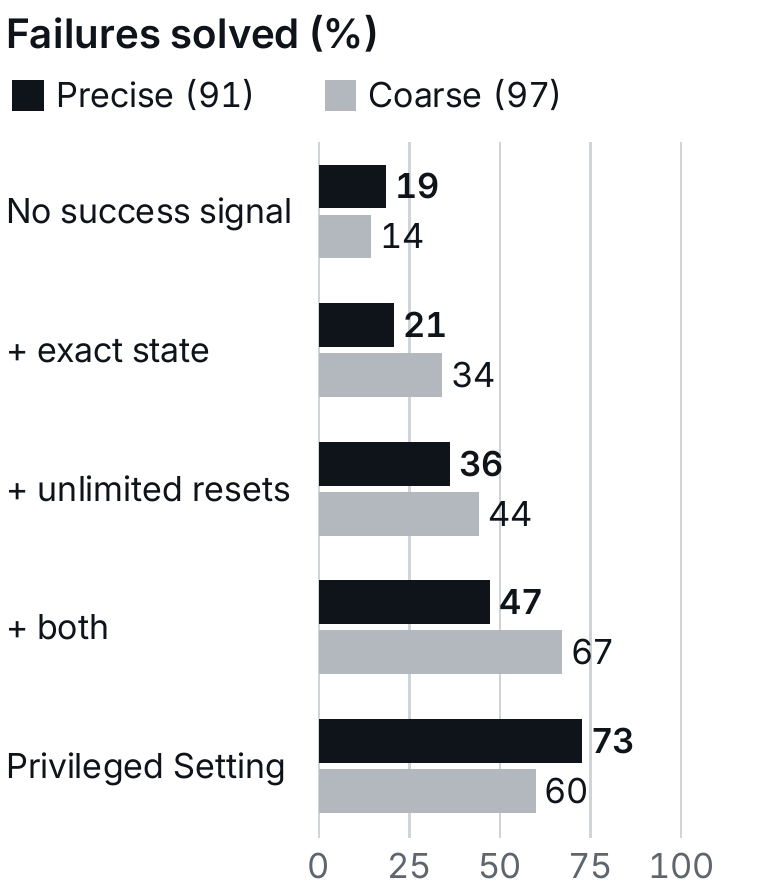}
    \caption{}
    \label{fig:ablation-stages}
  \end{subfigure}
  \caption{(a) Success rate of GPT-6 Astra in Codex on the 391 core tasks that have both settings, as features of the \evalsetting{Privileged} Setting are added to the \evalsetting{Standard} interface. (b) Failure rate of each stage type compared in at least two suites, eight agents. (c) Share of GPT-6 Astra's failures at precise and coarse stages that each run solves.}
  \label{fig:f-settings}
\end{figure}

Adding resets closes more of this gap than adding the exact state. In controlled runs of GPT-6 Astra in Codex, we add one or both of two features to the \evalsetting{Standard} interface: a query for the exact state of every object, and unlimited resets with saved states (\Cref{fig:ablation-ladder}, \Cref{app:ablation}). On the 391 core tasks that have both settings, the success rate rises from 28\% to 45\% with the state, to 53\% with resets and to 68\% with both. On their \evalsetting{Privileged} tests, it is 77\%. With resets, 66\% of the successes come in an episode that starts from a saved state, so the agent retries a hard stage without repeating the whole task (\Cref{app:an-settings}).

\begin{figure}[tb]
  \centering
  \includegraphics[width=\linewidth]{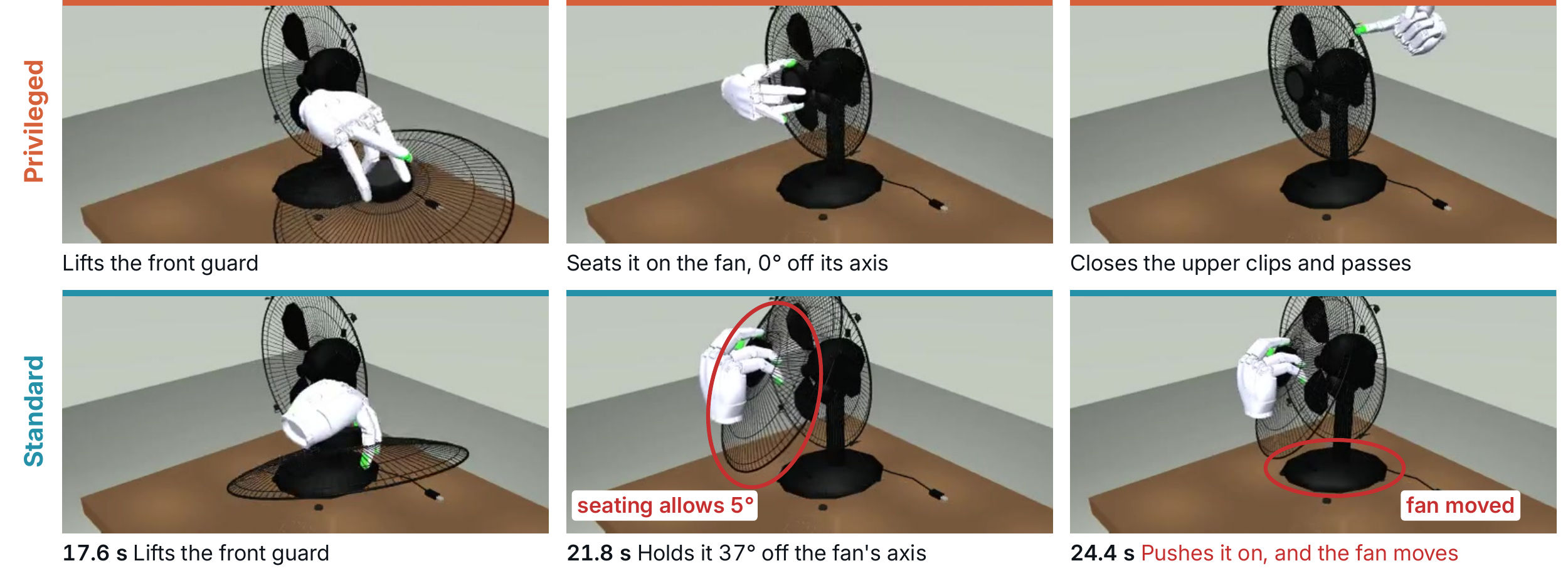}
  \caption{Claude Opus 5.5 installing the front guard of a fan (DexHandExam) in the \evalsetting{Privileged} and the \evalsetting{Standard} Setting. Red marks where the \evalsetting{Standard} attempt goes wrong. Times are seconds on the replay of the \evalsetting{Standard} episode.}
  \label{fig:case-fan-frames}
\end{figure}

\looseness=-1 Stages that need precision fail much more often in the \evalsetting{Standard} Setting. From the stage labels of the eight agents, we compute the failure rate of each stage type in each setting: the number of stages of the type at which attempts fail, divided by the number that attempts reach. We compare the settings within each suite (\Cref{fig:fa-transfer-a}). A stage type is precise when it needs a tight fit, a target pose or path, or contact with a moving object, and coarse when it needs only a region, a container, a grasp or the state of a mechanism (\Cref{app:fa-modes}). Among the types compared in at least two suites, the failure rate rises by 24 to 46 percentage points at precise types and by 1 to 20 points at coarse types. Assembling, for example, rises from 34\% to 80\%, and placing from 23\% to 35\%. If we count only stages that the same agent reaches on the same task in both settings, the difference between the two groups remains significant: precise stages rise by 25 points and coarse stages by 15 (\Cref{app:an-settings}). \Cref{fig:case-fan-frames} shows a precise stage that fails in the \evalsetting{Standard} Setting. There, Claude Opus 5.5 holds the front guard of a fan 36 to 39 degrees off the fan's axis, where seating allows 5 degrees, and pushing it further moves the whole fan. \Cref{app:fa-cases} gives the full attempt and two more paired examples (\Cref{fig:fa-case-f7}).

At precise stages, resets help GPT-6 Astra, but the exact state alone does not. We take the \evalsetting{Standard} tests that GPT-6 Astra failed in the baseline run and group them by the type of the stage at which they failed (\Cref{fig:ablation-stages}). Some of these failures would succeed in any second attempt. To estimate how many, we use a run that adds neither feature and only hides the success signal. Of the 91 failures at precise stages, this run solves 19\%, the run with the state 21\% and the run with resets 36\%, with 31 of its 33 successes coming after a reset or a restored state. Compared with the run without the success signal, the state thus solves about two more of the 91 failures, and resets about 16 more. At coarse stages, by contrast, the state alone solves 34\% of the 97 failures, against 14\% for the run without the success signal (\Cref{app:an-settings}).

\subsection{At Which Stage of a Task Do Agents Fail?}
\label{sec:f-difficulty}

\begin{findingbox}{Where Agents Fail}
\begin{itemize}
  \item \textbf{Failures on a task tend to happen at the same stage.} For the three strongest agents, 56\% of the failure pairs of a test share a stage on average, against 31\% by chance.
  \item \textbf{This stage also tends to be the same in the \evalsetting{Privileged} and \evalsetting{Standard} settings.} Leaving out tasks where it is the first stage in both settings, 57\% of tasks match, against 9\% by chance.
  \item \textbf{Stages that change a material or move the robot's body have the highest failure rates.} Once reached, they fail at 66\% and 45\%.
\end{itemize}
\end{findingbox}

\begin{figure}[tb]
  \centering
  \begin{subfigure}[t]{0.5\linewidth}
    \includegraphics[width=\linewidth]{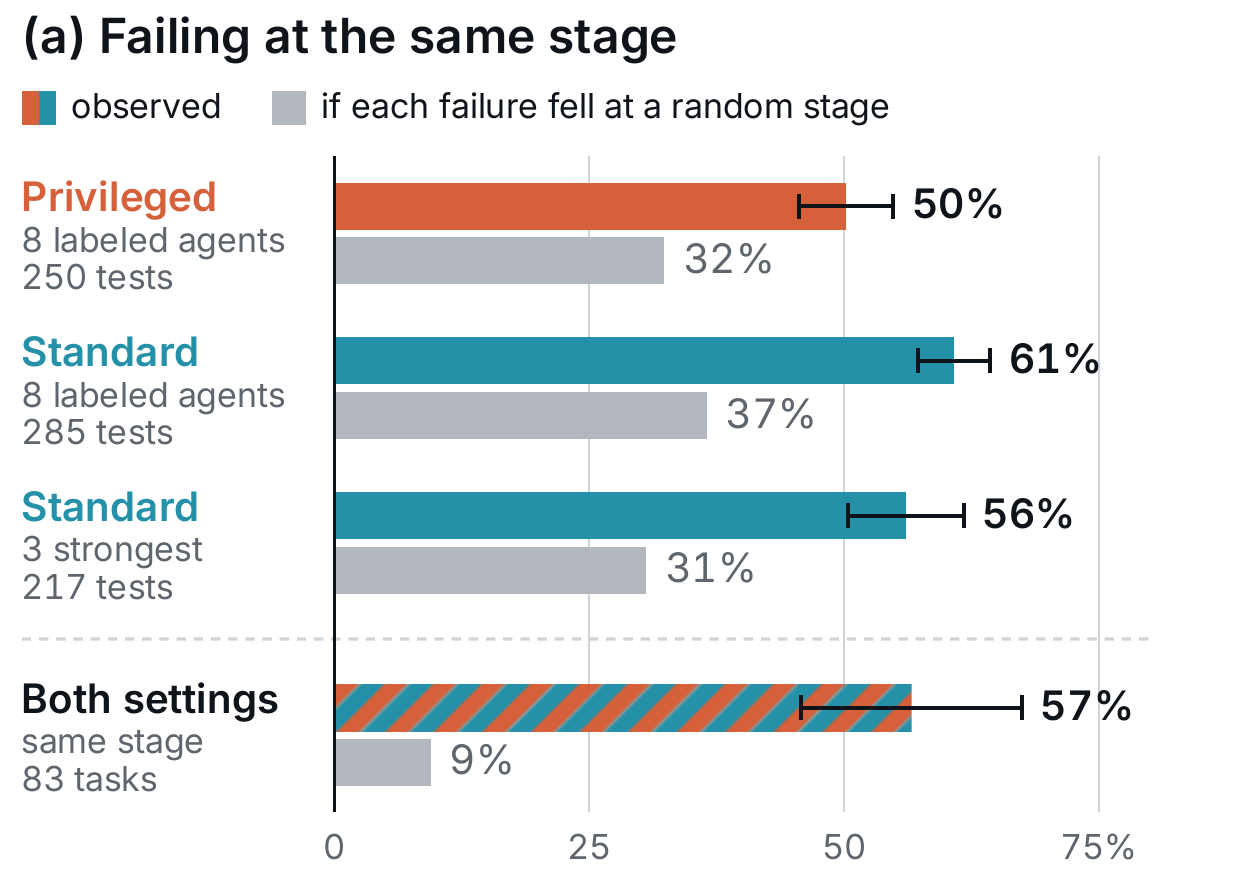}
    \phantomsubcaption
    \label{fig:fa-where-a}
  \end{subfigure}%
  \begin{subfigure}[t]{0.5\linewidth}
    \includegraphics[width=\linewidth]{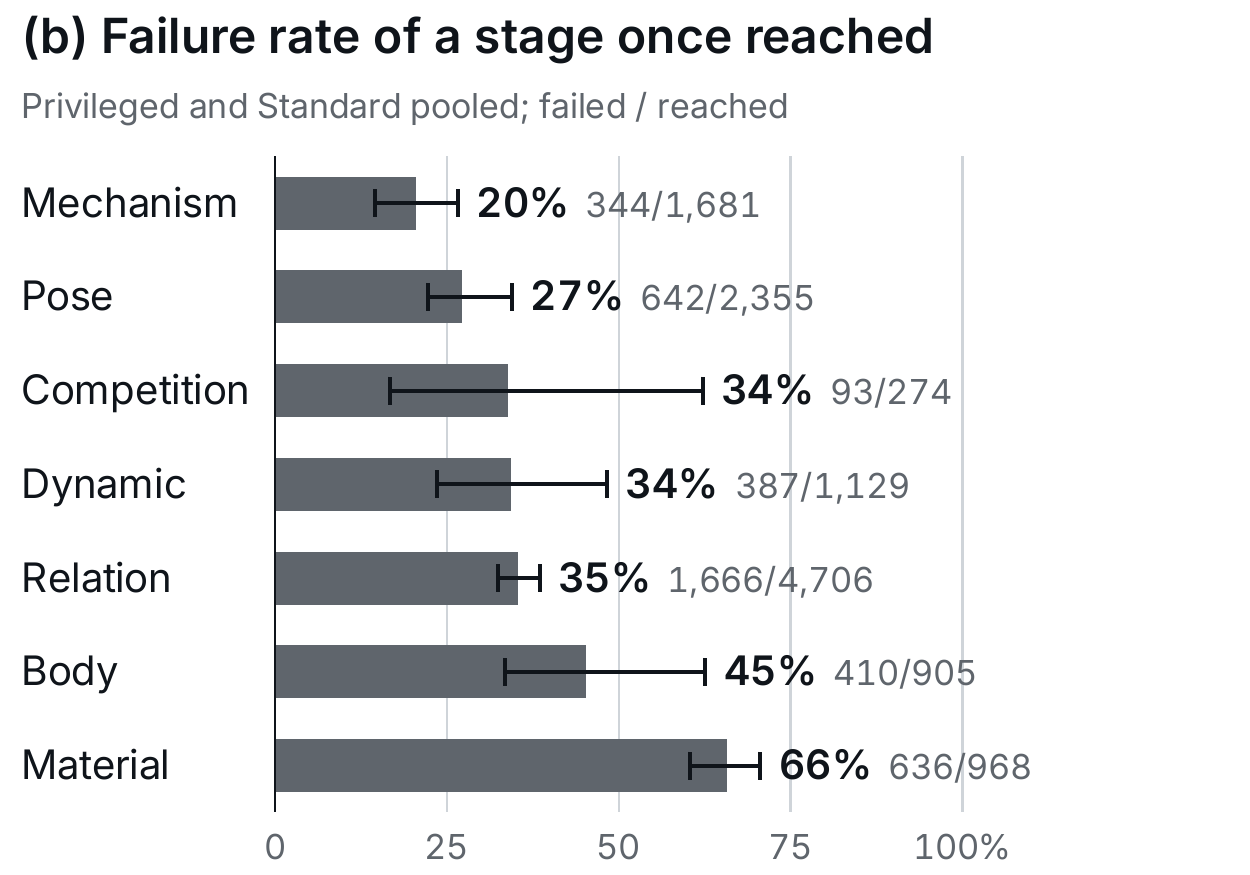}
    \phantomsubcaption
    \label{fig:fa-taxonomy-a}
  \end{subfigure}
  \caption{(a) Concentration of failure stages, mean over tests, and share of tasks with the same most common failure stage in both settings, without the tasks where it is the first stage in both. Gray bars: random failure stages; black lines: 95\% intervals. (b) Failure rate of a stage once reached, by category, both settings pooled (\Cref{tab:fa-pooled-rates}).}
  \label{fig:f-difficulty}
\end{figure}

\begin{figure}[tb]
  \centering
  \includegraphics[width=\linewidth]{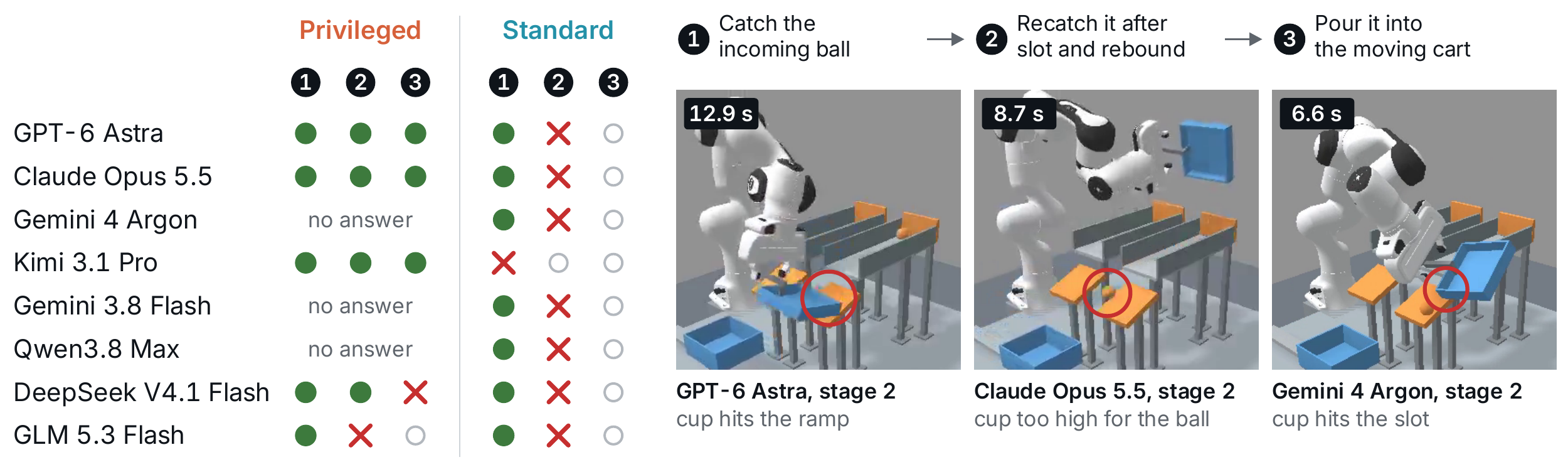}
  \caption{Catching a ball twice (RoboDynamicExam): the stages that the attempt of each agent reached and the stage at which it failed, in each setting, and \evalsetting{Standard} replay frames at three failures at stage 2.}
  \label{fig:case-ball}
\end{figure}

\looseness=-1 Failures on a task tend to happen at the same stage. The concentration of failure stages measures this (\Cref{fig:fa-where-a}). For each test, it is the share of pairs of failed attempts, by different agents, that failed at the same stage. We use the attempts with a located failure stage and average this share over tests. The baseline draws each failure stage at random, with the probability of each relative position in the task taken from the failures on the other tasks of the setting (\Cref{app:fa-stats}). Over the eight agents, the concentration is 61\% in the \evalsetting{Standard} Setting, against 37\% for the baseline. The five weaker agents often fail before they reach any stage of a task (\Cref{sec:f-agents}), so much of this agreement could come from failures at the first stage. We therefore check the three strongest agents separately. Over the 217 \evalsetting{Standard} tests with at least two located failures among them, the concentration is 56\%, against 31\% for the baseline built from their failures on the other \evalsetting{Standard} tasks. The most common failure stage of a test is usually the first stage of its task: of the 264 \evalsetting{Standard} tests whose failures have a single most common stage, it is the first stage in 65\% (\Cref{app:an-difficulty}). \Cref{fig:case-ball} shows a shared stage that is not the first. In the \evalsetting{Standard} Setting, all eight agents fail this test, and seven of them catch a thrown ball in a cup and then fail, in different ways, at the recatch. In \Cref{fig:fa-case-batteries} in \Cref{app:fa-cases}, the three strongest agents each put the first of three batteries into a bin and then tip the robot over on the way to the second, whereas four of the five weaker agents fail at the first battery.

The stage at which agents fail also tends to be the same in the two settings. We take the 188 tasks whose failures have a single most common stage in both the \evalsetting{Privileged} and the \evalsetting{Standard} Setting. In 105 of them, this stage is the first stage in both settings, and we leave these tasks out. Of the other 83 tasks, 47 (57\%) have the same common stage in both settings (\Cref{fig:fa-where-a}). If the failure stages of each setting were drawn at random as above, this share would be 9\%, again leaving out draws in which both common stages are the first stage (\Cref{app:fa-stats}).

\looseness=-1 Stages that change a material or move the robot's body have the highest failure rates (\Cref{fig:fa-taxonomy-a}). The failure rate of a stage category is the number of stages at which attempts fail divided by the number that they reach, with the \evalsetting{Privileged} and \evalsetting{Standard} settings pooled. Material stages, such as wiping, fail at 66\%, body stages, such as walking, at 45\%, and mechanism stages, such as opening a door, at 20\%. Each category occurs mostly in particular suites and with particular robots, so these rates do not show that the category itself causes the failures (\Cref{app:an-difficulty}).

\subsection{How Do Agents Differ?}
\label{sec:f-agents}

\begin{findingbox}{Differences between Agents}
\begin{itemize}
  \item \textbf{Weaker agents are much more likely than stronger agents to fail before making the required contact.} The five weaker agents do so at 9.1\% to 18.5\% of contact stages, the three strongest at 2.7\% to 3.7\%.
  \item \textbf{In the \evalsetting{Standard} Setting, the three strongest agents solve partly different sets of tests.} Between them, Claude Opus 5.5 and Gemini 4 Argon solve 64 tests that GPT-6 Astra fails.
  \item \textbf{Switching harnesses can change a model's success rate by up to 12 percentage points.} On the 456 core tests of the \evalsetting{Standard} Setting, GPT-6 Astra solves 34\% in its own Codex and 22\% in Claude Code.
\end{itemize}
\end{findingbox}

\begin{figure}[tb]
  \centering
  \begin{subfigure}[t]{0.5475\linewidth}
    \includegraphics[width=\linewidth]{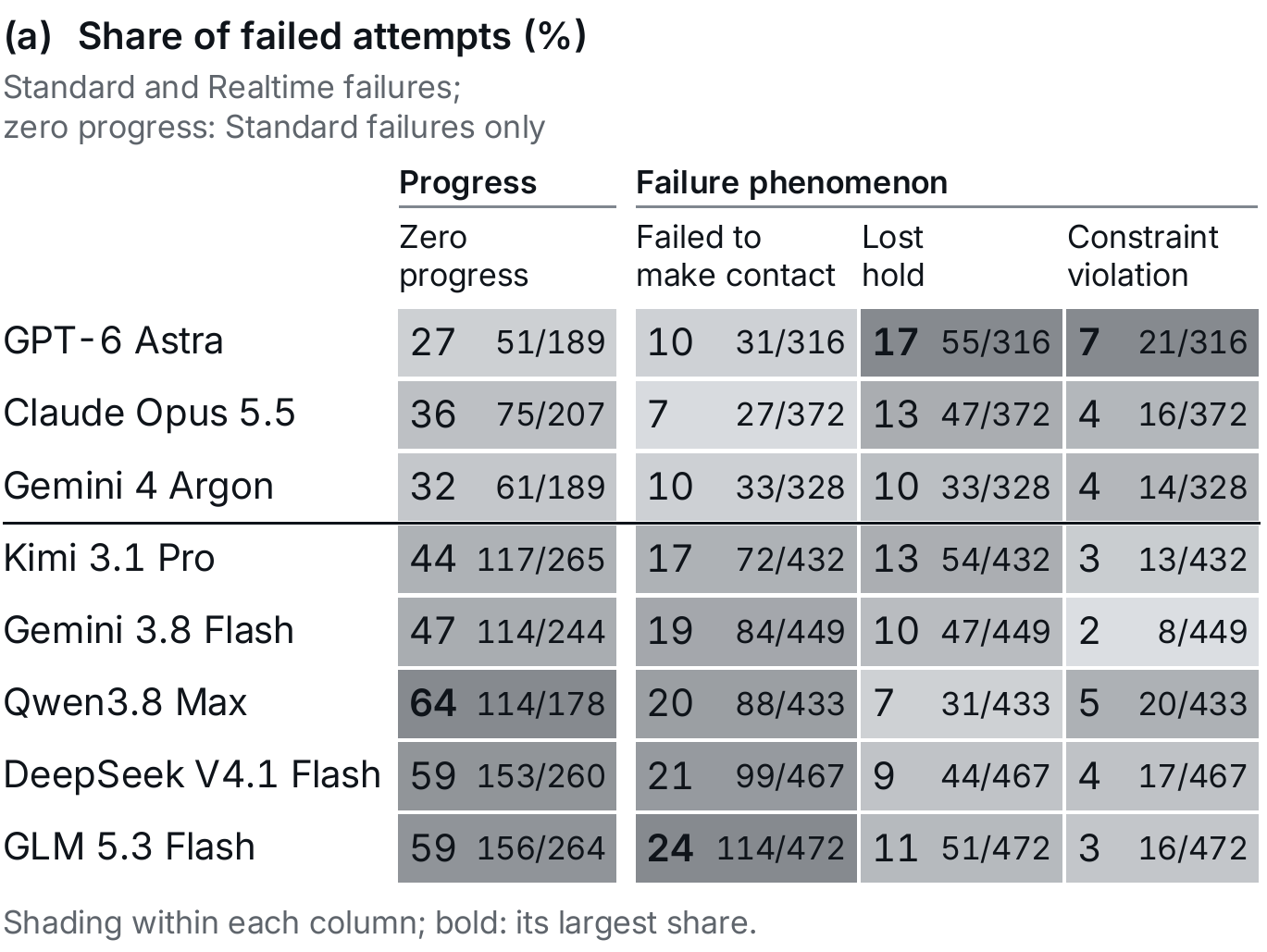}
    \phantomcaption
    \label{fig:fa-agents-a}
  \end{subfigure}%
  \begin{subfigure}[t]{0.4525\linewidth}
    \includegraphics[width=\linewidth]{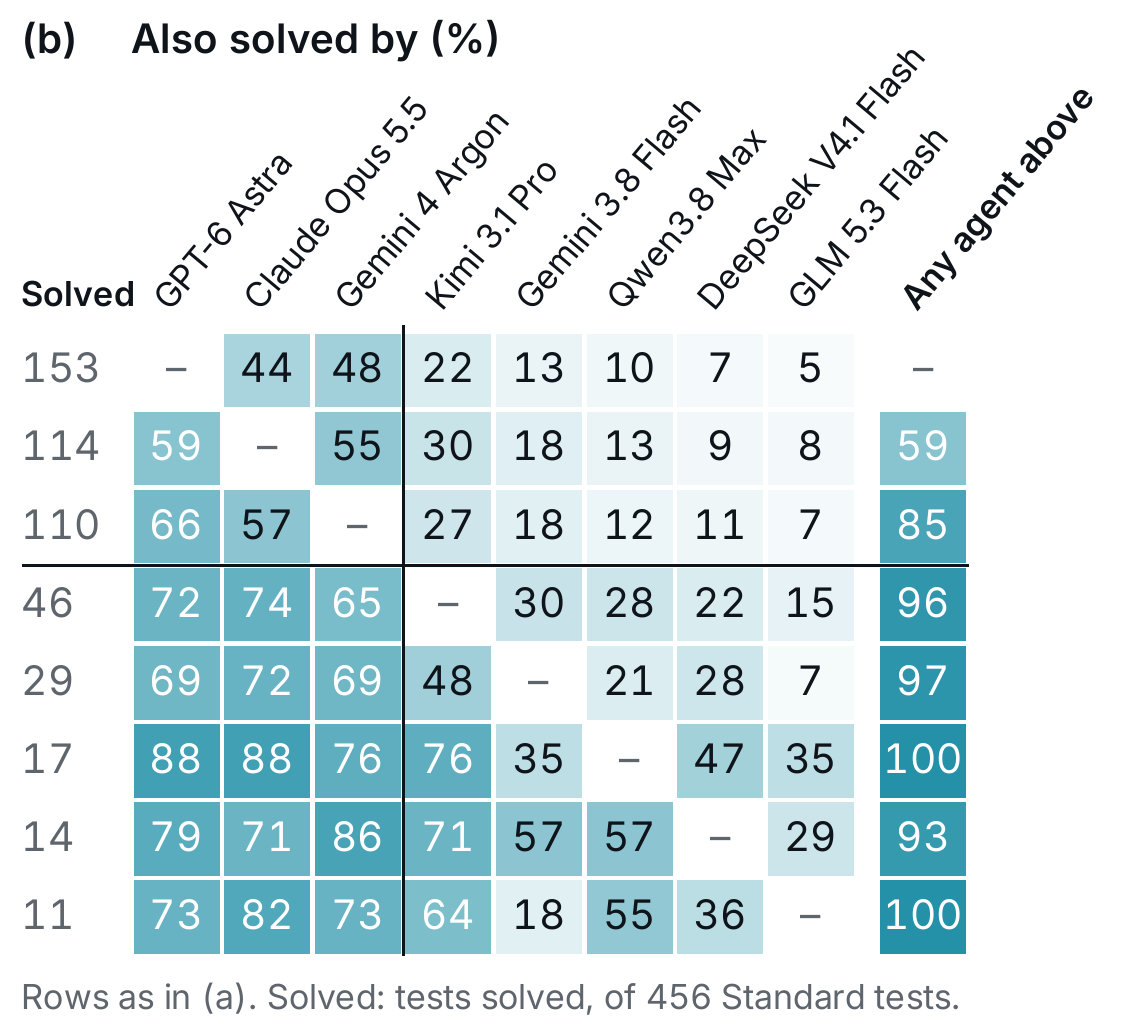}
    \phantomcaption
    \label{fig:solve-matrix}
  \end{subfigure}
  \par\smallskip
  \begin{subfigure}[t]{\linewidth}
    \includegraphics[width=\linewidth]{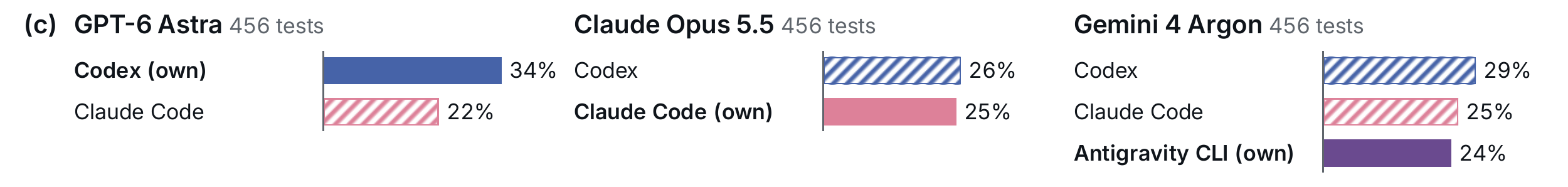}
    \phantomcaption
    \label{fig:ablation-harness}
  \end{subfigure}
  \caption{\looseness=-1 (a) Failure profile of each agent; zero progress means that the attempt reached no stage. (b) Share of the row agent's solved \evalsetting{Standard} tests (first column) that the column agent also solves; the last column uses any agent above it in \Cref{fig:leaderboard}. (c) Success rate by model and harness on the 456 core tests of the \evalsetting{Standard} Setting; hatched bars mark other harnesses.}
  \label{fig:f-agents}
\end{figure}

\begin{figure}[tb]
  \centering
  \includegraphics[width=0.87\linewidth]{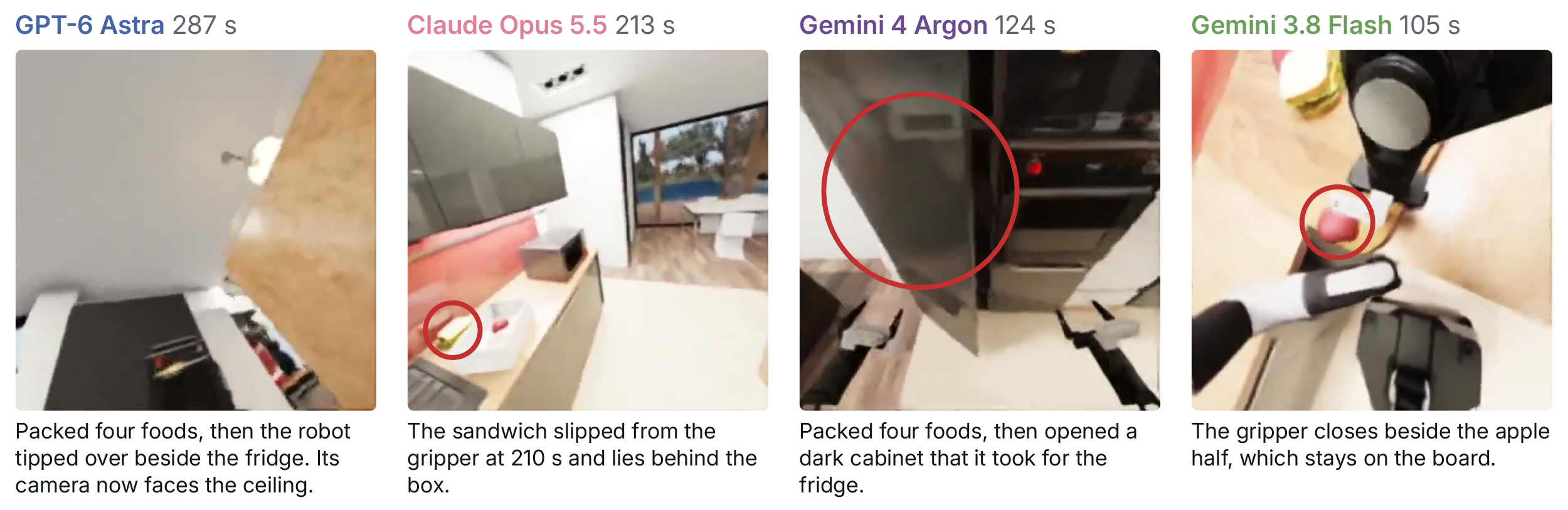}
  \caption{Preparing a lunch box (BEHAVIOR-1K, \evalsetting{Standard} Setting): a head-camera frame of four failed attempts, with the failure circled.}
  \label{fig:lunch-frames}
  \vspace{-6pt}
\end{figure}

The five weaker agents more often fail without making the required contact. In the failure profiles of \Cref{fig:fa-agents-a}, this happens in 17\% to 24\% of their failed \evalsetting{Standard} and \evalsetting{Realtime} attempts, against 7\% to 10\% for the three strongest agents. Of their failed \evalsetting{Standard} attempts on tasks with at least two stages, they reach no stage at all in 44\% to 64\%, against 27\% to 36\% (\Cref{tab:fa-agents}). Counted per contact stage that they reach, the weaker agents fail without contact at 9.1\% to 18.5\% of these stages, against 2.7\% to 3.7\% (\Cref{tab:fa-agents}). \Cref{fig:lunch-frames} shows a BEHAVIOR-1K test, preparing a lunch box, that all eight agents failed in the \evalsetting{Standard} Setting. The three strongest agents each packed three or four foods, Kimi 3.1 Pro packed two, and the other four weaker agents none. \Cref{app:fa-cases} describes the attempts and two more tests (\Cref{fig:fa-case-f8}). GPT-6.1 Sol, which wrote most of the labels, is from the same family as GPT-6 Astra, which could bias these comparisons (\Cref{app:fa-stats}).

The three strongest agents succeed on partly different tests. In the \evalsetting{Standard} Setting, no agent's solved tests are a subset of another agent's (\Cref{fig:solve-matrix}). GPT-6 Astra solves 59\% of Claude Opus 5.5's solved \evalsetting{Standard} tests and 66\% of Gemini 4 Argon's. In the \evalsetting{Privileged} Setting, these shares are 92\% and 95\% (\Cref{fig:solve-matrix-priv}). Between them, Claude Opus 5.5 and Gemini 4 Argon solve 64 \evalsetting{Standard} tests that GPT-6 Astra fails. Each agent made one attempt per test, so part of this difference may come from the variation between runs. Of the 69 \evalsetting{Standard} tests that at least one weaker agent solves, the three strongest agents solve 65. For each weaker agent, 93\% to 100\% of its solved tests are also solved by at least one agent above it in \Cref{fig:leaderboard}, as the last column of \Cref{fig:solve-matrix} shows (\Cref{app:an-agents}).

\looseness=-1 The harness also changes the result (\Cref{fig:ablation-harness}). We ran GPT-6 Astra in Claude Code, Claude Opus 5.5 in Codex, and Gemini 4 Argon in both. On the 456 core tests of the \evalsetting{Standard} Setting, GPT-6 Astra solves 34\% in Codex and 22\% in Claude Code. Gemini 4 Argon solves 29\% in Codex, 25\% in Claude Code and 24\% in its own Antigravity CLI, and Claude Opus 5.5 solves 26\% in Codex and 25\% in Claude Code. Cross-provider runs go through an adapter (\Cref{app:ablation}). Antigravity CLI ends the session when Gemini's content filter blocks a reply, which may explain part of Gemini 4 Argon's gap (\Cref{app:an-agents}).

\subsection{Which Control Approach Do Agents Use?}
\label{sec:f-control}

\begin{findingbox}{How Agents Control the Robot}
\begin{itemize}
  \item \textbf{In the \evalsetting{Standard} and \evalsetting{Realtime} settings of a soccer task, agents trained a policy when given a GPU and a training library, and only trained policies won.}
  \item \textbf{In the \evalsetting{Standard} Setting, agents mostly control the robot directly.} They do so in 61\% of the sampled \evalsetting{Standard} attempts.
  \item \textbf{In the \evalsetting{Privileged} and \evalsetting{Realtime} settings, agents mostly write programs to control the robot.} They do so in 82\% of the sampled \evalsetting{Privileged} attempts and 97\% of the \evalsetting{Realtime} attempts.
\end{itemize}
\end{findingbox}

\begin{figure}[tb]
  \centering
  \begin{subfigure}[t]{0.6\linewidth}
    \includegraphics[width=\linewidth]{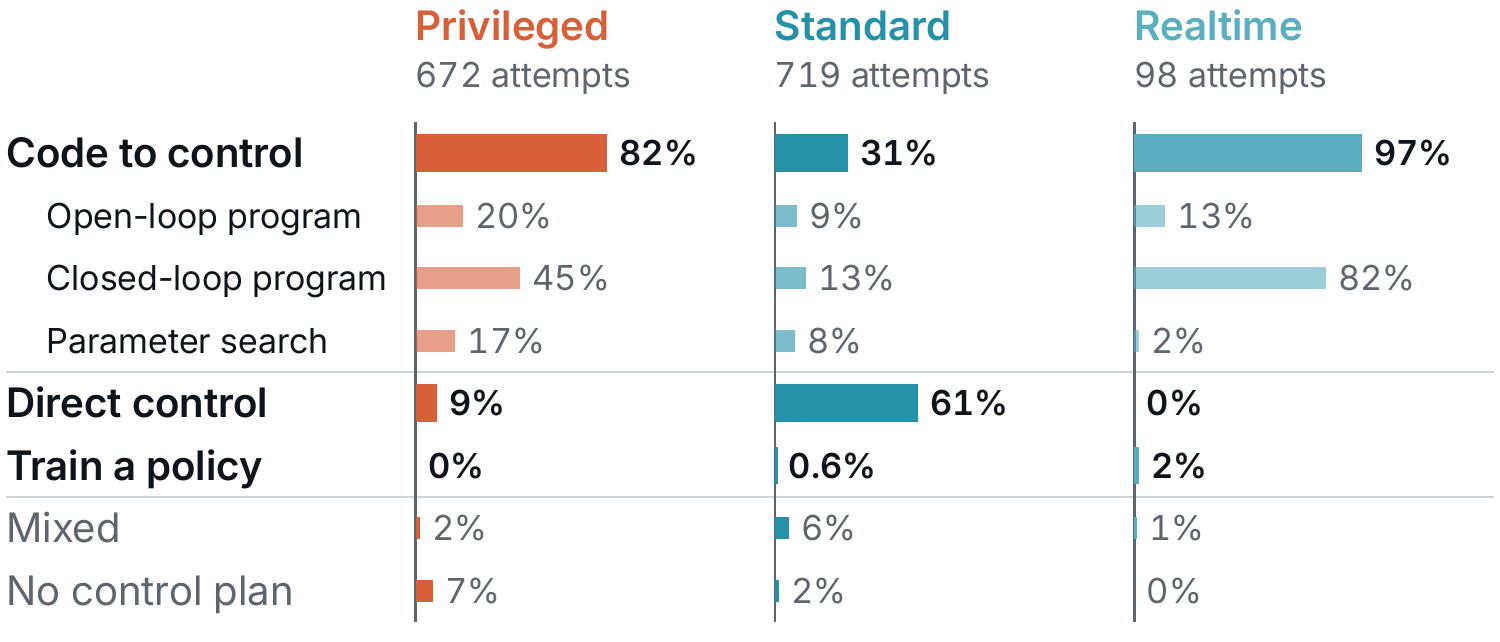}
    \caption{}
    \label{fig:approach-setting}
  \end{subfigure}\hfill
  \begin{subfigure}[t]{0.375\linewidth}
    \includegraphics[width=\linewidth]{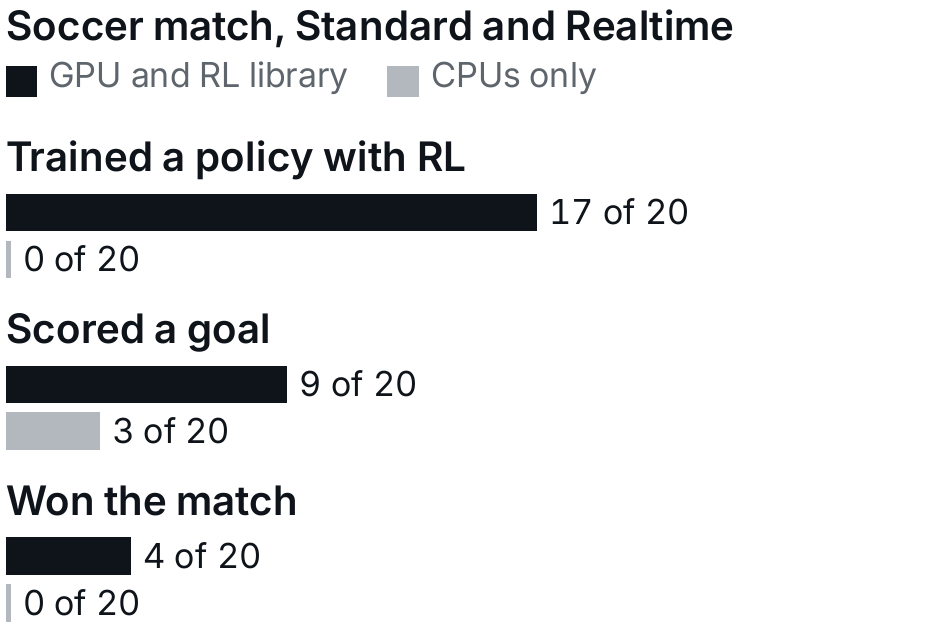}
    \caption{}
    \label{fig:approach-soccer}
  \end{subfigure}
  \caption{(a) Share of each approach by setting in 1,489 sampled attempts. Mixed attempts switched approach midway or used two approaches equally. (b) Soccer attempts in the \evalsetting{Standard} and \evalsetting{Realtime} settings that trained a policy, scored or won.}
  \label{fig:approach}
\end{figure}

\begin{figure}[tb]
  \centering
  \includegraphics[width=\linewidth]{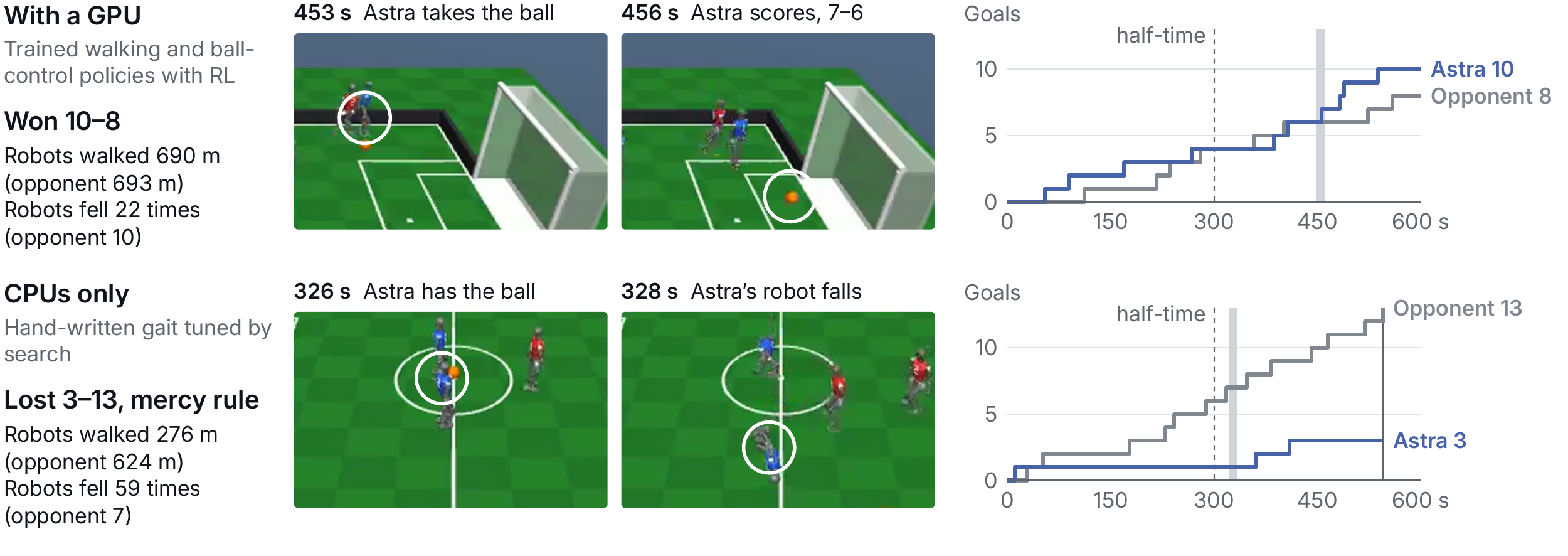}
  \caption{GPT-6 Astra in Codex on the two soccer versions in the \evalsetting{Standard} Setting. Each row gives the approach and result, two replay frames and the goals over match time.}
  \label{fig:approach-case}
\end{figure}

In the \evalsetting{Standard} and \evalsetting{Realtime} settings of a soccer task, agents trained a policy when given a GPU and a training library, and only trained policies won. The soccer match of Embodied Olympics - Match comes in two versions with the same frozen scene and opponent (\Cref{fig:approach-soccer}). In both, the agent controls two humanoid robots joint by joint, so it needs a walking gait. The GPU version adds a GPU, a reinforcement-learning library, GPU copies of the match for training and an example of training in its instruction. On this version, 10 of 12 attempts trained a policy with reinforcement learning in the \evalsetting{Standard} Setting, and 7 of 8 did in the \evalsetting{Realtime} Setting. Three attempts won in the \evalsetting{Standard} Setting and one in the \evalsetting{Realtime} Setting, all with a trained policy. On the CPU version, agents wrote gaits by hand, and none of the 12 \evalsetting{Standard} and 8 \evalsetting{Realtime} attempts won. The \evalsetting{Standard} counts include the runs of the harness comparison, so some models appear more than once (\Cref{tab:soccer-tiers}). In the \evalsetting{Privileged} Setting, agents also won on the CPU version, but every win there reused the opponent's walking network. The two versions differ in several ways, and this one task cannot show which difference matters. GPT-6 Astra, for example, trained policies and won 10--8 with the GPU, and tuned a hand-written gait and lost 3--13 without it (\Cref{fig:approach-case}).

\looseness=-1 In the \evalsetting{Standard} Setting, agents mostly control the robot directly, and in the \evalsetting{Privileged} and \evalsetting{Realtime} settings they mostly write programs (\Cref{fig:approach-setting}). A coding agent labeled the control approach of 1,489 sampled attempts (\Cref{app:an-control}). Where the simulation pauses between requests, 61\% of the sampled \evalsetting{Standard} attempts send one action chunk at a time and look at the camera images or sensor readings before choosing the next. With the full state, 82\% of the \evalsetting{Privileged} attempts compute the motion in a program. When the clock keeps running, 97\% of the \evalsetting{Realtime} attempts use a program, and in 82\% the program runs the control loop. Six of the 1,489 attempts were labeled as training a policy.

%% file: sections/5_related_work/related_work.tex
\section{Related Work}
\label{sec:related}

This section reviews the closest lines of work. \Cref{app:related} gives an extended review.

\paragraph{Benchmarks for coding agents.} \looseness=-1 Coding agents and other language-model agents are evaluated mostly on digital tasks, such as resolving issues in real repositories~\citep{jimenez2023swebench,miserendino2025swelancer}, completing tasks in a terminal~\citep{merrill2026terminalbench}, machine learning engineering and research~\citep{chan2024mlebench,starace2025paperbench}, and operating websites and desktop applications~\citep{zhou2023webarena,xie2024osworld}. The interface between the model and the computer affects how often an agent succeeds~\citep{yang2024sweagent}, and other work measures how often agents exploit the tests that grade them~\citep{zhong2025impossiblebench}. The \efe{} extends these evaluations to robotics tasks. In its \evalsetting{Standard} and \evalsetting{Realtime} settings, the agent must interpret sensor observations and often cannot undo an action. In every setting, a program that the agent cannot change grades each attempt.

\paragraph{Language models for robots.} Language models have been used to sequence skills into plans~\citep{ahn2022saycan,huang2022inner}, to write robot programs that call perception and control primitives~\citep{liang2022code}, to specify value maps or keypoint constraints that a planner turns into motions~\citep{huang2023voxposer,huang2024rekep}, and to write reward functions~\citep{ma2024eureka}. Most of these systems pair a model with an interface designed for it. The \efe{} evaluates unchanged general-purpose coding agents on the whole loop of reading an interface, choosing a control approach, carrying it out and revising it. In embodied benchmarks for language models~\citep{shridhar2021alfworld,li2024embodied,yang2025embodiedbench}, the model chooses from an action space that the benchmark defines. The exam gives the agent the robot's control interface, so the agent also decides how its actions are produced.

\paragraph{Agentic robots and their harnesses.} \looseness=-1 Recent work runs general-purpose agents on robots~\citep{qin2026agentic} with the three control approaches of \Cref{fig:approaches}. In direct control, GPT-6 Astra serves as the policy on manipulation benchmarks~\citep{zhang2026unexpected,zhou2026gpt6,jedlicki2026memory}. In code to control, coding agents write programs for simulated and real robots~\citep{tsui2026demonstration,jia2026agent,huang2026rax,ilie2026fetch2}, often with skill libraries or other added tools~\citep{lu2026aspire,hu2026simex}. In train a policy, agents train policies in the real world or in simulation~\citep{xiao2026enpire,li2026harbor}. Most of these studies fix the control approach in advance. The exam supports all three approaches and leaves the choice to the agent. Harnesses for these agents differ in the action interface they provide~\citep{chen2026showharness,gao2026hexashell,chu2026openrua}, in whether the agent can call learned policies as tools~\citep{li2026roboclaw,galanti2026physical}, and in the support they add for tracking state and checking outcomes~\citep{wang2026thea,fang2026emerge}. The exam runs each harness unmodified with the models of its own provider, and through an adapter with other providers' models (\Cref{app:ablation}), so different harnesses can be compared on the same tasks.

\paragraph{Benchmarks for robots.} \looseness=-1 Most robotics benchmarks focus on visuomotor policy learning and evaluate learned policies, such as vision-language-action models~\citep{brohan2023rt2,kim2024openvla,black20240,team2025gemini}, on manipulation, household and humanoid tasks~\citep{liu2023libero,chen2025robotwin,li2024behavior1k,nasiriany2024robocasa,sferrazza2024humanoidbench}. A learned policy receives only a fixed set of observations. A coding agent can run any program on its computer, so the exam adds rules about what the agent may access (\Cref{sec:settings}) and adapts 26 existing benchmarks (\Cref{app:tasks-adapted}). Benchmarks for agentic robots have also appeared. CaP-X~\citep{fu2026capx} varies the level of abstraction of the primitives that coding agents compose, RobotWorld~\citep{yang2026robotworld} spans manipulation, locomotion, driving and aerial control, and EmbodiedSWE~\citep{shen2026embodiedswe} and BiGym~2.0~\citep{zhang2026bigym} evaluate coding agents on dexterous and humanoid tasks. The \efe{} covers 47 robots of nine types. It also separates what an agent can do with full access to the simulator from what it can do through the robot's sensors alone.

%% file: sections/6_conclusion/conclusion.tex
\section{Conclusion}
\label{sec:conclusion}

We introduced the \efe{}, a unified evaluation framework that connects any coding agent to simulated robots and lets the agent execute actions in three ways. The exam contains 1,156 tasks on 47 robots. On the 391 hard core tasks that have both settings, the strongest agent reaches a success rate of 77\% in the \evalsetting{Privileged} Setting but only 30\% in the \evalsetting{Standard} Setting. In controlled runs with this agent, adding unlimited resets closes more of this gap than adding the exact object state, and at stages that need precision, resets help but the exact state alone does not. Failures on a task tend to happen at the same stage, weaker agents are much more likely than stronger agents to fail before making the required contact, and in the \evalsetting{Standard} and \evalsetting{Realtime} settings of a soccer task, agents trained a policy when given a GPU and a training library, and only trained policies won. The strongest agent already solves most tasks when it has full access to the simulator, but in the \evalsetting{Standard} and \evalsetting{Realtime} settings, which are designed to mirror how a real robot is deployed and tested, it still has much room for improvement.

\paragraph{Future work.} \looseness=-1 Running each core test several times will separate failures that an agent repeats from failures of a single run, and it will give confidence intervals for the comparisons between agents and between settings. We will also add more \evalsetting{Realtime} tasks, since the \evalsetting{Realtime} Setting has the largest share of tests that no agent solves. Finally, because the \evalsetting{Standard} Setting gives the agent only the robot's sensor readings and control interface, the same interface can be used on physical hardware once there is a way to grade the outcome. In a first demonstration, two agents played a buzz wire game with a physical robot arm (\Cref{app:real}), and we plan to evaluate agents on real robots systematically through this interface.

\paragraph{Limitations.} Each agent makes a single attempt per test, the stage-level and control-approach labels come from agents, and all results are in simulation. \Cref{app:limitations} discusses these and further limitations.

%% file: appendix_body.tex
\clearpage
\appendix
\pdfbookmark[0]{Appendix}{efe-appendix}
\part*{\textcolor{citecolor}{Appendix}}
\pdfbookmark[1]{Contents}{efe-appendix-contents}
\efeappendixcontents
\clearpage
\suppressfloats[t]

\section{Suites and Tasks}
\label{app:tasks}

\input{appendix_sections/a_tasks/a_tasks}

\section{Protocol Details}
\label{app:protocol}

\input{appendix_sections/b_protocol/b_protocol}

\section{Interface, Grading and Physics}
\label{app:framework}

\input{appendix_sections/e_framework/e_framework}

\section{Additional Results}
\label{app:results}

\input{appendix_sections/g_results/g_results}

\section{Further Analysis}
\label{app:analysis}

\input{appendix_sections/h_analysis/h_analysis}

\section{Example Attempts}
\label{app:runs}

\input{appendix_sections/c_runs/c_runs}

\section{Controlled Runs}
\label{app:ablation}

\input{appendix_sections/f_ablation/f_ablation}

\section{Failure Labels}
\label{app:failure}

\input{appendix_sections/d_failure/d_failure}

\section{Real-Robot Demonstration}
\label{app:real}

\input{appendix_sections/j_real/j_real}

\section{Limitations}
\label{app:limitations}

\input{appendix_sections/i_limitations/i_limitations}

\section{Extended Related Work}
\label{app:related}

\input{appendix_sections/k_related/k_related}

%% file: appendix_sections/a_tasks/a_tasks.tex
\paragraph{What a task contains.}
We package every task in the Harbor format with a fixed directory structure. The file \texttt{instruction.md} gives the agent its prompt: the goal, the submission requirements, the limits, and every convention the agent cannot infer, including coordinate frames, units, the action layout, how to use the robot client, and the reset budget. Any goal specified as an image, video, or other file is placed inside the agent's container and named in \texttt{instruction.md}. We store the task configuration in \texttt{task.toml}, which specifies the time limit, the CPU, memory, and disk caps of the agent's container, the verifier settings, and task metadata (the simulator, the robot, and whether a reference solution exists). The \texttt{environment/} directory defines the runtime via a \texttt{Dockerfile} for the agent's workspace container and a \texttt{docker-compose.yaml} that adds the remaining containers (the simulator service in \evalsetting{Standard} and \evalsetting{Realtime}, and a resource monitor) along with their resource caps. Within it, \texttt{environment/instance.json} defines the frozen scene (the objects, the robot, and their initial state), which the simulator checks against a stored hash at every build and reset so every episode starts from the same state. The \texttt{tests/} directory contains the verifier: its own \texttt{Dockerfile} and \texttt{docker-compose.yaml}, the entry script \texttt{test.sh}, and the grading code with the success rule, all running in an isolated container that the agent never sees. When a task has a reference solution, we include \texttt{solution/solve.sh}, which validates the task and is never shown to the agent, and we provide \texttt{README.md} to describe the task for human readers. Finally, we pin every base image to a fixed tag or digest so every run of a task uses the exact same software.

\Cref{tab:stats} gives the size of the exam, and \Cref{tab:suites} lists its 35 suites, of which 9 were built for the exam and 26 are adapted from existing benchmarks, whose scenes, robots and success rules we keep. The Robots column counts distinct robot models in the suite. The same robot can appear in several suites, so the total counts each of the 47 robots once. A test is one task in one setting, and the Core column counts the tasks in the core set. Time limits range from 30 minutes to four hours by suite. Of the 1,156 tasks, 590 allow one hour and 351 allow two hours or more.

\begin{table}[t]
\centering
\small
\caption{\looseness=-1 Size of the exam. Rows list the number of suites (with the number built for the exam in parentheses), robots, tasks, tests, tasks with a full or partial reference solution, core tasks and core tests.}
\label{tab:stats}
\begin{tabular}{lr}
\toprule
Item & Count \\
\midrule
Suites (new) & 35 (9) \\
Robots & 47 \\
Tasks & 1,156 \\
Tests & 2,288 \\
Tasks with a full / partial reference solution & 686 / 78 \\
Core tasks & 456 \\
Core tests & 901 \\
\bottomrule
\end{tabular}
\end{table}

\begin{table}[p]
  \centering
  \footnotesize
  \setlength{\tabcolsep}{3.5pt}
  \caption{The suites of the exam. The 9 suites that we built come first, followed by the 26 suites adapted from existing benchmarks, which \cref{app:tasks-adapted} cites. The Kind column gives the kinds of robot body in the suite, nine over the exam (mobile manip.\ = mobile manipulator; car = wheeled car). The Robots column counts distinct robot models, the Tests column counts tests over all tasks, and the Core column counts tasks in the core set.}
  \label{tab:suites}
  \begin{tabularx}{\linewidth}{@{}l>{\raggedright\arraybackslash}p{3.4cm}>{\raggedright\arraybackslash}Xrrrr@{}}
    \toprule
    Suite & Simulation backend & Kind & Robots & Tasks & Tests & Core \\
    \midrule
    DexHandExam & MuJoCo & hand & 3 & 30 & 60 & 23 \\
    Embodied Olympics - Fun & MuJoCo & humanoid & 2 & 79 & 179 & 56 \\
    Embodied Olympics - Match & MuJoCo & arm, humanoid, mobile manip. & 5 & 14 & 41 & 13 \\
    Embodied Safety Exam & MuJoCo & arm, two arms, mobile manip. & 4 & 72 & 76 & 50 \\
    iSpatialGym & OmniGibson (Isaac Sim) & mobile manip. & 1 & 12 & 24 & 8 \\
    RoboARC & MuJoCo & arm, two arms & 3 & 106 & 212 & 58 \\
    RoboDynamicExam & ManiSkill / SAPIEN & arm, two arms & 2 & 25 & 69 & 10 \\
    RoboEngineer & MuJoCo + learned dynamics & aerial, car, humanoid, quadruped & 5 & 30 & 30 & 15 \\
    RoboPaint & ManiSkill / SAPIEN & arm & 1 & 43 & 86 & 22 \\
    \midrule
    ABC Sim & MuJoCo & two arms & 1 & 29 & 60 & 0 \\
    AM-Bench & Isaac Lab / Isaac Sim & aerial & 1 & 9 & 17 & 2 \\
    AutoBio & MuJoCo & arm, two arms & 4 & 9 & 18 & 0 \\
    BEHAVIOR-1K & OmniGibson (Isaac Sim) & mobile manip. & 1 & 100 & 200 & 71 \\
    BiGym 2.0 & MuJoCo & humanoid & 1 & 13 & 39 & 10 \\
    DexJoCo & MuJoCo & hand & 2 & 11 & 22 & 2 \\
    DexToolBench & MuJoCo & hand & 1 & 21 & 42 & 20 \\
    EmbodiedSWE & Isaac Lab / Isaac Sim & arm, two arms, humanoid & 3 & 24 & 48 & 12 \\
    HumanoidBench & MuJoCo & humanoid & 2 & 10 & 30 & 8 \\
    KinDER & MuJoCo & mobile manip. & 1 & 5 & 10 & 4 \\
    LabUtopia & Isaac Sim & arm & 1 & 9 & 18 & 0 \\
    LIBERO-MAX & robosuite / MuJoCo & arm & 1 & 64 & 128 & 2 \\
    ManiGuard & OmniGibson (Isaac Sim) & arm & 1 & 30 & 60 & 7 \\
    MetaWorld+ & MuJoCo & arm & 1 & 50 & 100 & 2 \\
    MolmoSpaces & MuJoCo & arm, mobile manip. & 2 & 8 & 16 & 2 \\
    MuJoCo Playground & MuJoCo & arm, two arms, quadruped & 5 & 13 & 27 & 7 \\
    MyoSuite & MuJoCo & musculoskeletal & 6 & 22 & 45 & 13 \\
    RoboCasa & robosuite / MuJoCo & mobile manip. & 1 & 29 & 58 & 2 \\
    RoboCasa-GR1 & robosuite / MuJoCo & humanoid & 1 & 24 & 48 & 1 \\
    RoboCasa365 & robosuite / MuJoCo & mobile manip. & 1 & 50 & 101 & 5 \\
    RoboDojo & Isaac Lab / Isaac Sim & two arms & 1 & 54 & 102 & 12 \\
    RoboLab & Isaac Lab / Isaac Sim & arm & 1 & 53 & 106 & 6 \\
    RoboTwin 2.0 & SAPIEN & two arms & 1 & 47 & 94 & 3 \\
    RoboWits & Genesis & two arms & 1 & 15 & 30 & 6 \\
    VLABench & MuJoCo (dm\_control) & arm & 1 & 36 & 72 & 2 \\
    YAMLab & Isaac Lab / Isaac Sim & two arms & 1 & 10 & 20 & 2 \\
    \midrule
    Total & & & 47 & 1,156 & 2,288 & 456 \\
    \bottomrule
  \end{tabularx}
\end{table}

\subsection{New Suites}
\label{app:tasks-new}

RoboARC, DexHandExam and RoboPaint require contact-rich manipulation of deformable objects, multi-part assemblies and
tools. In RoboARC, single- and dual-arm robots untie knots, thread ropes through holes, disentangle rings from wire
puzzles, guide a loop along a wire without touching it, and assemble IKEA furniture. Earlier work studied such tasks with dedicated systems for untangling cables~\citep{viswanath2022untangling} and with datasets of furniture assembly manuals~\citep{wang2022ikeamanual}. In DexHandExam, dexterous hands turn book pages, open jars and insert a USB stick,
and objects move only through contact and gravity. In RoboPaint, a robot arm reproduces a painting or a calligraphy
character with a brush or a pen, so the task tests fine tool control. Dedicated robot painters such as FRIDA~\citep{schaldenbrand2022frida} plan their strokes in a differentiable simulation of the painting process.

RoboDynamicExam and the two Embodied Olympics suites involve dynamic objects and opponents, so the agent has to act
under tight timing constraints. These three suites contain 53 of the 86 tasks that have a \evalsetting{Realtime} test.
RoboDynamicExam asks single- and dual-arm robots to catch, strike and hand over moving objects such as tossed balls
and sliding pucks. In Embodied Olympics - Fun, a humanoid performs athletic skills in 79 task families, such as dribbling a
ball through a slalom, and in Embodied Olympics - Match, robots play full matches of table tennis, badminton, soccer,
foosball or fencing against scripted or learned opponents. Robots have learned to play table tennis~\citep{dambrosio2024tabletennis}, badminton~\citep{ma2025badminton} and soccer~\citep{haarnoja2023soccer} with reinforcement learning and task-specific training.

The remaining three suites test abilities beyond executing a stated goal. In iSpatialGym, a mobile bimanual robot
receives the goal only as an image of what its head camera should show when the task is done, and it must rearrange the
scene or reposition itself until its camera view matches this goal image. Multimodal models have been evaluated on spatial reasoning by answering questions about videos of indoor scenes~\citep{yang2025thinking}, and iSpatialGym requires the agent to act on such reasoning. Embodied Safety Exam pairs hazardous
requests, whose literal execution would cause harm, with benign requests of the same kind, so the agent must refuse
the hazardous requests and complete the benign ones, even when the hazard is hidden or emerges during the task. Its verifier decides success
from what physically happens in the simulation. RoboEngineer asks the agent to tune a controller on a nominal
simulator that it can access, deploy the controller to an evaluation simulator with undisclosed dynamics and refine it
from the deployment feedback. This loop resembles sim-to-real transfer, which prior work addresses with domain and dynamics randomization~\citep{tobin2017domain,peng2017dynamics} and by aligning simulated with real physics~\citep{he2025asap}. RoboEngineer's drone tasks build on the learned dynamics models of Neural Lander and Neural-Fly~\citep{shi2019neurallander,oconnell2022neuralfly}, and its racing-car tasks port the MuJoCo vehicle model of AnyCar~\citep{xiao2024anycar}, a wheeled car, without its learned dynamics model.

\subsection{Adapted Suites}
\label{app:tasks-adapted}

\begin{sloppypar}
\looseness=-1 The other 26 suites come from existing benchmarks, and we group them by their main
embodiment, as \Cref{fig:tasks} (right) also does. The single-arm suites are LIBERO-MAX~\citep{zhang2026liberomax},
RoboLab~\citep{yang2026robolab}, MetaWorld+~\citep{mclean2025metaworld}, VLABench~\citep{zhang2024vlabench},
ManiGuard~\citep{peng2026maniguard}, EmbodiedSWE~\citep{shen2026embodiedswe}, LabUtopia~\citep{li2025labutopia},
MolmoSpaces~\citep{kim2026molmospaces} and AutoBio~\citep{lan2025autobio}, and they range from
language-conditioned tabletop manipulation to laboratory automation. The bimanual suites are
RoboDojo~\citep{chen2026robodojo}, RoboTwin 2.0~\citep{chen2025robotwin}, ABC Sim~\citep{abcsim2026}, RoboWits~\citep{lin2026robowits}
and YAMLab~\citep{yamlab2026}, and the mobile manipulation suites are BEHAVIOR-1K~\citep{li2024behavior1k},
RoboCasa365~\citep{nasiriany2026robocasa365}, RoboCasa~\citep{nasiriany2024robocasa} and
KinDER~\citep{huang2026kinder}. The first three include long-horizon household activities in house and kitchen scenes, and KinDER tests physical reasoning.
RoboCasa-GR1~\citep{nvidia2025gr00t}, BiGym 2.0~\citep{zhang2026bigym} and
HumanoidBench~\citep{sferrazza2024humanoidbench} cover humanoid manipulation and whole-body control.
DexToolBench~\citep{kedia2026simtoolreal} and DexJoCo~\citep{wang2026dexjoco} cover dexterous manipulation and tool use.
MyoSuite~\citep{caggiano2022myosuite} covers musculoskeletal control, MuJoCo Playground~\citep{zakka2025mujoco} covers quadruped
locomotion together with a few arm tasks, and AM-Bench~\citep{wang2026ambench} covers aerial manipulation.
\end{sloppypar}

%% file: appendix_sections/b_protocol/b_protocol.tex
\paragraph{Budgets.} Each attempt runs in a container with 2 to 16 CPU cores and 4 to 39 GiB of memory, with a GPU where the simulator or the task needs one, with a wall-clock limit of 30 minutes to four hours, depending on the suite, and with a token budget that no attempt reaches. The harness can reach only the API of its model and the package registries, except in Embodied Olympics - Fun, which runs with open network access. The harness's web search is off in every task. A proxy between the harness and the model provider forwards only the requests to the model's API.

\paragraph{Reference solutions and integrity checks.} Every task must fail when the robot does nothing and pass with its reference solution wherever one exists (686 tasks have a full reference solution, and 78 have a partial one). Tasks without a reference solution rely on the original success rule and a human review. Reference solutions never enter the agent's container, and the solutions of the original benchmarks are removed from the container images. Before a suite is added, an agent is instructed to pass its tasks without solving them. If the agent succeeds on any task, the suite is not added until its verifier is fixed.

\paragraph{Leaderboard rules.} Each agent makes one attempt per test. An attempt that ends with an infrastructure or API error is rerun, and the latest finished attempt counts. A test counts as solved when the verifier reports success for its attempt. We report each setting over all of its core tests, including those of Embodied Olympics - Fun, whose tasks run with open network access, and of the suites with their own runtime, Embodied Olympics - Match and RoboEngineer, and \Cref{fig:benchmarks} gives the results by suite. The harness comparison uses the 456 core tests of the \evalsetting{Standard} Setting. When we pool the three strongest agents, as in \Cref{sec:results,app:results}, each test counts once per agent, and rates are reported per attempt.

\paragraph{Approach labels.} \looseness=-1 For \Cref{sec:f-control}, an agent read the logs of a random sample of 1,489 attempts and assigned each attempt the approach that produced its graded answer. \emph{Code to control} covers three kinds of program. An open-loop program computes the whole motion in advance, as waypoints, key frames, inverse-kinematics solutions or a recorded sequence, and runs it without feedback. A closed-loop program reads the state or the cameras at every control step or action chunk and adjusts its commands. A parameter search finds a passing motion, or its parameters, by trying many episodes and resets or by running an optimizer such as the cross-entropy method (CEM), CMA-ES, or a grid or random search. In \emph{Direct control}, the agent sends one action chunk at a time and decides on the next one after looking at images or readings. In \emph{Train a policy}, the agent first trains a control policy, such as a neural network or a reinforcement-learning policy, which then controls the robot. Attempts that switched approach midway or used two approaches equally are labeled mixed, and attempts that only explored or never ran anything are labeled as having no control plan. \Cref{fig:approaches} shows an example of each approach.

%% file: appendix_sections/e_framework/e_framework.tex
\looseness=-1 This section expands \Cref{sec:test} with the protocol rules and each suite's values and examples.

\subsection{Interface}
\label{app:fw-interface}

\paragraph{\evalsetting{Privileged} Setting: a simulator library in the agent's container.} The agent's container holds the simulator of the suite and a small Python library that the exam provides, which we call the tooling. The tooling rebuilds the frozen scene, resets it, advances it by applying actions, reads the state of every object and records each applied action as one row of an array. The agent may also call the planning primitives of the original benchmark, such as the grasp and place primitives of RoboTwin 2.0 with its CuRobo planner, or write its own planner, optimizer or learned controller. Whatever the agent calls, the tooling records the low-level action row applied at every control step, and the answer is this array of actions, saved as a trajectory file. For RoboTwin 2.0, for example, a row has 34 columns (16 joint position targets, 16 joint velocity targets and two gripper commands) and one row is applied per 4\,ms physics step. In the \evalsetting{Privileged} tests of Embodied Olympics - Match and EmbodiedSWE the answer is instead a program (one that plays the match, or one that controls the robot through the task), and the verifier runs it in a fresh simulator.

\paragraph{\evalsetting{Standard} Setting: the simulator as a service.} In the \evalsetting{Standard} and \evalsetting{Realtime} settings, the simulator runs as the simulator service in a separate container that the agent cannot access. The agent reaches it over HTTP through a Python client library that the exam provides. One call returns the specification of the robot: the action types with their layouts, units and limits; every joint with its position, velocity and effort limits and drive gains; the grippers and tool frames; the cameras with their intrinsics and mounting; the layout of the joint readings; the conventions of the world frame; and the budgets. A second call returns observations: either the current camera images, without advancing the simulation, or the images after the simulation has run for a requested number of seconds with the last command held. A third call executes an action chunk of $T$ rows, each of dimension $d$, and returns the video of the execution. Where the service offers it, a fourth computes forward and inverse kinematics on the model of the robot without advancing physics. A reset call exists only in tasks that grant resets. Every response contains the camera images (RGB, and depth on request), one row of joint readings per image (joint positions and velocities, gripper opening, tool pose and an estimate of the drive effort), the camera calibration, the simulated time and the time the agent has left. Object poses, segmentation, scene files, the seed and the success rule are never sent. Contact forces on objects are not sent either, except as safety telemetry that a suite documents field by field, as in RoboARC, and that never gives an object's pose. Any other request, such as a query for object states or for success, returns an error. The images come back as binary data that the agent's code can store and process, so they do not have to pass through the model's context.

\paragraph{From a chunk to motion.} Joint position rows are $dt$ seconds apart, where $dt$ is a multiple of the control period and at most 0.5\,s. The service rejects a value of $dt$ that breaks this rule and does not round it. The service streams the rows to the drives of the joints in the way a position controller on a real robot follows a trajectory, either interpolating between rows or holding each row until the next. It clips any value outside the joint limits and reports each clipped entry in the response. End-effector rows go through the inverse kinematics or the motion planner of the suite. Where the original benchmark controls the robot through a higher-level controller, the agent sends the commands of that controller, as in RoboCasa and in the whole-body controller of BiGym 2.0. Each suite keeps its own control rate and physics step. Control rates range from 5\,Hz (the mobile robot in MolmoSpaces) to 500\,Hz (AutoBio, where one row is one 2\,ms physics step), and one control step runs up to 100 physics steps (VLABench). The success rule is checked after every action chunk and every observation request that advances physics, and in most suites the first success ends the episode. In the \evalsetting{Standard} tests of RoboDojo, a success does not end the episode, so that the end of the episode does not tell the agent that it has succeeded, which would give it a success check to try against until it passes. The episode ends at the task's step limit, when the agent ends it, at its time limit, or when a check of the task fails, and it passes if the success rule held at any point, which matches RoboDojo's own success within its step limit.

\paragraph{\evalsetting{Realtime} Setting: a simulation that keeps running.} The \evalsetting{Realtime} Setting uses the same calls. From the first request that advances physics, the simulation advances one control step every $S$ control periods of wall-clock time, whatever the agent does, where $S \geq 1$ is a slowdown factor that each task states. A submitted chunk is answered at once and replaces the queued actions starting at a control step that the agent specifies. When the queue is empty, the robot holds its last target. The agent can therefore run a program in its own container that observes the robot and sends new chunks in a loop. By default, $S$ is 1 when the simulator runs at least three times faster than real time, and larger otherwise, so that the service can run each step on time. The service records the steps it could not run on time, and the time the robot stood still because of them.

\paragraph{Suites with their own runtime.} RoboEngineer and Embodied Olympics - Match use their own runtime instead of the simulator service, in their \evalsetting{Standard} tests and in the \evalsetting{Realtime} tests of Embodied Olympics - Match. In RoboEngineer, the agent tunes a controller in a simulator that the agent can access and deploys the controller to a second simulator, which the agent cannot access and whose dynamics differ from those of the first in ways not disclosed to the agent. The verifier then scores the controller that the agent selects on episodes that the agent has not seen.

\paragraph{Step limits and time budgets.} The protocol adds no step limit to the \evalsetting{Standard} and \evalsetting{Realtime} settings. RoboDojo keeps its own step limit (\Cref{fig:limits}), and many other suites cap the episode length, which the instruction states. A \evalsetting{Privileged} trajectory may have at most a stated number of rows. The main limit is the agent's wall-clock time, from 30 minutes to four hours depending on the suite, and it is the same in every setting of a task. Each response reports the time left, counted from the first request of the agent. One call may return only a stated number of images and may run for at most 15 minutes by default, so that one call cannot exhaust the memory of the service or occupy it indefinitely.

\paragraph{Limits.} The limits of a test are the agent's wall-clock time and token budget, the resets that the task grants, the slowdown $S$ in the \evalsetting{Realtime} Setting, the per-call limits above, and the CPU, memory, process, disk and GPU memory caps of each container, which a resource monitor records and enforces where the container runtime does not. \evalsetting{Privileged} tests allow unlimited resets, and \evalsetting{Standard} and \evalsetting{Realtime} tests allow no resets unless the task grants a budget, which its instruction states (\Cref{sec:limits}). These values are part of the configuration of a result, and results are compared only between runs with the same configuration.

\subsection{Grading}
\label{app:fw-grading}

\paragraph{Success rules.} Each task is graded by one success rule, a condition on the simulator state. Adapted suites keep the success rule of the original benchmark and tighten it in some tasks. Examples are the success function of a RoboTwin 2.0 task, the BDDL goal of a BEHAVIOR-1K activity evaluated on OmniGibson object states, and the solved flag of MyoSuite. For the suites that we built, we wrote our own rules on the simulator state. The success rule of Embodied Olympics - Fun checks the suite's scoring checkpoints, legal contacts and fouls after every 1\,ms physics step, RoboDynamicExam checks ordered phases (a catch, a hold, a crossing, a release) after every 1\,ms physics step, and DexHandExam computes its checkpoints from contact pairs and contact forces. The instruction states each condition of the success rule with the same thresholds that the rule uses.

\paragraph{Settle and hold windows.} \looseness=-1 Some rules are checked over a time window. A hold window requires the goal to hold for a stated time, so that an object in flight or a grasp that is about to slip does not count. In BEHAVIOR-1K the goal must hold for 1\,s with the object released and at rest. MyoSuite requires its solved flag for 1\,s (5\,s for the Baoding balls), BiGym 2.0 requires its goal for 1\,s and no fall during the episode, and ManiGuard requires its goal for 10 consecutive control steps and no safety violation during the episode. A settle window keeps the simulation running with the last command held while the rule is checked, as in RoboLab, which holds the last command for 2\,s. Rules without a window are checked on a single state. A window is part of the rule of the task and applies in every setting. It is checked on the live simulator in the \evalsetting{Standard} and \evalsetting{Realtime} settings and after the last row of the replay in the \evalsetting{Privileged} Setting.

\paragraph{\evalsetting{Privileged} Setting: a fresh replay.} The verifier rebuilds the frozen scene in a fresh process, checks that its initial state matches the stored hash, replays the submitted actions open loop and applies the success rule to the replay. By default it replays once. In BEHAVIOR-1K, AM-Bench, RoboDojo and RoboLab, whose simulators do not replay bit for bit, it replays up to three times, each in a fresh process, and the attempt passes if any replay succeeds. Where the answer is a program, as in Embodied Olympics - Match and EmbodiedSWE, the verifier runs the program in a fresh simulator instead. Because only the actions are replayed, a change that the agent made directly to the state of its own simulator, such as moving an object directly, has no effect on the result. A file that is not a valid trajectory fails.

\paragraph{\evalsetting{Standard} and \evalsetting{Realtime} settings: the live episode.} \looseness=-1 The simulator service checks the success rule on the episode that it ran, and the verifier reports the result from the service's record of the episode. The service checks the hash of the start state when the episode begins and after every reset. When the agent's time is up, the program that runs the evaluation stops every process in the agent's container before it signals the service. From then on the service refuses every request that would advance the simulator, runs any settle window, checks the success rule on the live simulator and writes the result with the recorded trajectory. The verifier reports that result, and nothing is replayed to decide it. The result depends only on the state of the simulator, never on what the agent reports. An episode in which the agent's actions cause a simulator error fails. A failure of the infrastructure, such as a scene that does not rebuild to its stored start state or a service that crashed, gives no result, and the whole attempt can be rerun. With resets, only the last episode is graded. We also record whether the first episode, before any reset, succeeded, and report this as first-episode success.

\paragraph{Isolation.} \looseness=-1 The verifier runs in its own container, built from the task's grading files, without network, and receives only the files that the task declares for grading: the trajectory file (or the submitted program) in the \evalsetting{Privileged} Setting and the record of the simulator service, which checks the success rule, in the \evalsetting{Standard} and \evalsetting{Realtime} settings. It first removes from the agent's output directory every entry that is not a regular file, such as a link. No writable volume is shared between the agent's container and the containers of the simulator service and the verifier, and the simulator service opens no port on the host. The agent can never read the reference solution or the files of the verifier, and in the \evalsetting{Standard} and \evalsetting{Realtime} settings it cannot read the code of the success rule either. A program submitted as the answer runs under a user account without administrator rights and without network, cannot read the files of the simulator or the verifier, and every process it starts ends before the result is written. The result is written before the video of the attempt is rendered, and the video never changes it.

\subsection{Physics}
\label{app:fw-physics}

\paragraph{Simulators.} Each suite keeps the simulator of its original benchmark, and new suites use the simulator that suits their physics. MuJoCo~\citep{todorov2012mujoco} runs most suites, directly or through robosuite, dm\_control and MyoSuite, in versions 3.2 to 3.14 and mostly on the CPU. NVIDIA Isaac Sim and Isaac Lab with PhysX run RoboLab, RoboDojo, YAMLab, AM-Bench, LabUtopia and EmbodiedSWE, and OmniGibson on Isaac Sim runs BEHAVIOR-1K, ManiGuard and iSpatialGym. SAPIEN~3 with PhysX runs RoboTwin 2.0, and through ManiSkill~3 it also runs RoboPaint and RoboDynamicExam. Genesis runs RoboWits. RoboEngineer runs MuJoCo, and its drone tasks add learned dynamics models, a learned ground-effect model of a quadrotor and a wind-force model fitted to published flight logs.

\paragraph{What is simulated.} \looseness=-1 All suites simulate rigid bodies with frictional contact, joint limits and actuator limits. Several also simulate deformable and particle matter: cloth in the blanket tasks of RoboARC (a deterministic GPU solver) and the folding tasks of RoboDojo (GPU PhysX), ropes and cords as soft bodies in RoboARC, liquids as GPU PhysX particles in RoboDojo, and the object states of OmniGibson in BEHAVIOR-1K, where objects can be sliced, cooked or filled with particles. MyoSuite moves its limbs with simulated muscles instead of motors, and RoboPaint models brushes and pens on paper. Physics steps range from 1\,ms (Embodied Olympics - Fun and RoboDynamicExam) to 1/60\,s (LabUtopia), with 2\,ms in DexHandExam, DexJoCo and AutoBio and 1/120\,s in iSpatialGym, ManiGuard, AM-Bench and YAMLab.

\paragraph{Determinism.} \evalsetting{Privileged} grading replays the submitted actions, so when we build a task we check whether fresh replays of a trajectory end in the same final state on the hardware used for evaluation. Where they do not, the verifier replays up to three times (\Cref{app:fw-grading}). The frozen scene is also rebuilt and checked against its stored start-state hash before every episode. Where a task does not replay reliably at all, as with the GPU cloth and liquids of RoboDojo, the task is offered only in the \evalsetting{Standard} Setting, where the live episode is graded, and its video is a recording of the live episode rather than a replay.

\paragraph{Realism rules.} \looseness=-1 The robot moves only through its own actuators and the controllers stated in the task. In the \evalsetting{Standard} and \evalsetting{Realtime} settings the agent has no way to set the state of the simulator, and in the \evalsetting{Privileged} Setting only actions are replayed, so moving an object directly has no effect on the grade in any setting. The suites we built for the exam add stricter rules. In Embodied Olympics - Fun, the simulator never attaches a held object rigidly to the hand, never pins a body or moves it with a motion-capture constraint, and never sets a velocity directly after the start. Balls come from ball machines and goalkeepers are force-limited mechanisms. Each robot uses its own hands and actuators, modeled on measured actuator data from the manufacturer (torque-speed curves, rotor inertia and friction). Where we increased a parameter above the manufacturer's value, the increase is at most a factor of 1.5 to 2 and is stated in the task; the one stated exception is the grip on the high bar. DexHandExam has no rigid attachments or scripted object motion, so objects move only through contact and gravity, and in the furniture assembly tasks of RoboARC, the robot holds parts only by finger friction. Where a suite keeps a grasp aid, its documentation states it. In the knot, ring and rope tasks of RoboARC, a rigid attachment holds a grasped object and slips when the pull exceeds 60 to 65\,N for 20\,ms. Adapted suites keep the grasp model of the original benchmark, such as the assisted grasping of OmniGibson and the hand of MetaWorld+, which a motion-capture constraint moves.

%% file: appendix_sections/g_results/g_results.tex
This section describes the agents, the counting rules and the data behind \Cref{sec:experiments}. It then shows the success rates of \Cref{fig:leaderboard} as bars and gives further results by suite, by cost and by task type (\Cref{sec:results}).

\paragraph{Agents.} \looseness=-1 We evaluate eight agents, each a model running in a harness at high reasoning effort. The three strongest are GPT-6 Astra~\citep{openai2026gpt6astra} in Codex~\citep{openai2025codex}, Claude Opus 5.5~\citep{anthropic2026opus55} in Claude Code~\citep{anthropic2025claudecode} and Gemini 4 Argon in Antigravity CLI. The other five are Kimi 3.1 Pro and Qwen3.8 Max in Claude Code and three Flash-class models: Gemini 3.8 Flash in Antigravity CLI, GLM 5.3 Flash in Claude Code and DeepSeek V4.1 Flash in DeepSeek Harness (dsh), DeepSeek's own harness. We report each result for a model in a particular harness, because the way the harness exposes tools, manages the context and ends a session is part of what we measure. Switching harnesses can change a model's success rate by up to 12 percentage points (\Cref{sec:f-agents}).

\paragraph{Counting rules.} \Cref{app:protocol} gives the leaderboard rules. Each column of \Cref{fig:leaderboard} covers all core tests of one setting (456 \evalsetting{Standard}, 391 \evalsetting{Privileged} and 54 \evalsetting{Realtime} tests), and every core task has a \evalsetting{Standard} test. When we pool the three strongest agents, each test counts once per agent. For results by suite, we treat the IKEA assembly tasks of RoboARC as a separate suite.

\paragraph{Data.} Success rates are the results of the public leaderboard. The controlled runs add features of the \evalsetting{Privileged} Setting to the \evalsetting{Standard} Setting or run a model in a different harness.

\begin{figure}[t]
  \centering
  \includegraphics[width=\linewidth]{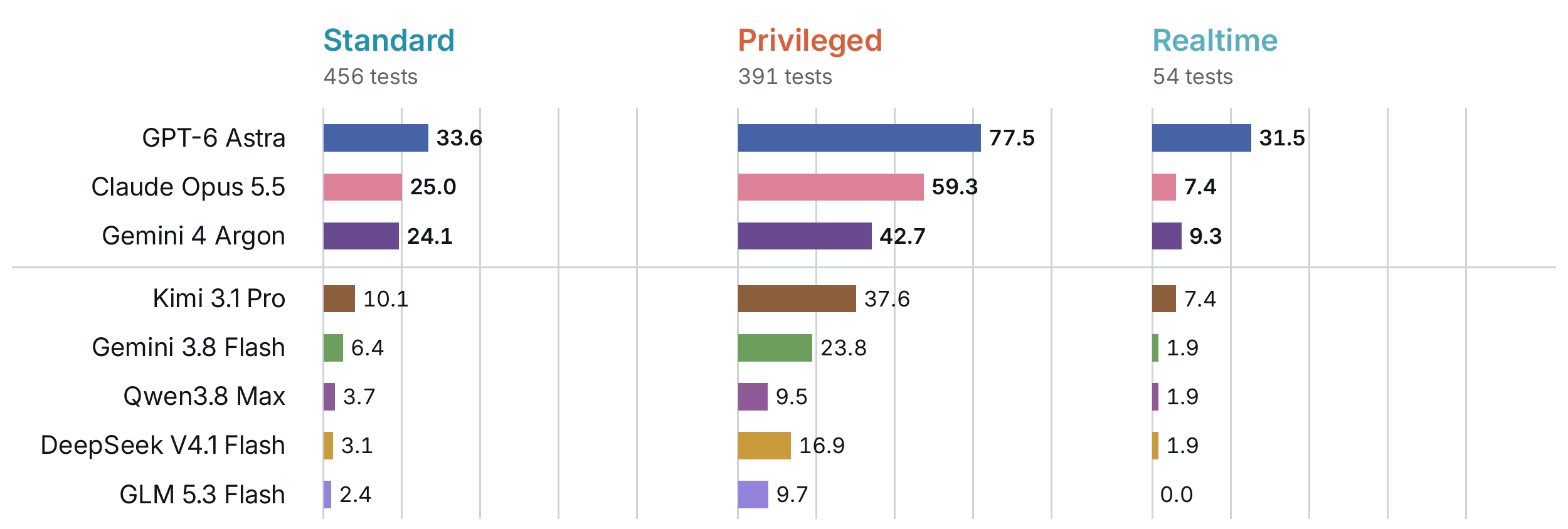}
  \caption{Success rate of each agent on the core set. Each bar gives the share of the core tests of one setting that one agent solves.}
  \label{fig:results}
\end{figure}

\paragraph{Success rate of each agent.} \Cref{fig:results} shows the success rates of \Cref{fig:leaderboard} as bars, for each setting. GPT-6 Astra leads in every column and solves a third of the \evalsetting{Standard} tests. The three strongest agents keep the same order in every column except the \evalsetting{Realtime} column, where Gemini 4 Argon is ahead of Claude Opus 5.5. Outside the core set, success rates are higher.

\paragraph{Paired outcomes.} \looseness=-1 \Cref{fig:transfer} in \Cref{sec:f-settings} splits the 391 core tasks that have both a \evalsetting{Privileged} and a \evalsetting{Standard} test by the settings in which each agent solves them. GPT-6 Astra solves 4 tasks in the \evalsetting{Standard} Setting but not in the \evalsetting{Privileged} Setting, and Gemini 4 Argon solves 18 such tasks. The \evalsetting{Privileged} result is therefore an approximate upper bound on the \evalsetting{Standard} result over a set of tasks, not on each single task (\Cref{sec:settings}).

\paragraph{Results by suite.} GPT-6 Astra leads or ties in most suites, while Claude Opus 5.5 and Gemini 4 Argon each lead in some (\Cref{fig:benchmarks}). We count only suites with at least five tests in the setting. In the \evalsetting{Standard} Setting, at least half of the tests are solved by none of the three strongest agents in 14 of 22 suites, among them DexToolBench, the IKEA assembly tasks, HumanoidBench and BEHAVIOR-1K. The three strongest agents solve at least 60\% of their attempts only in ManiGuard, whose safety rules are stated in the instruction. In the \evalsetting{Privileged} Setting, at least half of the tests are unsolved only in HumanoidBench, and the three strongest agents solve at least 70\% of their attempts in 10 of 20 suites: Embodied Olympics - Match, EmbodiedSWE, RoboDynamicExam, RoboCasa365, RoboWits, MuJoCo Playground, Embodied Olympics - Fun, BiGym 2.0, RoboPaint and ManiGuard.

\begin{figure}[t]
  \centering
  \includegraphics[width=\linewidth]{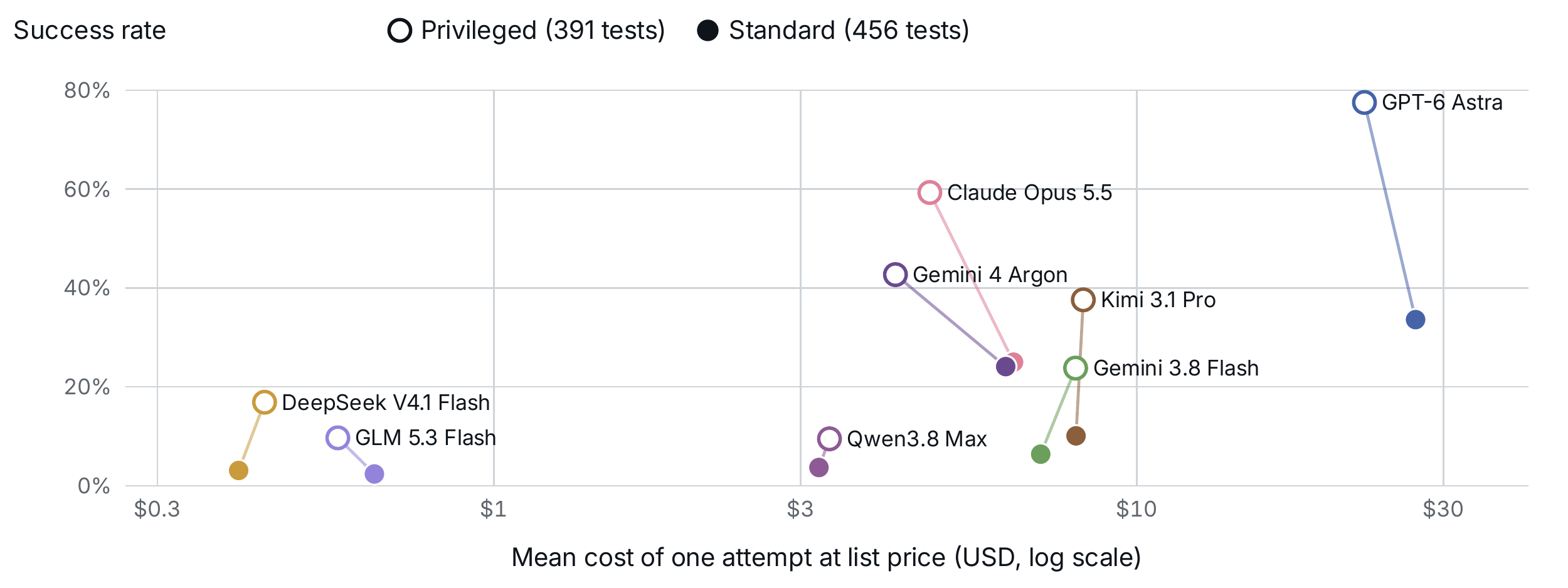}
  \caption{Success rate of each agent against the mean list-price cost of its attempts.}
  \label{fig:cost}
\end{figure}

\paragraph{Cost of an attempt.} \Cref{fig:cost} prices each attempt from the tokens that its harness logged, at the providers' list prices of November 2026. These are list-price costs, not billed costs. Each graded attempt is priced from the run that the leaderboard counts for it. In the \evalsetting{Standard} Setting, an attempt of GPT-6 Astra costs \$27.1 on average, about four times as much as one of Claude Opus 5.5 or Gemini 4 Argon. Kimi 3.1 Pro and Gemini 3.8 Flash spend more per attempt than these two agents and solve less than half as many \evalsetting{Standard} tests. For every agent, in the \evalsetting{Privileged} and in the \evalsetting{Standard} Setting, the median failed attempt costs more than the median solved attempt.

\begin{figure}[t]
  \centering
  \includegraphics[width=\linewidth]{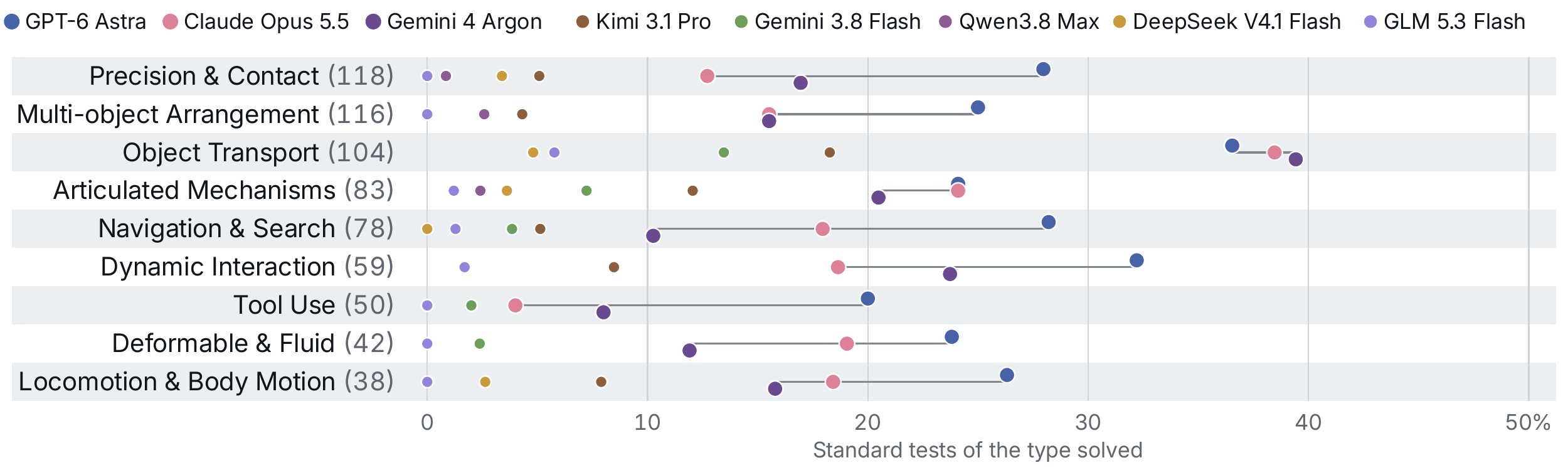}
  \caption{Share of the core tasks of each type whose \evalsetting{Standard} test each agent solves, with a gray line joining the three strongest agents. The number of core tasks of the type is in parentheses, and a task can have several types.}
  \label{fig:tasktype}
\end{figure}

\paragraph{Results by task type.} \looseness=-1 A task type describes a core task as a whole, whereas the stage-level failure labels of \Cref{sec:labels} assign a type to each stage of a task. By task type (\Cref{fig:tasktype}), GPT-6 Astra's lead on the \evalsetting{Standard} tests is widest in the Tool Use and Precision \& Contact types, and Gemini 4 Argon is ahead of it in Object Transport, at 39\%. For the three strongest agents together, Object Transport has the highest success rate of the nine task types, at 38\% of their attempts, and Tool Use has the lowest, at 11\%.

%% file: appendix_sections/h_analysis/h_analysis.tex
This section gives the full analyses behind the key takeaways of \Cref{sec:f-settings,sec:f-difficulty,sec:f-agents,sec:f-control}, in the same order. The definitions and statistics that these analyses use are in \Cref{app:ablation,app:failure}.

\subsection{Causes of the Drop from the \evalsetting{Privileged} to the \evalsetting{Standard} Setting}
\label{app:an-settings}

To measure how much two features that the \evalsetting{Standard} Setting lacks contribute to the drop, we ran GPT-6 Astra in Codex on the default tests of the core tasks with one or both of them added to the \evalsetting{Standard} interface (\Cref{fig:ablation-ladder}, \Cref{app:ablation}). The first feature is a query that returns the state of every object, including its pose, size and velocity. The second is unlimited resets together with calls that save the simulator state and restore it later. On the 391 core tasks that have both settings, GPT-6 Astra solves 28\% with the unmodified interface. The state query raises this rate to 45\%, and unlimited resets raise it to 53\%. With both features, the agent solves 68\%, against 77\% on their \evalsetting{Privileged} tests. \textbf{With unlimited resets, GPT-6 Astra uses saved states to retry a hard stage without repeating the whole task.} Of the 206 successes of the reset run on these tasks, 14\% came in the first episode, and 66\% came in an episode that started from a saved state.

The stage labels show at which kinds of stages the \evalsetting{Standard} Setting raises the failure rate (\Cref{fig:fa-transfer-a}). The failure rate of a stage type is the number of its stages at which attempts fail, divided by the number of its stages that attempts reach. We compare it between the two settings within each suite and average over the suites (\Cref{tab:fa-step-rates}). \Cref{app:fa-modes} defines the precise and coarse groups of stage types and gives the statistical tests. Among the stage types compared in at least two suites, the failure rate rises by 24 to 46 percentage points for precise types and by 1 to 20 points for coarse types. The failure rate of cleaning, which is in neither group, also rises by 51 points. Pooled within suites, the failure rate rises from 22\% to 56\% at precise stages and from 22\% to 38\% at coarse stages. \Cref{fig:case-fan-frames} shows an example, and \Cref{app:fa-cases} gives it in full. The difference between the two groups is smaller, but still significant, when the same agent reaches the same stage of a task in both settings. There, the rise is 25 points at precise stages and 15 points at coarse stages. At follow-path stages, most of them waypoints of DexToolBench, the rate rises by 11 points in this comparison. Without these stages, the rise at precise stages is 33 points.

\looseness=-1 Combining the controlled runs with the stage labels shows which feature helps at which kind of stage (\Cref{fig:ablation-stages}, \Cref{tab:fa-ablation-stages}). We take the \evalsetting{Standard} tests that GPT-6 Astra failed in the baseline run of the controlled runs and group them by the type of the failure stage. Of these failures, 91 are at precise stages and 97 are at coarse stages. Some of these failures would turn into successes if the test were simply run again. We therefore compare each run with the run without the success signal, a further controlled run in which the interface no longer reports when an episode succeeds. This run repeats the same tests and changes only the success signal, so we use it as a rough approximation of a second attempt. Its success rate is within 4 points of the baseline's (\Cref{tab:ablation}), but this run does not separate the variation between attempts from the effect of hiding the success signal. At precise stages, the state query solves 21\% of the failures, against 19\% for the run without the success signal. Unlimited resets solve 36\%, and 31 of these 33 successes came after a reset or a restored state. The agent did call the state query in the runs that offer it. \textbf{At precise stages, adding the state query to the resets raises the share solved from 36\% to 47\%.} At coarse stages, the state query alone solves 34\% of the failures, against 14\% for the run without the success signal, and the two features together solve 67\%, against 60\% in the \evalsetting{Privileged} Setting. With both features, the agent solves 47\% of the precise failures, compared with 73\% in the \evalsetting{Privileged} Setting, and most of this remaining gap lies at follow-path, thread and assemble stages. Neither added feature lets the agent advance the physics one step at a time or call planning primitives, as the \evalsetting{Privileged} Setting does, and these abilities may account for part of the remaining gap.

The comparisons in this subsection have limits. The controlled runs use one agent. The labels cannot separate errors of perception from errors of control (\Cref{app:fa-stats}).

\subsection{Where Agents Fail}
\label{app:an-difficulty}

\looseness=-1 Failures on a task tend to happen at the same stage (\Cref{fig:fa-where-a}). We measure this by the concentration of failure stages, which is the probability that two of the failed attempts on a test, chosen at random, failed at the same stage. Each agent makes one attempt per test, so the two attempts always come from different agents, and the measure pools the agents that failed the test. On the \evalsetting{Standard} tests that at least two agents failed at located stages, on tasks with at least two stages, the concentration averages 61\%, and we compare it with two baselines. In the first baseline, each agent fails at a stage chosen at random, with the probability of each stage position set by how often failures occur at that position on the other tasks of the same setting, so early stages are more likely. This baseline gives 37\%. In the second baseline, an agent fails at each stage with the failure rate of the stage's type on the other tasks, and this baseline gives 48\% (\Cref{tab:fa-common-stage}). For the three strongest agents alone, the concentration is 56\%, against 31\% for the first baseline built from their failures on the other tasks (\Cref{app:fa-stats}). \textbf{Failures therefore concentrate more than the stage types predict.} The most common failure stage is usually the first stage of the task. Of the 264 \evalsetting{Standard} tests whose failures have a single most common stage, this stage is the first stage in 65\%, and it is the first stage in 56\% of those whose tasks have at least two types of stages. In a RoboDynamicExam task, for example, seven of the eight \evalsetting{Standard} attempts caught a thrown ball in a cup and then failed at the recatch, in three different ways (\Cref{fig:case-ball}, with all eight attempts in \Cref{app:fa-cases}).

Failure rates differ between stage categories (\Cref{fig:fa-taxonomy-a}). Pooled over the \evalsetting{Privileged} and \evalsetting{Standard} settings, attempts fail at 66\% of the material stages (such as wiping or pouring) that they reach, at 45\% of the body stages (such as walking or keeping balance) and at 20\% of the mechanism stages. Within a category, the rate also depends on how the robot must act. Opening a door or a lid fails in 38\% of cases, whereas closing one fails in 19\% when it does not already hold at the start (\Cref{app:fa-stats}), and pressing fails in 3\% (\Cref{tab:fa-pooled-rates}). Relation stages, which place or fit one object relative to another (\Cref{tab:fa-steps}), fail more often when two arms act together or the legs carry the robot's weight than when one hand acts (\Cref{fig:fa-taxonomy-b} in \Cref{app:fa-stats}). \textbf{However, which stage types fail most differs from suite to suite.} When we rank each pair of stage types by their failure rates on all suites but one, the ranking holds in the held-out suite for 55\% of the 313 pairs, close to what a random ranking would give. Whether an attempt fails at a stage that it reaches is also predicted better by the identity of the agent than by the stage type (\Cref{app:fa-stats}). Each category occurs mostly in particular suites and with particular robots, so these rates do not show that the category itself causes the failures.

The most common failure stage of a task is also usually the same in both settings. Of the tasks whose failures have a single most common stage in both settings, 81\% have the same such stage in the two settings, compared with 49\% if each setting's failure stages were sampled independently. Excluding the tasks whose most common failure stage is the first stage in both settings, the share is still 57\%, compared with 9\% for random failure stages (\Cref{app:fa-stats}). \textbf{Of the attempts that get past the stage at which the other agents on the same test most often failed, 64\% pass the test in the \evalsetting{Privileged} Setting and 33\% in the \evalsetting{Standard} Setting.} Part of this link between getting past the stage and passing follows from the order of the stages, because an attempt that gets past the common stage is already closer to the end of the task.

\subsection{Differences between Agents}
\label{app:an-agents}

\paragraph{Tests solved by one agent and by another.} \looseness=-1 For two agents A and B, the cell in row A and column B of \Cref{fig:solve-matrix} gives the share of the tests that A solves that B also solves. The denominator of row A is the number of tests that A solves. It is given in parentheses after the name of A, and it equals A's count of solved tests in the \evalsetting{Standard} column of \Cref{fig:leaderboard}. The diagonal is left empty, because it would always be 100\%. The last column gives the share of A's solved tests that at least one agent above A in \Cref{fig:leaderboard} also solves. A test breaks the order when some agent solves it and GPT-6 Astra fails it. We count such a test under the highest agent in \Cref{fig:leaderboard} that solves it, so every agent above this agent fails the test. We use the \evalsetting{Standard} Setting in the main text because every core task has a \evalsetting{Standard} test and all agents reach the robot through the same interface there. Each of the five weaker agents solves 11 to 46 \evalsetting{Standard} tests. A single test therefore changes a share in its row by 2 to 9 percentage points. Among these five agents, the order of \Cref{fig:leaderboard} does not hold. Gemini 3.8 Flash, for example, solves 2 of the 11 \evalsetting{Standard} tests that GLM 5.3 Flash solves.

\begin{figure}[h]
  \centering
  \includegraphics[width=0.625\linewidth]{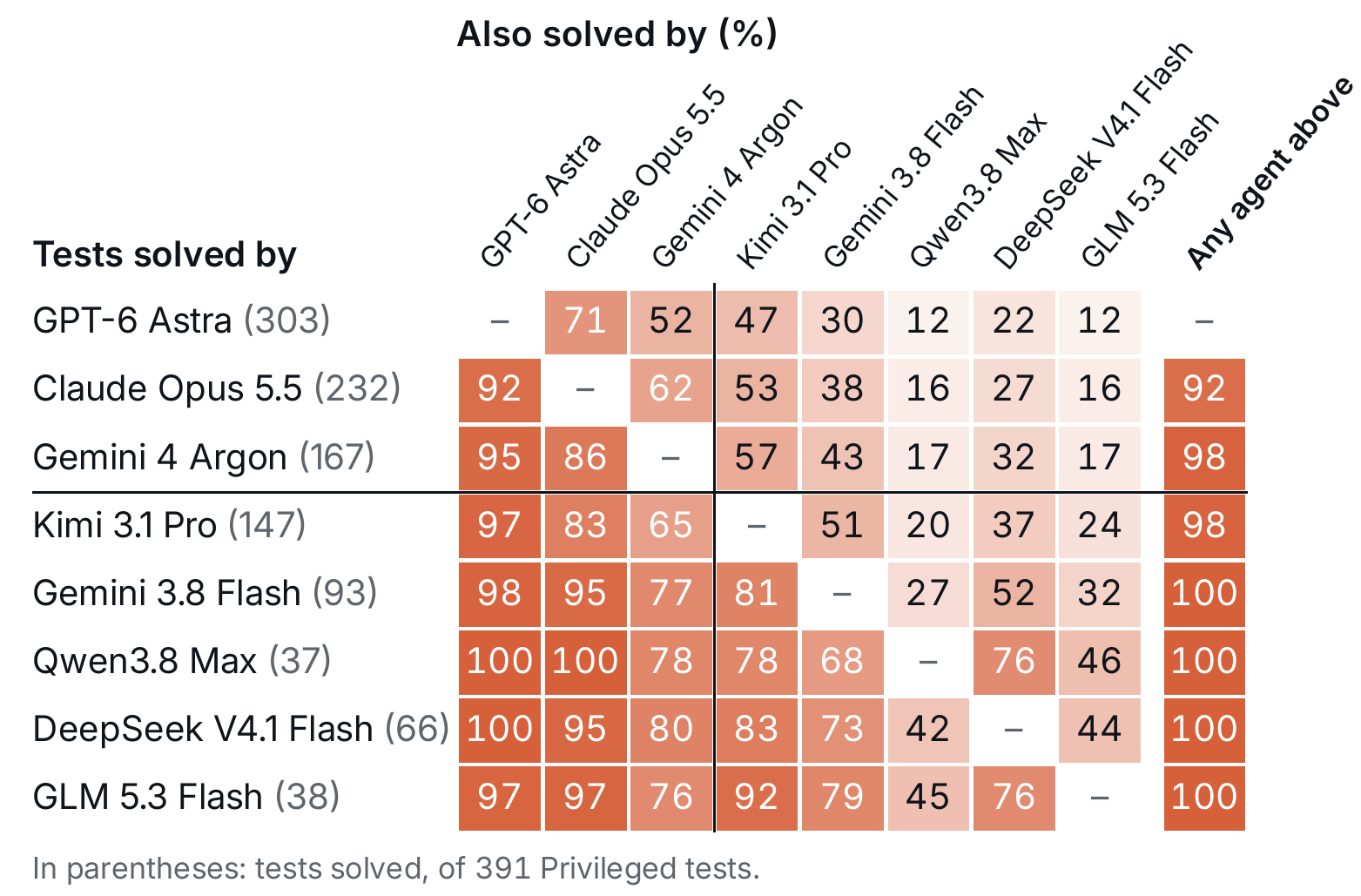}
  \caption{Share of the \evalsetting{Privileged} tests solved by the row agent that the column agent also solves, as in \Cref{fig:solve-matrix}, with the agents in the order of \Cref{fig:leaderboard}.}
  \label{fig:solve-matrix-priv}
\end{figure}

\looseness=-1 In the \evalsetting{Privileged} Setting, the order of \Cref{fig:leaderboard} holds more closely. GPT-6 Astra solves more than 90\% of the \evalsetting{Privileged} tests that Claude Opus 5.5 or Gemini 4 Argon solves (\Cref{fig:solve-matrix-priv}), and only 24 of the 327 \evalsetting{Privileged} tests that some agent solves break the order (\Cref{tab:counter}), against 68 of 221 in the \evalsetting{Standard} Setting. Every agent except GPT-6 Astra solves at most 5 \evalsetting{Realtime} tests, so we do not draw a \evalsetting{Realtime} matrix. Of the 20 \evalsetting{Realtime} tests that some agent solves, 3 break the order: Claude Opus 5.5 solves 2 of them and Gemini 4 Argon 1.

\begin{table}[h]
  \centering
  \small
  \caption{Tests that break the order of \Cref{fig:leaderboard}, by suite and setting. ``Tests'' gives the number of tests of the suite in the setting. Each test that breaks the order is counted in the column of the highest agent in \Cref{fig:leaderboard} that solves it, and ``Weaker'' sums the five weaker agents. The six suites with the most such \evalsetting{Standard} tests and BEHAVIOR-1K, which has the most such \evalsetting{Privileged} tests, are listed, and ``Other suites'' sums the rest. RoboARC excludes its IKEA assembly tasks, which are counted under Other suites.}
  \label{tab:counter}
  \setlength{\tabcolsep}{3.5pt}
  \begin{tabular}{@{}lrrrrrrrr@{}}
    \toprule
     & \multicolumn{4}{c}{\evalsetting{Standard}} & \multicolumn{4}{c}{\evalsetting{Privileged}} \\
    \cmidrule(lr){2-5}\cmidrule(l){6-9}
    Suite & Tests & \begin{tabular}[b]{@{}c@{}}Claude\\Opus 5.5\end{tabular} & \begin{tabular}[b]{@{}c@{}}Gemini 4\\Argon\end{tabular} & Weaker & Tests & \begin{tabular}[b]{@{}c@{}}Claude\\Opus 5.5\end{tabular} & \begin{tabular}[b]{@{}c@{}}Gemini 4\\Argon\end{tabular} & Weaker \\
    \midrule
    Embodied Safety Exam & 50 & 8 & 5 & 0 & -- & -- & -- & -- \\
    RoboARC & 50 & 8 & 2 & 1 & 50 & 5 & 1 & 0 \\
    Embodied Olympics - Fun & 56 & 5 & 1 & 0 & 56 & 1 & 0 & 0 \\
    ManiGuard & 7 & 4 & 0 & 0 & 7 & 0 & 0 & 0 \\
    RoboDojo & 12 & 0 & 2 & 1 & 12 & 0 & 0 & 0 \\
    Embodied Olympics - Match & 13 & 1 & 1 & 1 & 13 & 2 & 0 & 0 \\
    BEHAVIOR-1K & 71 & 1 & 1 & 0 & 71 & 6 & 1 & 1 \\
    Other suites & 197 & 20 & 5 & 1 & 182 & 4 & 1 & 2 \\
    \midrule
    All & 456 & 47 & 17 & 4 & 391 & 18 & 3 & 3 \\
    \bottomrule
  \end{tabular}
\end{table}

\begin{figure}[h]
  \centering
  \includegraphics[width=0.67\linewidth]{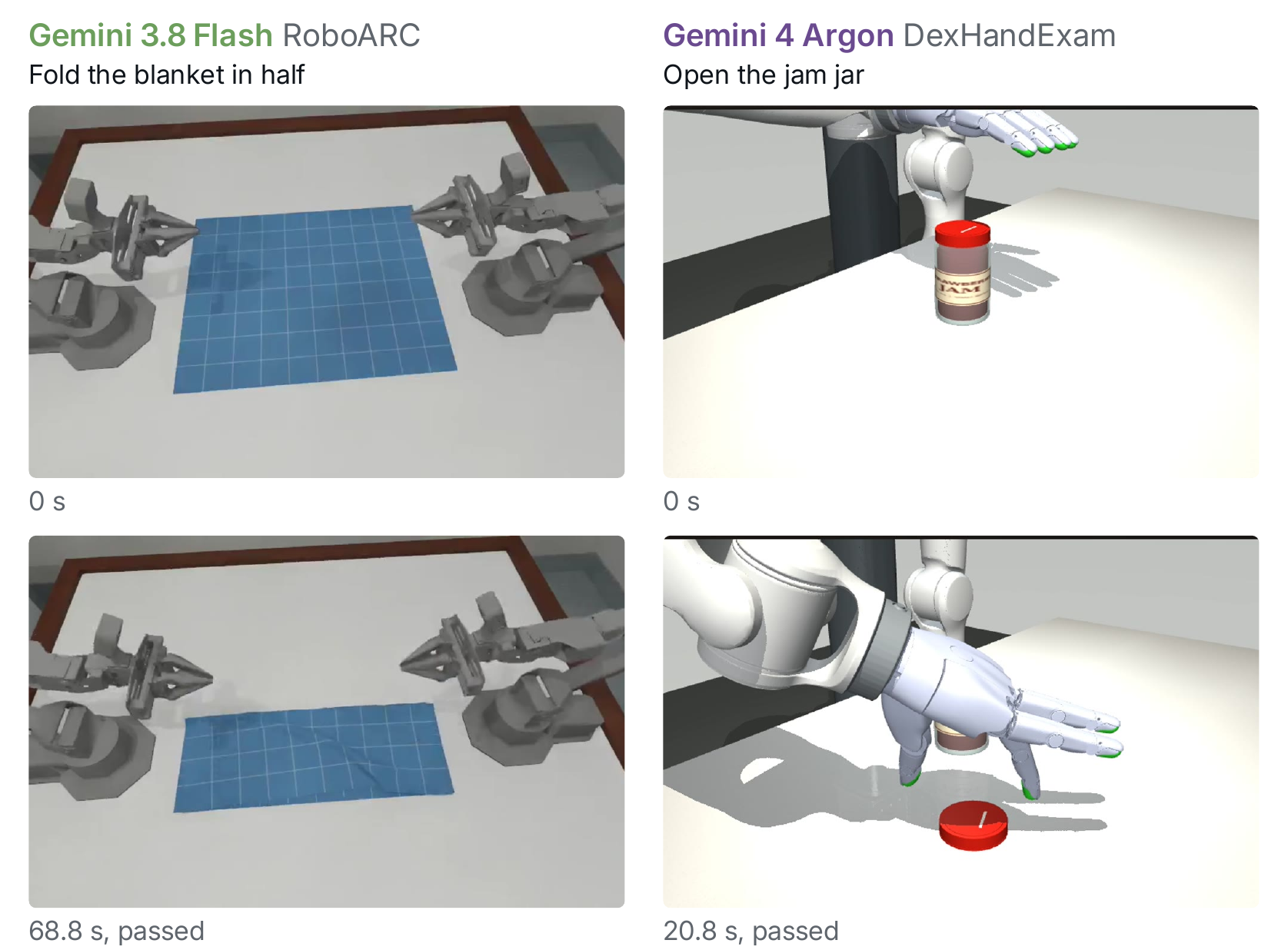}
  \caption{First and last frames of the graded episodes of two \evalsetting{Standard} tests that break the order of \Cref{fig:leaderboard}. Gemini 3.8 Flash folds a blanket in RoboARC, and Gemini 4 Argon opens a jam jar in DexHandExam.}
  \label{fig:counter-cases}
\end{figure}

\paragraph{Where the order breaks.} \Cref{tab:counter} counts the tests that break the order by suite. In the \evalsetting{Standard} Setting, Embodied Safety Exam has the most, 13 of its 50 tests, followed by RoboARC with 11 of 50. In the \evalsetting{Privileged} Setting, BEHAVIOR-1K has the most, 8 of 71. We checked three of these tests, two of which \Cref{fig:counter-cases} shows. In the RoboARC blanket test, which all seven other agents failed, Gemini 3.8 Flash computed the corner positions from camera images and depth. It grasped the two far corners with both grippers and folded the blanket in half. It then ran four adjustment programs before the episode ended with success. In an iSpatialGym test, whose goal is given as a picture of the finished scene, DeepSeek V4.1 Flash placed a wrapped food package on a plate. The simulator service ended the episode with success at control step 407 of the 1,900 that the task allows. Gemini 4 Argon opened a jam jar with a dexterous hand in a DexHandExam test, which every other agent failed. These checks cannot show that the agents that failed would fail again. The leaderboard counts one attempt per agent and test. A test that one agent solves and another fails may therefore reflect the variation between runs as well as a difference between the agents.

\paragraph{How the agents fail.} \looseness=-1 When GPT-6 Astra and another agent both fail a test after meeting different numbers of stages, GPT-6 Astra has usually met more of the task's stages (\Cref{tab:fa-agents}). The kind of failure differs as well (\Cref{fig:fa-agents-a}). The five weaker agents fail without contact at 9.1\% to 18.5\% of the contact stages they reach, against 2.7\% to 3.7\% for the three strongest agents, and the weaker agents' rate is higher in all 19 suites with at least 20 such stages per group (\Cref{app:fa-stats}). A lost hold makes up a larger share of GPT-6 Astra's failures than of any other agent's, but GPT-6 Astra also reaches more stages that need a hold. Per stage reached that needs a grasp, it loses the hold in 5.6\% of cases, against 7.5\% for the other seven agents, and the 95\% interval of the difference, $-0.2$ to 3.6 points, includes zero (\Cref{app:fa-stats}). In a BEHAVIOR-1K task that all eight agents failed, for example, the three strongest agents each packed three or four foods into a lunch box and failed in three different ways, Kimi 3.1 Pro packed two, and the other four weaker agents none (\Cref{fig:case-lunch} in \Cref{app:fa-cases}).

\paragraph{Harness.} \looseness=-1 The harness also changes the result (\Cref{fig:ablation-harness}). We ran GPT-6 Astra in Claude Code, Claude Opus 5.5 in Codex, and Gemini 4 Argon in Codex and in Claude Code, with the same tasks, time limits and reasoning effort as in the main evaluation (\Cref{app:ablation}). On the 456 core tests of the \evalsetting{Standard} Setting, Gemini 4 Argon solves 29\% in Codex and 25\% in Claude Code, compared with 24\% in its own Antigravity CLI. The difference to Codex is significant, and the difference to Claude Code is not. GPT-6 Astra solves 34\% in Codex and 22\% in Claude Code, and this difference is significant also when only the first episode counts. Claude Opus 5.5 solves 26\% in Codex and 25\% in Claude Code. Part of Gemini 4 Argon's gap may come from how each harness handles a reply that Gemini's content filter blocks: Antigravity CLI ends the session, whereas Codex sends the request again and Claude Code asks the model to continue. Attempts with a blocked reply pass at 5\% in Antigravity CLI and at about the rate of unblocked attempts in Claude Code.

\begin{figure}[h]
  \centering
  \includegraphics[width=\linewidth]{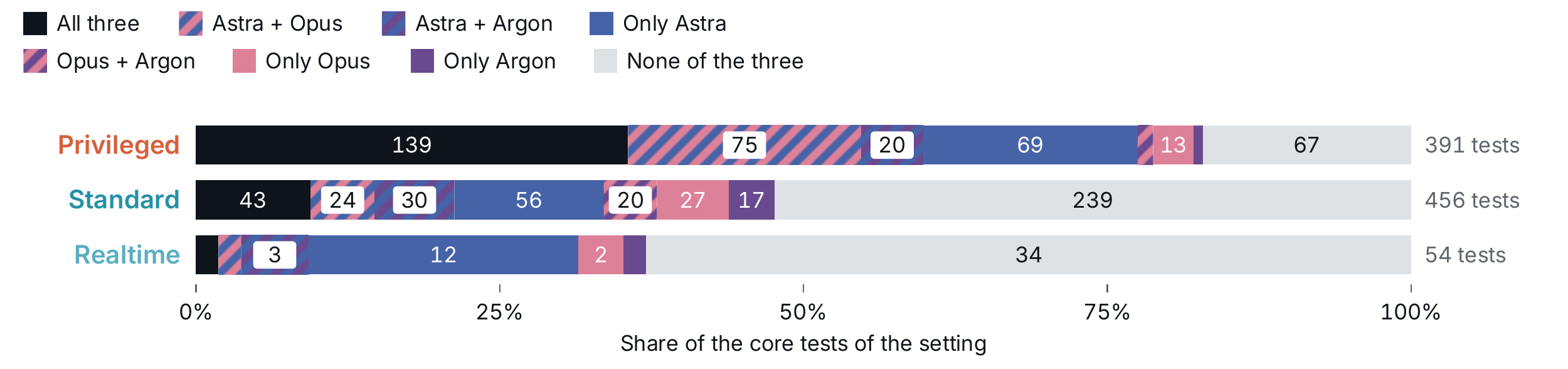}
  \caption{\looseness=-1 Core tests of each setting, grouped by which of the three strongest agents solve them.}
  \label{fig:overlap}
\end{figure}

\paragraph{Overlap among the three strongest agents.} \Cref{fig:overlap} groups the core tests of each setting by which of the three strongest agents solve them. \Cref{sec:results,sec:f-agents} give the totals. Here, an agent solves a test alone when the other two of the three fail it. In the \evalsetting{Standard} Setting, all three solve 43 tests and none of them solves 239. GPT-6 Astra alone solves 56, Claude Opus 5.5 alone solves 27 and Gemini 4 Argon alone solves 17. In the \evalsetting{Privileged} Setting, all three solve 139 tests and none of them solves 67. GPT-6 Astra alone solves 69, Claude Opus 5.5 alone solves 13 and Gemini 4 Argon alone solves 3. The \evalsetting{Privileged} tests that only GPT-6 Astra solves include tasks of precise contact and of arranging several objects. The \evalsetting{Standard} tests that only Claude Opus 5.5 solves include object transport tasks. In a ManiGuard task, for example, Claude Opus 5.5 closed the hinged lid of a glass jar and carried the jar to the goal, whereas GPT-6 Astra knocked the jar over (\Cref{app:runs}).

\subsection{Control Approaches}
\label{app:an-control}

\looseness=-1 A separate coding agent read the logs of 1,489 attempts, sampled at random, and labeled each attempt with the approach that produced its graded answer (\Cref{app:protocol}). Many \evalsetting{Privileged} tests give the agent a GPU to run the simulator, and their instructions name no training library. Apart from these, only eight tests give the agent a GPU: the GPU versions of the soccer match below and of a table tennis match, in each setting, and two RoboEngineer tests that ask the agent to train a walking policy.

The labels split \emph{Code to control} into three kinds. Open-loop programs compute the whole motion in advance. Closed-loop programs read the robot's state or cameras as they run. Parameter searches try many episodes to find a passing motion. In the \evalsetting{Privileged} Setting, agents use the simulator to compute or search for a motion before they submit it. Most of their programs in this setting are closed-loop, and 114 of the 549 programs search over many episodes. \textbf{The same agent can use different control approaches for the same task in different settings.} On a RoboTwin 2.0 task, GPT-6 Astra scripted a hand-over of the bottles with the simulator's motion primitives in the \evalsetting{Privileged} Setting. In the \evalsetting{Standard} Setting, it sent end-effector commands one at a time and checked the camera images after each command (\Cref{app:runs}).

\looseness=-1 In the \evalsetting{Standard} Setting, \emph{Direct control} places the model inside the control loop, so the robot waits for the model's reasoning before each chunk. \textbf{In the \evalsetting{Realtime} Setting, the environment keeps running, and most attempts run the control loop in a program.} In a badminton match of Embodied Olympics - Match, GPT-6 Astra let a program track the shuttlecock and swing the racket. At each pause, it revised the program from the program's logs and won 11--4 (\Cref{app:runs}).

The soccer match has a GPU version and a CPU version. In both soccer versions, each team has two Booster T1 humanoids controlled at the joint level, and the agent's team plays a team with a learned policy on the same frozen scene, with a limit of four hours. The GPU version gives the agent a GPU with 12 GiB of memory, the mjlab simulator with the rsl\_rl reinforcement-learning library and PyTorch, and training copies of the match that run on the GPU. Its instruction lets the agent train a controller or write one, and it gives an example of training a walking policy with rsl\_rl. The CPU version has no GPU and no machine-learning framework, and its instruction allows learning code in NumPy or SciPy.

\Cref{tab:soccer-tiers} counts the attempts of the eight agents on both versions in each setting, and of the four runs of the harness comparison (\Cref{app:ablation}) in the \evalsetting{Standard} Setting.

\begin{table}[ht]
  \centering
  \small
  \caption{Attempts of the agents on the two versions of the soccer task, in their own harness and in the harness comparison. ``Played with it'' counts the attempts that played the match with the policy they trained. The \evalsetting{Privileged} Setting does not record goals.}
  \label{tab:soccer-tiers}
  \begin{tabular}{llccccc}
    \toprule
    Setting & Version & Attempts & Trained with RL & Played with it & Scored a goal & Won \\
    \midrule
    \evalsetting{Standard} & GPU & 12 & 10 & 9 & 7 & 3 \\
             & CPUs only & 12 & 0 & 0 & 3 & 0 \\
    \evalsetting{Realtime} & GPU & 8 & 7 & 5 & 2 & 1 \\
             & CPUs only & 8 & 0 & 0 & 0 & 0 \\
    \evalsetting{Privileged} & GPU & 8 & 5 & 0 & -- & 2 \\
             & CPUs only & 8 & 0 & 0 & -- & 3 \\
    \bottomrule
  \end{tabular}
\end{table}

\textbf{In the \evalsetting{Standard} and \evalsetting{Realtime} settings, 17 of the 20 attempts with the GPU trained a policy with reinforcement learning, and none of the 20 with only CPUs did.} Of these 17, 14 played the match with the trained policy. DeepSeek V4.1 Flash, in the \evalsetting{Standard} Setting, wrote a joint controller by hand instead. With only CPUs, agents wrote walking gaits by hand, for example with inverse kinematics or a capture-point model. Of the 15 attempts whose final reports we read, 10 tuned the gait with CMA-ES, the cross-entropy method, or a grid or random search. One attempt, by GPT-6 Astra in Codex in the \evalsetting{Realtime} Setting, distilled its hand-written controller into a small network by supervised learning. We do not count it as training with reinforcement learning.

With the GPU, agents scored in 9 of 20 matches and won 4, all with a trained policy. With only CPUs, they scored in 3 of 20 matches and won none. Counting only the attempts in each agent's own harness, the GPU attempts trained in 13 of 16 cases and won 3 times, and the 16 CPU attempts neither trained nor won. The same models appear in several settings and harnesses, so the attempts are not independent, and we report these counts without a significance test.

\Cref{fig:approach-case} shows GPT-6 Astra in Codex on both versions of the soccer match in the \evalsetting{Standard} Setting. With the GPU, it trained walking and ball-control policies and combined them with hand-written tactics. Its robots walked 690 m, against 693 m for the opponent's, and it won 10--8. With only CPUs, it wrote a gait by hand and tuned it with differential evolution and CMA-ES. Its robots walked less than half as far as the opponent's and fell 59 times. The mercy rule ended the match early at 3--13. In this example, the two robots of GPT-6 Astra touched the ball 150 times with the GPU and 54 times with only CPUs. The opponent's robots touched it 142 and 97 times, respectively. The goal times in the figure come from the referee state that the agent queried after each goal, and the frames come from the replay of each attempt.

\textbf{In the \evalsetting{Privileged} Setting, every win reused the opponent's walking network.} With the GPU, 5 of 8 attempts ran reinforcement-learning code and 2 won. With only CPUs, 3 of 8 attempts won. In this setting, the agent can replay the same match in its own simulator. According to their code or final reports, all five winning attempts reused the opponent's walking network that comes with the simulator, and none used a policy that it had trained.

We decided whether an attempt trained a policy from three signals: calls to a reinforcement-learning library or gradient updates in the agent's code, the GPU memory that the resource monitor recorded, and the attempt's final report, which another agent read. The three signals agree on every \evalsetting{Standard} and \evalsetting{Realtime} attempt with the GPU. The comparison covers one task and one frozen scene. The two versions differ at the same time in the GPU, the library, the training copies of the match and the instruction, so the comparison cannot separate their effects.

%% file: appendix_sections/c_runs/c_runs.tex
This section describes further attempts that illustrate the findings. Every attempt is public, and most can be replayed step by step on the exam's website.

\paragraph{RoboLab, the white mug in the center of the table (\evalsetting{Standard}, solved by Gemini 4 Argon).} Gemini 4 Argon located the table and the mug in the point clouds of three calibrated depth cameras and found that the mug, 9~cm across, was wider than the gripper's largest opening. It therefore grasped the mug's wall at the rim, with one finger inside the mug, and moved the mug to the center of the table with inverse-kinematics targets sent as joint positions. To let go, it set the mug down and moved the inner finger to the middle of the mug before lifting the gripper. The episode, its only one, succeeded after about half an hour. GPT-6 Astra failed this test.

\paragraph{DexToolBench, red brush (\evalsetting{Standard}, solved by none of the three strongest agents).} \looseness=-1 A DexToolBench task asks the agent to move a tool with a dexterous hand through goal poses, 38 in this task. Claude Opus 5.5 pushed the brush until it stuck out over the table edge, grasped it and reached goals 0 to 4. Goal 5 needed a 90-degree turn about the brush's long axis. The palm-down grasp could not make it, and the brush flew out of the hand as the arm changed configuration.

\paragraph{RoboARC, IKEA Stegesund (\evalsetting{Privileged}, failed).} With the full state, GPT-6 Astra pressed the halves together, slid in the shelf and drove the first bolt. Its hex key turned only 120 degrees per insertion, so one bolt took 40 minutes, and at the three-hour limit the second bolt was still 3.7~mm short. None of the three strongest agents solved this task.

\paragraph{BEHAVIOR-1K, setting a table (\evalsetting{Privileged}, failed).} GPT-6 Astra spent 62 minutes placing two plates, and the robot then tipped over against the kitchen island on the way to the fridge.

\paragraph{RoboARC, threading a rope through six holes (\evalsetting{Standard}, failed).} \looseness=-1 Claude Opus 5.5 mapped the wall, its holes and the coiled rope from depth point clouds. The rope slipped from the gripper whenever the robot turned its wrist, and the agent threaded none of the six holes.

\paragraph{RoboDojo, inserting a key (\evalsetting{Standard}, failed).} \looseness=-1 Claude Opus 5.5 located the key and the lock from depth images but dropped the key at least nine times during hand-overs and large joint moves. Its most accurate insertion, at minute 36, was about 9~mm off and hit the face of the lock.

\paragraph{RoboDynamicExam, toss, recatch and deliver (\evalsetting{Standard}, failed).} \looseness=-1 In this task, unlike that of \Cref{fig:case-ball}, the robot catches a thrown ball in a cup, tosses it up out of the cup itself, catches it again and releases it into a moving box. GPT-6 Astra caught the ball using depth images, but its first episodes ended early (one ended when the cup grazed a moving box) and used up its three allowed resets. In the last episode it caught and tossed the ball, but on the second catch the ball's contact with the tilted cup was flagged as a limit violation.

\paragraph{RoboPaint, a character in running script (\evalsetting{Standard}, failed).} \looseness=-1 With no resets available, GPT-6 Astra calibrated the stroke width on a practice sheet, wrote five strokes and then retouched the character in one long stroke for 38 minutes. The overlap and edge scores met their thresholds, but the long stroke raised the total length of the inked path to 2,048~mm, above the limit of 1,450~mm. In addition, one of the nine regions was only 65.7\% covered, below the required 80\%.

\paragraph{HumanoidBench, opening a door (\evalsetting{Standard}, failed).} Claude Opus 5.5 kept the humanoid standing with ankle feedback and located the handle with stereo cameras. It tried more than ten ways of opening the door (pressing, hooking, pushing up and using both hands) and pulled the handle back by up to 26~cm, but the door never opened. The agent used 45 resets in 27 minutes.

\paragraph{Embodied Olympics - Fun, dribble slalom (\evalsetting{Privileged} solved, \evalsetting{Standard} failed).} \looseness=-1 In the \evalsetting{Privileged} Setting, Gemini 4 Argon saved and restored exact simulator states to try candidate kicks in eight parallel processes and passed all 13 checkpoints of the slalom. In the \evalsetting{Standard} Setting, it estimated the positions of the robot and the ball from the pixels of a fixed camera, about 9~cm off to the side, and its best episode passed four poles before the 120 minutes ran out.

\paragraph{BiGym 2.0, closing the dishwasher (\evalsetting{Privileged}, solved).} \looseness=-1 Gemini 4 Argon read the dishwasher's code and collision shapes and pushed both racks in. The robot then crouched, hooked the door's front edge and swung the door shut along an arc. Its answer, a 693-step trajectory, passed on replay.

\paragraph{RoboCasa365, defrosting meat in the microwave (\evalsetting{Privileged}, solved).} Claude Opus 5.5 used Jacobian inverse kinematics and closed-loop control of the base to take the box out of the freezer, put it in the microwave, close the door and press start. The agent submitted its answer at minute 46, and the answer passed on replay.

\paragraph{ManiGuard, dusty transfer (Gemini 4 Argon, \evalsetting{Privileged}, solved).} Working in a copy of the simulator that it kept running, Gemini 4 Argon wiped 18 dust particles from the box and then tilted the tray so that the potato rolled in without being touched. It finished at minute 26, and its answer passed on replay.

\paragraph{RoboPaint, cursive character (\evalsetting{Privileged}, solved).} \looseness=-1 GPT-6 Astra read how the brush model draws a stroke, wrote a C++ program that reproduces this model and fitted three strokes with it. The first strokes that the agent executed passed every check, and the agent submitted its answer after four minutes.

\paragraph{BEHAVIOR-1K, bringing in plywood (\evalsetting{Privileged} solved, \evalsetting{Standard} failed).} A mobile robot with two arms must carry three plywood sheets from the garden onto the corridor floor. In the \evalsetting{Privileged} Setting, Claude Opus 5.5 kept a copy of the simulator running, wrote separate scripts for grasping, opening the door and carrying, and tested each stage in this copy, where a fall of the robot required only a reset. Its answer passed after 72 minutes. In the \evalsetting{Standard} Setting, the same agent picked up two of the sheets, one in each gripper, but then drove the base backwards and sideways with a single command and did not check the camera images while the base moved, until the robot hit a curb and tipped over.

\paragraph{ManiGuard, dusty transfer (GPT-6 Astra, \evalsetting{Privileged} solved, \evalsetting{Standard} failed).} \looseness=-1 The robot must wipe a box clean and then tip a tray so that a potato rolls into the box without being touched. GPT-6 Astra touched or flung the potato in its first three \evalsetting{Privileged} episodes, then reset and passed with a new wrist angle. In the \evalsetting{Standard} Setting, the test failed because a finger touched the potato once while the robot regrasped the tray, although the potato later reached the box.

\paragraph{DexToolBench, screwdriver (\evalsetting{Standard}, failed).} \looseness=-1 Claude Opus 5.5 estimated the axis of a screwdriver from the point cloud but could not determine how far the handle had rolled about that axis, and it used all 20 resets that this task allows without reaching any of its 76 goal poses.

\paragraph{ManiGuard, closing a jar (\evalsetting{Standard}, solved by Claude Opus 5.5).} The robot must close the hinged lid of a glass jar and carry the jar to a goal without breaking the task's safety rules. Claude Opus 5.5 estimated the jar's shape and the position of the hinge from the depth point cloud, slid the robot's closed fingers under the lid and swung the lid shut about the hinge. It then lifted the jar by 1~cm to check the grasp before carrying the jar to the goal. GPT-6 Astra knocked the jar over while pushing the lid.

\paragraph{RoboDynamicExam, toss and recatch (\evalsetting{Privileged}).} Claude Opus 5.5 used the arm's Jacobian to compute the largest downward acceleration of the point from which the ball is tossed, and it chose its toss and recatch to stay within this limit.

\paragraph{RoboWits, tipping a cube out of a container (\evalsetting{Privileged}).} Claude Opus 5.5 tried six pour positions to tip the cube out of the container and kept the steadiest.

\paragraph{RoboTwin 2.0, three bottles into a dustbin (\evalsetting{Privileged} and \evalsetting{Standard}).} In the \evalsetting{Privileged} Setting, GPT-6 Astra read the task's source code, scripted the grasps with the simulator's motion primitives, handed the bottles from one arm to the other in mid-air and submitted its answer after five minutes. In the \evalsetting{Standard} Setting, it sent end-effector commands one at a time and checked the camera images after each command. It knocked one bottle over, picked the bottle up again with the other arm and finished about 30 seconds before the time limit.

\paragraph{Embodied Olympics - Match, badminton (\evalsetting{Realtime}, won).} \looseness=-1 A program written by GPT-6 Astra ran in the background, located the shuttlecock with two cameras, fitted a drag model to its flight and swung the racket. Whenever play paused, the agent read the program's logs and revised the program. It won 11--4 against an opponent that picks its shots with a learned policy.

%% file: appendix_sections/f_ablation/f_ablation.tex
This section defines the controlled runs of \Cref{sec:f-settings,sec:f-agents} (\Cref{fig:ablation-ladder,fig:ablation-harness}).

\paragraph{Baseline.} Every controlled run of \Cref{fig:ablation-ladder} uses GPT-6 Astra in Codex with the same reasoning effort, time limit and task versions as the leaderboard run of \Cref{fig:leaderboard}, and the baseline is that leaderboard run. Each run covers the default tests of the 456 core tasks, and \Cref{tab:ablation} and \Cref{fig:ablation-ladder} compare the runs on the 391 core tasks that have both a \evalsetting{Privileged} and a \evalsetting{Standard} test.
Each run changes one or two features of the \evalsetting{Standard} interface and nothing else. Apart from these features, every task behaves exactly as in the leaderboard run.

\paragraph{Object-state query.} We add two read-only calls to the \evalsetting{Standard} interface: one lists the objects in the scene, and one returns their state. The state consists of position, orientation, size and bounding box, linear and angular velocity, and the joint positions of the robot and of articulated objects such as doors and drawers. In Embodied Olympics - Match, the state of the opponent and of the ball counts as object state. The calls do not advance the simulation or use up simulated time, and they never return success flags, scores, thresholds or goal regions. The only progress value they return is the paint coverage of the canvas in RoboPaint. We remove the fields that would reveal what the agent is meant to find out itself: goal markers and zones, objects that appear only later in a BEHAVIOR-1K task, hint notes in RoboCasa365, and hazard labels in Embodied Safety Exam.

\paragraph{Unlimited resets with saved states.} The reset budget of every task is raised to 1,000, and the interface gains calls that save the current simulator state under a handle and restore it later. The simulator service still grades the last episode as it runs. Restoring a saved state does not erase a rule violation that has already been recorded. In Embodied Safety Exam, for example, recorded violations remain after a restore.

\paragraph{Both.} This run combines the two features above.

\paragraph{No success signal.} The interface no longer reports that an episode has succeeded. The episode does not end at success, and no success flag, progress or score is returned. The agent ends the episode itself with a finish call, or the episode ends when the time runs out or the task's own failure rule is triggered. The simulator service checks the success rule on the state at the end of the episode, so a success that was reached and then undone counts as a failure.

\paragraph{Harness comparison.} For \Cref{fig:ablation-harness}, we ran GPT-6 Astra in Claude Code, Claude Opus 5.5 in Codex, and Gemini 4 Argon in Codex and in Claude Code, with the same tasks, time limits and reasoning effort as the leaderboard runs. Codex and Claude Code support only their own provider's API format, so an adapter passes their requests to the other provider and translates the answers back. The adapter also adapts the system prompt and model metadata, so these runs only approximate a native setup. We score every run on the 456 core tests of the \evalsetting{Standard} Setting, the tests of the \evalsetting{Standard} leaderboard. Each group of bars in \Cref{fig:ablation-harness} compares a model's runs on these 456 tests, and we test differences with a paired exact McNemar test.

\paragraph{Further results of the controlled runs.} \Cref{tab:ablation} gives the success rates of each run. Without the success signal, the success rate stays within a few points of the baseline.

\looseness=-1 The agent uses the saved states to retry. With unlimited resets, only 14\% of its successes came in the first episode, and 66\% came in an episode that started from a saved state. The saved states let the agent return to the moment before a hard stage instead of repeating the whole task. The state query helps most where the scene is hard to measure from cameras, as in iSpatialGym, RoboDojo and BEHAVIOR-1K, and it changes little in Embodied Olympics - Fun and DexHandExam.

\begin{table}[h]
  \centering
  \small
  \caption{Success rate (\%) of GPT-6 Astra in Codex in the controlled runs, on the 391 core tasks that have both a \evalsetting{Privileged} and a \evalsetting{Standard} test, and in the \evalsetting{Realtime} column on the 50 \evalsetting{Realtime} tests of these tasks. The \evalsetting{Standard} column includes the \evalsetting{Realtime} tests. The \evalsetting{Privileged} row uses the \evalsetting{Privileged} tests of the same tasks.}
  \label{tab:ablation}
  \begin{tabular}{lcc}
    \toprule
    Run & \evalsetting{Standard} & \evalsetting{Realtime} \\
    \midrule
    \evalsetting{Standard} (baseline) & 28 & 30 \\
    + object-state query & 45 & 38 \\
    + unlimited resets and save/load states & 53 & 64 \\
    + both & 68 & 80 \\
    \evalsetting{Privileged} & 77 & 84 \\
    \midrule
    No success signal & 31 & 26 \\
    \bottomrule
  \end{tabular}
\end{table}

\begin{table}[h]
  \centering
  \small
  \caption{Success rate (\%) of each model in each harness, on the 456 core tests of the \evalsetting{Standard} Setting. The first number counts the final result, and the second counts first-episode success. The model's own harness is marked with an asterisk.}
  \label{tab:harness}
  \begin{tabular}{lccc}
    \toprule
    Model (tests) & Codex & Claude Code & Antigravity CLI \\
    \midrule
    GPT-6 Astra (456) & 34 / 26$^{*}$ & 22 / 18 & -- \\
    Claude Opus 5.5 (456) & 26 / 21 & 25 / 21$^{*}$ & -- \\
    Gemini 4 Argon (456) & 29 / 25 & 25 / 24 & 24 / 21$^{*}$ \\
    \bottomrule
  \end{tabular}
\end{table}

\paragraph{Further results of the harness comparison.} \Cref{tab:harness} gives the success rate of each model in each harness, counted by the final result and by first-episode success, that is, success in the first episode before any reset (\Cref{app:fw-grading}). Gemini 4 Argon does better in Codex than in its own harness (paired exact test, $p=0.02$ for the final result and $p<0.05$ for first-episode success). In Claude Code, the difference is not significant under either count ($p=0.61$ and $p=0.20$). In both harnesses, BEHAVIOR-1K contributes the largest part of its gain. GPT-6 Astra does better in Codex than in Claude Code under both counts ($p<0.001$), and most of this difference is already present in the first episode. For Claude Opus 5.5, the difference is not significant under either count. These runs are reported separately from the leaderboard of \Cref{fig:leaderboard}.

%% file: appendix_sections/d_failure/d_failure.tex
\looseness=-1 This section describes the stage-level failure labels of \Cref{sec:labels}. \Cref{app:fa-procedure} explains how they were made, \Cref{app:fa-taxonomy} gives the taxonomy, and \Cref{app:fa-exclusions} lists the excluded attempts. \Cref{app:fa-breakdown} breaks the failures down by suite, robot body, agent and setting. \Cref{app:fa-stats} gives further statistics, in the order of \Cref{sec:f-settings,sec:f-difficulty,sec:f-agents}, and the limitations of the labels. \Cref{app:fa-cases} describes the case studies of the figures in full, with further examples. The labels are released with the attempts at \nolinkurl{huggingface.co/datasets/MLL-Lab/efe-failure-analysis}.

\subsection{Labeling Procedure}
\label{app:fa-procedure}

\paragraph{Task graph.} For each of 427 core tasks, the labeling agent, GPT-6.1 Sol in Codex, wrote one graph of stages, such as the graph of \Cref{fig:fa-method-a} in \Cref{sec:labels}, from the task instruction, the verifier's code and a successful attempt when one exists. A stage is a change in the world that a person would list when describing how to do the task, such as ``open the fridge'' or ``put the first mango in the fridge''. Conditions on how well a stage must be done, such as thresholds, alignment or ``without touching X'', belong to that stage's condition and are not stages of their own. A condition that is already true at the start and must still be true at the end, such as a closed fridge, becomes a stage of its own that is met from the start, so that the progress map shows when an attempt breaks it. Each stage has a stage type, its prerequisite stages, a condition that can be checked on the replay video, and the contact it needs. The contact is a list of pairs of the robot part that touches and how it touches (who $\times$ how), one list for each way of doing the stage. The edges of the graph give the order in which the stages must be met for the task to pass; a stage that holds from the start can hold before its prerequisites. The stage types fall into nine categories (\Cref{tab:fa-steps}). For example, material stages change the state of a material, such as pouring or cleaning, and body stages move the robot's own body, such as walking or keeping balance. The 427 graphs have 2,411 stages, 5.6 per task on average.

\paragraph{Progress map.} For each attempt, the labels give the time intervals during which each stage is met, on the timeline of the graded episode's replay video (\Cref{fig:fa-method-b}). The condition of a stage can stop holding after it has been met, so the progress map can go down. Progress is read from the replay video, or from the logs where there is no video; attempts without a submitted trajectory usually have no readable progress. An attempt reaches a stage if it meets the stage's condition at some moment or if the stage is its failure stage.

\paragraph{Failure stage, phenomena and root causes.} Each failed attempt is assigned at most one failure stage, so that failures can be counted by stage and compared across attempts and agents. A failure stage is located when the labels give one.
 If an irreversible event occurred, such as a fall or a violation of a task rule that makes the attempt fail, the failure stage is the stage during which the event occurred, unless an earlier stage had already failed permanently. Otherwise, it is the stage with the earliest uncorrected mistake among the stages whose condition was never met. Mistakes that the agent later corrected do not count. The labels also record where the attempt was when it failed: during the stage, between stages, while exploring, before the first stage or at the final success check. Up to two failure phenomena describe what visibly went wrong at the failure stage, and up to two root causes describe why it happened, with a video time or log entry supporting each cause. A time limit, a used-up budget or the agent stopping is never a phenomenon or a cause: the label names what was going wrong when the attempt ended. A root cause is given only when the evidence supports it, and it may stay at the category level ``Control'' when the subtype cannot be determined, as in 47\% of the analyzed failures.

\paragraph{Writer and reviewer.} GPT-6.1 Sol in Codex, at high reasoning effort, wrote every graph and the label files of most attempts from the inputs above, and Claude Sonnet 5.5 in Claude Code wrote those of the remaining attempts from the same inputs with the same prompts. Claude Sonnet 5.5 in Claude Code reviewed each file, with up to two rounds of fixes and a final confirmation round. The prompts were frozen before the final labeling. \Cref{sec:f-control} uses a second set of labels, which give the control approach of a sample of attempts (\Cref{app:protocol}). Agents wrote both sets of labels.

\paragraph{Human oversight.} We defined the taxonomy (\Cref{app:fa-taxonomy}) and settled it in team meetings before the final labeling, and the agents assign its values to each task and attempt. Earlier versions of the labels were reviewed by hand and in team meetings. Each error we found was written into the prompts as a worked example: the task or attempt, the wrong label, the right label and the rule. Most of the worked examples in the final prompts come from this review, and the rest come from rechecks by agents. The labels used here were then written again with the final prompts.

\paragraph{Core set covered.} The labels cover 6,779 attempts on 857 tests of 427 tasks, with at most one labeled attempt per agent and test.

\subsection{Taxonomy}
\label{app:fa-taxonomy}

\Cref{tab:fa-steps,tab:fa-contact,tab:fa-phenomena,tab:fa-causes} list the values of the taxonomy. A root cause can also stay at the category level, such as Control (\Cref{app:fa-procedure}). Stage types let us compare stages across tasks. A stage type names the robot action that directly brings about the stage's change in the world, and the types fall into the nine categories of \Cref{tab:fa-steps}. In the task of \Cref{fig:fa-method}, for example, opening the fridge is a mechanism stage of type open, and putting the tomato in is a relation stage of type put in. The contact of a stage says who touches, such as a gripper, both arms or the mobile base, and how, such as grasping, pushing or supporting the robot's weight. Failure phenomena describe what visibly went wrong at the failure stage. In the same example, Claude Opus 5.5 fails with a loss of balance, and DeepSeek V4.1 Flash fails with a lost hold. Root causes, such as a wrong perception or an unsuitable grasp, describe why a failure happened.

\begin{table}[ht]
  \centering
  \footnotesize
  \caption{Stage types, by category. Competition stages (score, defend, win) are counted by category only.}
  \label{tab:fa-steps}
  \begin{tabularx}{\linewidth}{@{}l>{\raggedright\arraybackslash}X@{}}
    \toprule
    Category & Stage types \\
    \midrule
    Relation & place, put in, take out, gather, stack, insert, assemble, disassemble, screw/unscrew, thread/hang, connect, hand over \\
    Pose & pick up, regrasp, push, pull, flip over, adjust pose, follow path, stabilize \\
    Dynamic & throw, strike, catch, dribble \\
    Mechanism & open, close, rotate, press, switch on/off \\
    Material & fold, flatten, untangle, pour, scoop, cut, clean, stir, draw/write, change temperature, ignite \\
    Body & navigate, traverse, jump, perform acrobatics, hang and swing, maintain balance, maintain pose \\
    Safety & refuse, warn, mitigate \\
    Information & search, inspect, communicate \\
    Competition & score, defend, win \\
    \bottomrule
  \end{tabularx}
\end{table}

\begin{table}[ht]
  \centering
  \footnotesize
  \caption{Contact of a stage: who touches and how.}
  \label{tab:fa-contact}
  \begin{tabularx}{\linewidth}{@{}l>{\raggedright\arraybackslash}X@{}}
    \toprule
    Who & \\
    \midrule
    Fingers/gripper & one hand or gripper touches the object directly, also when the pose of a held tool is graded \\
    Bimanual & two hands or arms act on the stage together \\
    Tool & the robot acts through a held or mounted tool \\
    Legs/torso & feet, legs, head, torso or landing gear on a support surface \\
    Mobile base & wheels or a wheeled base, while moving \\
    No contact & no contact to label, as when flying, observing or reporting \\
    \midrule
    How & \\
    \midrule
    Grasp & the hand closes on the object \\
    Push/sweep & push, nudge or sweep without closing the hand \\
    Press/wipe & press or wipe with pressure on the contact surface \\
    Cradle/scoop & support, catch or scoop the object from below \\
    Hook & hook with a finger or a tool \\
    In-hand adjustment & reorient or reposition the object with the fingers without letting go \\
    Support & the body or base bears weight, as in standing or landing \\
    Avoid contact & pass without touching a named object \\
    \bottomrule
  \end{tabularx}
\end{table}

\begin{table}[ht]
  \centering
  \footnotesize
  \caption{\looseness=-1 Failure phenomena: what visibly went wrong at the failure stage, up to two per attempt.}
  \label{tab:fa-phenomena}
  \begin{tabularx}{\linewidth}{@{}ll>{\raggedright\arraybackslash}X@{}}
    \toprule
    Category & Phenomenon & Meaning \\
    \midrule
    Contact & Failed to make contact & the contact the stage needs never existed, as in a missed grasp or fingers that never closed on the object \\
     & Lost hold & held and then lost: slipped or dropped while held or carried \\
     & Abnormal contact force & visible or sensed wrong force, such as a part not fully inserted or a latch not engaged \\
     & Failed to release & the object stays in the hand or tool when the stage needs it to leave \\
     & Unstable after release & after a planned release the object falls over, rolls away or bounces \\
     & Drags the base & manipulating a part moves the body that should stay still \\
    Motion & Inaccurate pose & the hand, tool or object misses the required position or orientation \\
     & Obstruction/collision & blocked by contact, or knocking over or displacing objects \\
     & Out of reach & evidence of a reachability or joint limit \\
     & Dynamic timing error & contact with or release of a moving object too early or too late \\
    Whole body & Loss of balance & any robot loses support or tips over \\
     & Gait failure & walking, running or jumping falls short without a fall \\
    Constraint & Constraint violation & a task rule broken, with no more specific phenomenon \\
    Reporting & Missing report, incorrect report & a graded information act, such as a warning or a required answer, missing or wrong \\
    Not attempted & Not attempted & a stage the graded attempt needed was never started \\
    \bottomrule
  \end{tabularx}
\end{table}

\begin{table}[ht]
  \centering
  \footnotesize
  \caption{\looseness=-1 Root causes: why the failure happened, up to two per attempt, each with its evidence.}
  \label{tab:fa-causes}
  \begin{tabularx}{\linewidth}{@{}ll>{\raggedright\arraybackslash}X@{}}
    \toprule
    Category & Root cause & Meaning \\
    \midrule
    Cognition & Task understanding & the behavior shows the task was misread \\
     & Reasoning/planning & wrong goal, stage order or overall strategy \\
     & Perception & the robot acted on a wrong estimate of where or what something is \\
     & Safety judgment & records show a hazard was not recognized or misjudged \\
     & False completion belief & a false belief that the task was done affected stopping or decisions \\
    Control & Grasp and contact design & the contact geometry or load layout does not suit the stage \\
     & Trajectory or IK & goal and contact plan are feasible, but no trajectory or IK solution is found or followed \\
     & Feedback control & tracking or stabilization under feedback fails \\
     & Parameter tuning & parameters shown to be wrong \\
    Delivery & Delivery operation error & recording, saving or checking the submitted answer went wrong, such as recording without a reset \\
    Task or environment & Task or environment & every legal solution is ruled out, or an event the agent cannot react to \\
    \bottomrule
  \end{tabularx}
\end{table}

\subsection{Excluded Attempts}
\label{app:fa-exclusions}

The eight agents failed 5,272 of their 6,779 labeled attempts. The analyses in \Cref{sec:f-settings,sec:f-difficulty,sec:f-agents} exclude 500 of these failures, in two groups that overlap in 83 attempts. \Cref{tab:fa-excluded} counts them by agent and setting. Both groups are failures under the rules of the exam. They are excluded only from the statistics computed from the failure labels.

\begin{table}[ht]
  \centering
  \footnotesize
  \caption{Failed attempts excluded from the failure analysis, by agent and setting. P, S and R stand for the \evalsetting{Privileged}, \evalsetting{Standard} and \evalsetting{Realtime} settings. The column ``with failure stage'' counts filter cutoffs whose labels also give a failure stage with a physical phenomenon. The column ``Excluded'' counts filter cutoffs and missing trajectories, and an attempt in both groups counts once. GPT-6 Astra, Claude Opus 5.5 and DeepSeek V4.1 Flash are not listed: they have no filter cutoffs and 1, 2 and 4 missing trajectories in the \evalsetting{Privileged} Setting.}
  \label{tab:fa-excluded}
  \begin{tabular}{@{}llrrrrr@{}}
    \toprule
    Agent & Setting & Failed & Filter cutoffs & with failure stage & Missing trajectory & Excluded \\
    \midrule
    Gemini 4 Argon & P & 221 & 108 & 9 & 134 & 163 \\
     & S & 326 & 36 & 22 & 0 & 36 \\
     & R & 47 & 9 & 3 & 0 & 9 \\
    Kimi 3.1 Pro & P & 238 & 0 & 0 & 49 & 49 \\
    Gemini 3.8 Flash & P & 291 & 4 & 1 & 97 & 98 \\
     & S & 400 & 3 & 2 & 0 & 3 \\
    Qwen3.8 Max & P & 311 & 1 & 0 & 94 & 94 \\
    GLM 5.3 Flash & P & 340 & 0 & 0 & 41 & 41 \\
    \midrule
    All eight agents & & 5,272 & 161 & 37 & 422 & 500 \\
    \bottomrule
  \end{tabular}
\end{table}

\paragraph{Filter cutoffs, 161 attempts.} In these attempts the model provider's content filter stopped the agent's session, and the attempt ended or submitted nothing. Gemini 4 Argon accounts for 153 of them, Gemini 3.8 Flash for 7, and Qwen3.8 Max for one. Antigravity CLI, the harness of both Gemini models, ends the session after a blocked reply, whereas Codex sends the request again and Claude Code asks the model to continue (\Cref{app:an-agents}). In 37 of the 161 attempts, the labels also give a failure stage with a physical phenomenon, and the column ``with failure stage'' of \Cref{tab:fa-excluded} counts them. Of these 37, 34 are Gemini 4 Argon's, 22 of them in the \evalsetting{Standard} Setting. Because almost all of them are Gemini 4 Argon's, its failure profile in \Cref{fig:fa-agents-a} and \Cref{tab:fa-agents} is the statistic that this choice affects most.

\paragraph{Missing trajectories, 422 attempts.} In the \evalsetting{Privileged} Setting the agent's answer is a recorded trajectory that the verifier replays or, in Embodied Olympics - Match, a program that the verifier runs. In these attempts the agent never submitted an answer, and the verifier recorded neither a trajectory nor a replay. We exclude these attempts because they leave no submitted trajectory or graded replay to label. Apart from the 83 that are also filter cutoffs, a missing trajectory is the agent's own failure. Gemini 4 Argon has 134 such attempts, of which 79 are also filter cutoffs. Gemini 3.8 Flash has 97, Qwen3.8 Max 94, Kimi 3.1 Pro 49 and GLM 5.3 Flash 41. For Gemini 3.8 Flash and Qwen3.8 Max they are about a third of all \evalsetting{Privileged} failures, 97 of 291 and 94 of 311.

\paragraph{Failures without readable progress, 101 attempts.} Besides the 500 failures above, the stage-level statistics, namely the failure rates of stages and the predictions of the failure stage, exclude 101 failed attempts whose labels give no stage progress, because neither the video nor the log shows when the stages held. These statistics therefore use 6,178 attempts, passed and failed: the 6,779 labeled attempts without the 500 excluded failures and these 101. The 101 attempts remain in the statistics that do not need stage progress, such as the failure profiles of \Cref{tab:fa-agents}. The 72 of them with a located failure stage, that is, a failure stage that the labels give, also remain in the concentration of failure stages and in the common failure stage.

\subsection{Failures by Suite, Robot Body, Agent and Setting}
\label{app:fa-breakdown}

\begin{figure}[p]
  \centering
  \includegraphics[width=\linewidth]{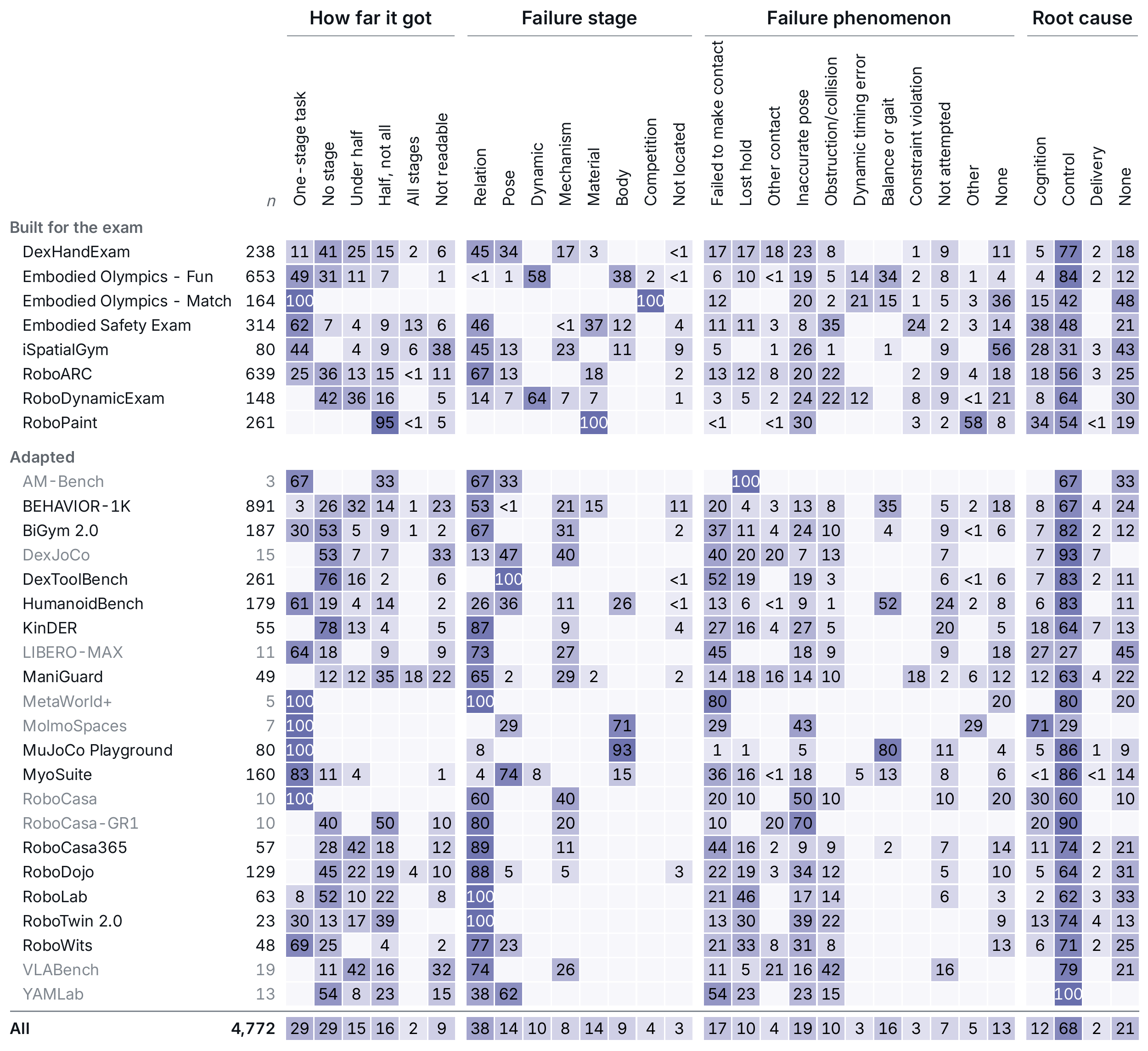}
  \caption{Failures by suite. Each cell gives the share (\%) of the suite's analyzed failures in the column, with the three settings pooled, and $n$ is the number of failures. A failure can have up to two phenomena and up to two root causes, so these two blocks can add up to more than 100\%.}
  \label{fig:fa-breakdown-benchmark}
\end{figure}

\begin{figure}[t]
  \centering
  \includegraphics[width=\linewidth]{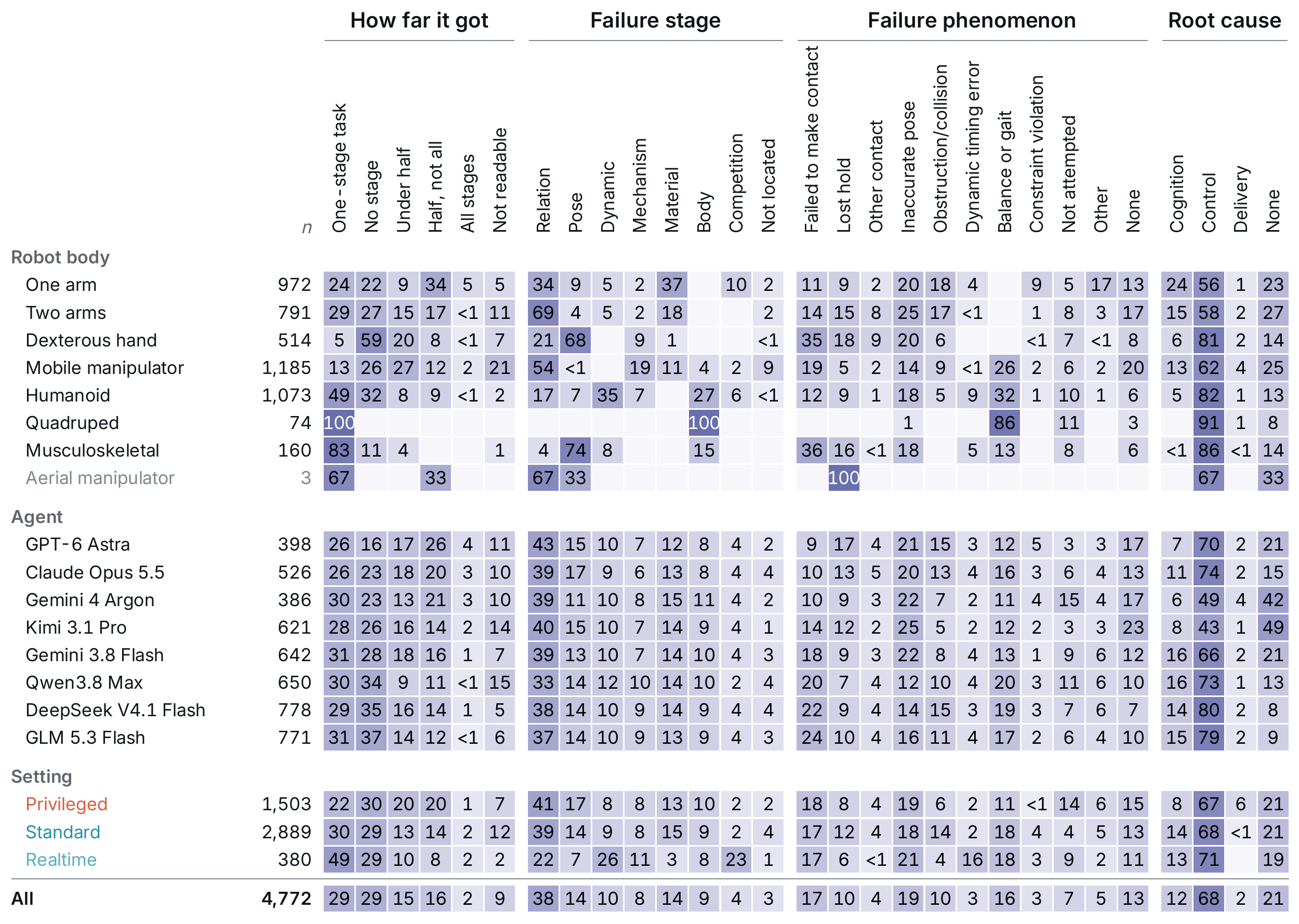}
  \caption{Failures by robot body, agent and setting, in the same format as \Cref{fig:fa-breakdown-benchmark}. Here too, the phenomena and the root causes can add up to more than 100\%.}
  \label{fig:fa-breakdown-robot}
\end{figure}

\looseness=-1 \Cref{fig:fa-breakdown-benchmark,fig:fa-breakdown-robot} break the 4,772 analyzed failures down by suite, robot body, agent and setting. Each row splits the group's failures among the labels in four blocks, so its values are shares of the failures and should not be read as failure rates. The first block says how far the failed attempt got. Failures on tasks with a single stage form their own column. For longer tasks, the block uses the attempt's peak progress, the largest share of the task's stages met at one moment, counting stages already met at the start. Its columns are no stage, more than none but under half, at least half but not all, all stages, and not readable. Peak progress is not readable when the status of at least one stage could not be read, and this column holds 448 failures, among them 98 of the 101 failures without any readable progress (\Cref{app:fa-exclusions}). The second block gives the category of the failure stage. The third and fourth blocks give the failure phenomena and the root causes. A failure can have up to two of each, so these two blocks can add up to more than 100\%. The rows pool the three settings, except the rows of the settings themselves. The robot body is the kind of body of the task's robot, as in \Cref{tab:suites}.

Shares are rounded, $<$1 marks a share above zero but below 1\%, and empty cells are zero. Gray labels mark groups with fewer than 20 failures. The phenomena and root causes are those of \Cref{tab:fa-phenomena,tab:fa-causes}, and None means that a failure has no label in the block. Other contact pools the four rarer contact phenomena, and Balance or gait pools loss of balance and gait failure. Other covers out of reach, reporting and defects that the taxonomy does not name, such as gaps in a painting or folds left in a cloth. No analyzed failure lies at a safety or information stage or has a task or environment root cause, so these columns are omitted.

\paragraph{By suite.} \Cref{fig:fa-breakdown-benchmark} gives the rows of the suites. The categories of the failure stages differ across suites, and in some suites all failures lie at one kind of stage. Every RoboPaint failure is at a material stage, every Embodied Olympics - Match failure is at a competition stage, and every DexToolBench failure with a located failure stage is at a pose stage. In Embodied Olympics - Match and DexToolBench, every stage of every task is of that category. How far the failed attempts get also differs. In DexToolBench and KinDER, most failed attempts never meet a stage of the task, while in RoboPaint almost every failed attempt meets at least half of the stages. In Embodied Safety Exam and ManiGuard, some failed attempts meet all stages at one moment. In Embodied Safety Exam, most of these attempts have a constraint violation. Failures at the same kind of stage can still differ between suites. In both RoboLab and RoboCasa365, most failures lie at relation stages, but the most common phenomenon of RoboLab failures is a lost hold, and in RoboCasa365 it is a failure to make the contact the stage needs. Among the suites with at least 20 failures, a cognitive root cause is most frequent in Embodied Safety Exam, RoboPaint and iSpatialGym, at 28 to 38\% of their failures, compared with under 8\% in ten others, such as HumanoidBench, MyoSuite and DexToolBench.

\paragraph{By robot body.} The upper rows of \Cref{fig:fa-breakdown-robot} give the robot bodies. Each kind of body comes from a few suites: three quarters of the failures of mobile manipulators are BEHAVIOR-1K's, and all those of quadrupeds come from MuJoCo Playground. The bodies also differ in their mix of settings. These rows therefore do not isolate the effect of the body. Loss of balance is the most common phenomenon for humanoids, quadrupeds and mobile manipulators. Failure to make contact is the most common for dexterous hands and for the musculoskeletal models of MyoSuite, and most failed attempts with a dexterous hand never meet a stage of the task. For robots with one or two arms, collisions and inaccurate poses are the most common phenomena apart from task-specific defects.

\paragraph{By agent and setting.} The lower rows of \Cref{fig:fa-breakdown-robot} give the agents and the settings. For every agent, relation stages hold the largest share of the failures, from 33 to 43\%, and the shares of the other categories differ by about five points or less between agents. The agents differ in how far they get and in what goes wrong, as in \Cref{sec:f-agents}. The failures of the five weaker agents more often meet no stage of the task, and more often lack the contact that the stage needs. Gemini 4 Argon's failures more often have the not-attempted label. Between the settings, \evalsetting{Realtime} failures more often lie at competition and dynamic stages and more often involve a timing error. \evalsetting{Privileged} failures more often have the not-attempted label. Collisions are more common in \evalsetting{Standard} failures, and losses of balance are more common in \evalsetting{Standard} and \evalsetting{Realtime} failures than in \evalsetting{Privileged} ones. The settings cover different mixes of tests, so these comparisons do not isolate the effect of the setting.

\subsection{Further Statistics}
\label{app:fa-stats}

\paragraph{Failure rate of a stage.} \looseness=-1 The failure rate of a stage is the number of attempts that failed at the stage divided by the number of attempts that reached it. Attempts that never reached the stage are not counted, so the rate measures whether an attempt that gets to the stage fails there. For the rate of a stage type, each stage of that type counts separately, so an attempt that reaches three put-in stages adds three stages. A stage counts as failed when it is the attempt's failure stage, and as met when its condition held at some moment, even if it already held at the start or stopped holding later. To compare the two settings without mixing suites, we compute the rate within each suite and average over the suites in which attempts reach the stage at least eight times in each setting, weighted by the smaller of the two counts. \Cref{tab:fa-step-rates,tab:fa-contact-rates} give the results.

\begin{table}[ht]
  \centering
  \footnotesize
  \caption{Failure rate of each stage type in percent, compared within suites: the types compared in at least two suites, as in \Cref{fig:fa-transfer-a}, and the types of one suite with a weight of at least 25. P is the \evalsetting{Privileged} and S the \evalsetting{Standard} Setting. The Suites column gives the number of suites compared, and the Weight column the sum of the smaller counts. Types compared in only one suite are less reliable. Close includes stages that already hold at the start, such as a closed fridge, which count as reached even when the attempt never acts on them.}
  \label{tab:fa-step-rates}
  \begin{tabular}{@{}lrrrrr@{}}
    \toprule
    Stage type & Suites & Weight & P & S & S $-$ P \\
    \midrule
    Clean & 2 & 68 & 33 & 84 & +51 \\
    Fold & 1 & 35 & 38 & 89 & +51 \\
    Assemble & 2 & 56 & 34 & 80 & +46 \\
    Strike & 2 & 221 & 11 & 55 & +44 \\
    Perform acrobatics & 1 & 29 & 28 & 71 & +43 \\
    Flatten & 1 & 36 & 42 & 84 & +42 \\
    Adjust pose & 5 & 71 & 23 & 64 & +41 \\
    Dribble & 1 & 42 & 14 & 55 & +41 \\
    Catch & 2 & 100 & 18 & 57 & +39 \\
    Throw & 1 & 56 & 61 & 100 & +39 \\
    Insert & 2 & 46 & 25 & 62 & +36 \\
    Draw/write & 1 & 165 & 59 & 94 & +35 \\
    Jump & 2 & 134 & 36 & 66 & +30 \\
    Traverse & 2 & 95 & 27 & 54 & +27 \\
    Stack & 2 & 23 & 28 & 54 & +26 \\
    Disassemble & 2 & 47 & 44 & 70 & +26 \\
    Thread/hang & 3 & 208 & 24 & 49 & +25 \\
    Follow path & 4 & 497 & 23 & 47 & +24 \\
    Put in & 12 & 797 & 28 & 48 & +20 \\
    Open & 4 & 265 & 29 & 47 & +18 \\
    Pick up & 5 & 84 & 29 & 47 & +18 \\
    Switch on/off & 2 & 24 & 10 & 25 & +15 \\
    Place & 10 & 484 & 23 & 35 & +12 \\
    Close & 7 & 264 & 5 & 9 & +4 \\
    Press & 2 & 96 & 1 & 2 & +1 \\
    \bottomrule
  \end{tabular}
\end{table}

\begin{table}[ht]
  \centering
  \footnotesize
  \caption{Failure rate by contact in percent, compared within suites as in \Cref{tab:fa-step-rates}. Bimanual stages are the hardest in both settings, and among the values of how, the failure rates of in-hand adjustment and of cradling or scooping rise the most.}
  \label{tab:fa-contact-rates}
  \begin{tabular}{@{}llrrrrr@{}}
    \toprule
     & Value & Suites & Weight & P & S & S $-$ P \\
    \midrule
    Who & Legs/torso & 6 & 844 & 27 & 64 & +37 \\
     & Bimanual & 10 & 741 & 39 & 71 & +32 \\
     & Tool & 12 & 910 & 24 & 55 & +31 \\
     & Fingers/gripper & 26 & 3,420 & 24 & 46 & +22 \\
     & Mobile base & 6 & 1,169 & 26 & 43 & +18 \\
     & No contact & 3 & 173 & 11 & 24 & +13 \\
    \midrule
    How & In-hand adjustment & 4 & 96 & 29 & 70 & +41 \\
     & Cradle/scoop & 8 & 268 & 23 & 57 & +34 \\
     & Support & 11 & 1,956 & 27 & 52 & +25 \\
     & Avoid contact & 3 & 230 & 24 & 49 & +25 \\
     & Press/wipe & 6 & 389 & 32 & 55 & +23 \\
     & Grasp & 25 & 3,391 & 24 & 46 & +22 \\
     & Hook & 4 & 186 & 31 & 50 & +19 \\
     & Push/sweep & 16 & 640 & 17 & 31 & +14 \\
    \bottomrule
  \end{tabular}
\end{table}

\paragraph{The \evalsetting{Privileged} and \evalsetting{Standard} settings.}
\label{app:fa-modes}
\looseness=-1 The two settings differ in two conditions that matter for \Cref{sec:f-settings}. In the \evalsetting{Privileged} Setting the agent reads the exact pose of every object and can reset the simulator without limit, so it can compute a precise motion from the full state, test it and try again within the time limit before it submits an answer. In the \evalsetting{Standard} Setting it must estimate poses from camera images and depth, the episode is graded as it runs, and the task limits the number of resets, so a precise stage is attempted on estimates and an error can remain in the graded episode. The comparisons of \Cref{sec:f-settings} concern the stages of a task. They do not concern how the agent controls the robot, which changes with the setting (\Cref{sec:f-control}). We split the stage types into two groups by their definitions in the taxonomy (\Cref{tab:fa-steps}). A stage type is precise if it needs a tight fit, a mated or threaded relation, a target pose or path, or contact with a moving object. The 13 precise types are insert, assemble, screw/unscrew, thread/hang, connect, adjust pose, follow path, stabilize, regrasp, throw, strike, catch and dribble. A stage type is coarse if it needs only to reach a region, a container, a grasp or the open or closed state of a mechanism. The 12 coarse types are place, put in, take out, gather, pick up, push, pull, open, close, press, switch on/off and navigate. All other types, such as the material types and the body types other than navigate, are in neither group. \Cref{sec:f-settings}, \Cref{app:an-settings} and \Cref{tab:fa-ablation-stages} use these groups.

\looseness=-1 Compared within suites as in \Cref{tab:fa-step-rates}, the failure rate rises from 22\% to 56\% at precise stages, in 9 suites, and from 22\% to 38\% at coarse stages, in 25 suites. With suite fixed effects and intervals clustered by task, the precise stages rise by 19 points more than the coarse stages (95\% interval 13 to 25). These comparisons use different attempts in the two settings. When the same agent reaches the same stage of a task in both settings, the failure rate rises by 25 points at precise stages and by 15 points at coarse stages, and the difference between the two groups, 10 points, is smaller but significant (95\% interval 5 to 16 points). Follow-path stages, most of them waypoints of DexToolBench, make up 38\% of the precise stages in this comparison, and their rate rises by 11 points. Without them, the rate at precise stages rises by 33 points, 18 points more than at coarse stages (95\% interval 12 to 25). Among the types compared in at least two suites in \Cref{tab:fa-step-rates}, the rates of striking, assembling, adjusting a pose and catching rise by 35 to 41 points in this paired comparison, and those of closing, placing and picking up rise by 3 to 17 points.

\looseness=-1 The labels cannot show how much of the loss comes from perception and how much from control (see the limitations at the end of this section). The labels also cannot show how much comes from the missing state and how much from the missing resets. The controlled runs of \Cref{app:ablation} separate the two at the level of success rates: on these tests, adding either one alone closes a third to a half of the gap. Joined with the labels, they also separate the two by stage type.

\paragraph{State and resets at precise stages.} We join the controlled runs of GPT-6 Astra to its labeled failures in the \evalsetting{Standard} Setting, using the results of \Cref{app:ablation}. Of these failures, 256 have a located failure stage. Of them, 91 are at precise stages and 97 at coarse stages, split as above. Of the rest, 36 are at material stages and 32 at other stages. \Cref{tab:fa-ablation-stages} gives the share of them that each run solves.

\looseness=-1 Each run makes one attempt per test, so some failures would turn into successes simply by trying again. The run without the success signal approximates such a second attempt. It is not an exact repeat, since the agent must end the episode itself, and its success rate is within 4 points of the baseline's (\Cref{tab:ablation}). It solves 19\% of the failures at precise stages and 14\% of those at coarse stages.

At precise stages (\Cref{tab:fa-ablation-stages}), 22 failures are solved by exactly one of the state-query run and the reset run, and the reset run solves 18 of them (exact McNemar test, $p=0.004$).

The logs show how the agent used each feature. Of the 33 precise failures that the resets solve, 31 succeed only after a reset or a restored state. The agent did call the state query in the runs that have it. The run with both solves more of these failures than the run with resets alone (16 failures that only one of the two runs solves, 13 of them by the run with both, $p=0.021$).

Let the recovery of a run be its share of solved failures minus that of the run without the success signal, both on the same failures. On the precise failures that have a \evalsetting{Privileged} test, the recovery of the run with both features is 53\% of the recovery of the \evalsetting{Privileged} Setting, with a 95\% interval from 35\% to 70\% over failures. The \evalsetting{Privileged} Setting solves 23 more precise failures than the run with both. Stages that follow a path, thread or assemble account for 15 of these 23 failures. At striking and catching stages, the run with both solves exactly the failures that the \evalsetting{Privileged} Setting solves.

These comparisons have limits. Precise and coarse failures come from different suites. Most precise failures are in Embodied Olympics - Fun, DexToolBench and RoboARC, and 38 of the 97 coarse failures are in BEHAVIOR-1K. Leaving out any one suite reverses none of the comparisons between runs or between the two groups of stages. The intervals and tests treat the failed tests as independent.

\begin{table}[ht]
  \centering
  \footnotesize
  \caption{Share (\%) of GPT-6 Astra's labeled failures in the \evalsetting{Standard} Setting that each controlled run solves, by type of failure stage, with 95\% intervals over failures. Material stages include drawing, pouring and cleaning. The \evalsetting{Privileged} row uses the \evalsetting{Privileged} test of the same task.}
  \label{tab:fa-ablation-stages}
  \setlength{\tabcolsep}{4pt}
  \begin{tabular}{@{}lcccc@{}}
    \toprule
    Run & Precise (91) & Coarse (97) & Material (36) & Other (32) \\
    \midrule
    No success signal & 19 (11--26) & 14 (8--22) & 28 (14--42) & 16 (3--28) \\
    + object-state query & 21 (13--30) & 34 (25--43) & 42 (25--58) & 28 (13--44) \\
    + unlimited resets and save/load states & 36 (26--46) & 44 (35--54) & 47 (31--64) & 31 (16--47) \\
    + both & 47 (37--58) & 67 (58--76) & 69 (56--83) & 41 (22--56) \\
    \evalsetting{Privileged} & 73 (64--81) & 60 (51--69) & 83 (69--94) & 72 (56--88) \\
    \bottomrule
  \end{tabular}
\end{table}

\paragraph{Concentration of failure stages.} \looseness=-1 The concentration of a test measures how much the agents that failed it agree on the stage at which they failed. We measure it with the Simpson index: for a test that $n \geq 2$ agents failed at located stages, with $n_k$ of them at stage $k$, it is $D = \sum_k n_k(n_k-1) / \bigl(n(n-1)\bigr)$, the share of pairs of these agents that failed at the same stage. We average $D$ over the tests of a setting whose tasks have at least two counted stages, and resample tests 2,000 times for the 95\% interval. If each agent's failure stage is sampled independently with probabilities $q_k$, the expected value of $D$ is $\sum_k q_k^2$. The random-stage value, the gray bars of \Cref{fig:fa-where-a}, assumes that each agent fails at a random stage but keeps one property of the observed failures: where in a task they usually fall. First, we order the $K$ stages of each task graph topologically, breaking ties by their order in the task file, so that the $i$-th stage covers the interval $[(i-1)/K, i/K]$. Second, over all failures in the other tasks of the same setting that have at least two stages, we build a ten-bin histogram of the relative position $(i - 1/2)/K$ of the failure stage. Third, each stage receives the mass of this histogram over its interval, and these masses are the probabilities $q_k$. A simpler uniform baseline, $q_k = 1/K$ for a task with $K$ stages, gives 28\% in the \evalsetting{Privileged} Setting, 28\% in the \evalsetting{Standard} Setting and 30\% in the \evalsetting{Realtime} Setting. This baseline ignores that early stages fail more often, so the main text does not use it. The value of $D$ for the actual failure stages exceeds the random-stage value in 74\% of \evalsetting{Standard} and 64\% of \evalsetting{Privileged} tests, compared with about 39\% in both if the failure stages were sampled from the random-stage probabilities $q_k$. The 95\% interval of the mean difference is 21 to 28 points in the \evalsetting{Standard} Setting and 14 to 22 points in the \evalsetting{Privileged} Setting. In the \evalsetting{Realtime} Setting, which \Cref{sec:f-difficulty} excludes because only 27 tests qualify, the concentration is 53\%, compared with 43\% for random failure stages, and the interval of the difference, 1 to 20 points, lies just above zero. On tasks with at least three stages, the concentration in the \evalsetting{Standard} Setting is 50\% over 194 tests, compared with 26\% for random failure stages and 18\% for the uniform baseline. The comparison with random failure stages assumes that, apart from where in a task failures usually fall, the failure stages of different agents are independent. Because each agent makes one attempt per test, every pair counted in $D$ compares two different agents. For the three strongest agents alone, we use the \evalsetting{Standard} tests that at least two of them failed at located stages, and we build the histogram from their failures on the other tasks. Over these 217 tests, the concentration is 56\% (95\% interval 50\% to 62\%), compared with 31\% for random failure stages, and the 95\% interval of the difference is 20 to 31 points. With the histogram of the eight agents, the random-stage value is 35\%. Pairs that agree at a stage after the first contribute 28 points of the 56\%, compared with 13 for random failure stages. In 34\% of these tests, exactly two of the three agents have located failures, so the concentration on each such test is either 0 or 1. Among the located \evalsetting{Standard} failures, 62\% of those of the five weaker agents are at the first stage, compared with 41\% for the three strongest. In the \evalsetting{Privileged} Setting, 69 tests qualify, and 78\% of them have exactly two located failures among these agents. The concentration is 40\%, compared with 25\% for random failure stages, and the 95\% interval of the difference is 5 to 25 points.

\paragraph{Shared failure stages.} We count the \evalsetting{Standard} tests with at least three stages that four or more agents failed. In 64 of these 190 tests, at least 80\% of the failures fall on one stage. \Cref{tab:fa-bottlenecks} lists one example per suite among the 64. The \evalsetting{Realtime} test of the same RoboDynamicExam task as \Cref{fig:case-ball} shows the same shared stage: seven agents failed at the recatch.

\begin{table}[ht]
  \centering
  \footnotesize
  \caption{\looseness=-1 Examples of shared failure stages in the \evalsetting{Standard} Setting, one per suite. The Stages column gives the number of stages of the task. The column ``Failed here'' gives the failed attempts that failed at this stage over all failed attempts with a located failure stage, without the attempts excluded in \Cref{app:fa-exclusions}.}
  \label{tab:fa-bottlenecks}
  \begin{tabularx}{\linewidth}{@{}l>{\raggedright\arraybackslash}Xr>{\raggedright\arraybackslash}Xlr@{}}
    \toprule
    Suite & Task & Stages & Shared stage & Type & Failed here \\
    \midrule
    BEHAVIOR-1K & clean up branches and twigs & 5 & first branch into the bin & put in & 8/8 \\
    Embodied Safety Exam & three spice jars onto a tray & 3 & second jar onto the tray & place & 8/8 \\
    RoboARC & thread a fiber bundle through rings around a post & 18 & first fiber through its ring & thread/hang & 7/8 \\
    RoboCasa365 & defrost meat in the microwave & 4 & container into the microwave & put in & 7/7 \\
    DexHandExam & install the front guard of a fan & 5 & seat the front guard & assemble & 6/7 \\
    RoboDynamicExam & catch, return through a slot, deliver to a moving cart & 3 & recatch after the rebound & catch & 7/8 \\
    RoboDojo & put two slices of bread into a toaster and press the lever & 3 & first slice into a toaster slot & put in & 6/7 \\
    \bottomrule
  \end{tabularx}
\end{table}

\paragraph{The common failure stage.} \looseness=-1 A test has a single most common failure stage when one stage holds more of its located failures than any other. In all, 203 \evalsetting{Privileged} and 264 \evalsetting{Standard} tests have one. For the statements of \Cref{sec:f-difficulty} built on it, the stages of a task are ordered by their depth in the task graph, with ties broken by the order in the task file. To ask whether stage types predict this stage, we estimate, for each setting, how often each stage type fails once an attempt reaches it, using the tasks other than the test's own. Each rate is shrunk toward the setting's mean rate with the weight of ten attempts. A second variant uses the rate of each type within the test's own suite, shrunk toward the pooled rate of the type in the same way. In a sequential model with failure rates $h_1, \dots, h_K$ in stage order, the chance that the failure stage is stage $k$ is proportional to $h_k \prod_{j<k}(1-h_j)$. The predicted stage is the one with the highest chance, and the concentration expected under these rates is $\sum_k q_k^2$ for the normalized chances $q_k$. To measure how strongly getting past the common stage predicts passing the test, we take for each attempt the most common failure stage of the other agents' failed attempts on the same test, when at least two of them failed there and no other stage ties with it. We count the attempt as past this stage when its progress map shows the stage met at some moment. Two agents share a failure phenomenon when their labels name at least one phenomenon in common, and pairs in which either attempt has no phenomenon are excluded. Agents that fail at the same stage need not fail there in the same way: in the \evalsetting{Standard} Setting, two agents that failed at the same stage share a failure phenomenon in 40\% of pairs, as in the RoboDynamicExam task of \Cref{fig:case-ball} (\Cref{app:fa-cases}).

\looseness=-1 \Cref{tab:fa-common-stage} compares the common failure stage with what the stage types predict, with rates from the other tasks and with rates from the same suite, and \Cref{sec:f-difficulty} reports the rates from the other tasks. A prediction from the stage types finds the common stage in 73\% of the tests, and always naming the first stage finds it in 65\%. On tasks with at least two types of stages, the two rules find it in 67\% and 56\%. The concentration it predicts is 36\% in the \evalsetting{Privileged} and 48\% in the \evalsetting{Standard} Setting, against 50\% and 61\% observed. \Cref{tab:fa-past-common} shows how strongly getting past the common stage predicts passing the test. Part of this link follows from the order of the stages, since an attempt that reached more stages is closer to the end. The settings still differ: in the \evalsetting{Privileged} Setting most attempts that pass the common stage go on to pass the test, while in the \evalsetting{Standard} Setting fewer than half do, and most of the others fail at a later stage.

\begin{table}[ht]
  \centering
  \footnotesize
  \caption{\looseness=-1 Stage types and the common failure stage, in percent: how often a rule names the most common failure stage of a \evalsetting{Standard} test, over the 264 \evalsetting{Standard} tests that have one and over those among them whose tasks have at least two types of stages, and the concentration of failure stages, observed and expected if each agent failed independently at the given rates.}
  \label{tab:fa-common-stage}
  \setlength{\tabcolsep}{4pt}
  \begin{minipage}[t]{0.48\linewidth}
    \centering
    \begin{tabular}{@{}lrr@{}}
      \toprule
      Naming rule & All & \makecell[r]{Two or more\\stage types} \\
      \midrule
      Stage-type rates, other tasks & 73 & 67 \\
      Stage-type rates, same suite & 75 & 71 \\
      Always the first stage & 65 & 56 \\
      \bottomrule
    \end{tabular}
  \end{minipage}\hfill
  \begin{minipage}[t]{0.5\linewidth}
    \centering
    \begin{tabular}{@{}lrr@{}}
      \toprule
      Concentration & \evalsetting{Privileged} & \evalsetting{Standard} \\
      \midrule
      Observed & 50 & 61 \\
      Stage-type rates, other tasks & 36 & 48 \\
      Stage-type rates, same suite & 40 & 51 \\
      Random failure stages & 32 & 37 \\
      \bottomrule
    \end{tabular}
  \end{minipage}
\end{table}

\begin{table}[ht]
  \centering
  \footnotesize
  \caption{Getting past the common failure stage, in percent. For each attempt, the common stage is the most common failure stage of the other agents on the same test. The table covers the 1,235 \evalsetting{Privileged} and 1,874 \evalsetting{Standard} attempts that have such a stage and a readable status at it.}
  \label{tab:fa-past-common}
  \begin{tabular}{@{}lrr@{}}
    \toprule
     & \evalsetting{Privileged} & \evalsetting{Standard} \\
    \midrule
    Passed the test, all attempts covered & 30 & 10 \\
    Passed the test, attempts past the common stage & 64 & 33 \\
    Passed the test, past a common stage that is not the first & 76 & 41 \\
    Failed exactly at the common stage, attempts that never met it & 77 & 83 \\
    \bottomrule
  \end{tabular}
\end{table}

\paragraph{The common failure stage across settings.} In all, 188 tasks have a single most common failure stage in both the \evalsetting{Privileged} and the \evalsetting{Standard} Setting, and it is the same stage in 152 of them, or 81\%. For a baseline, we sample, for each of these tasks and each setting, as many failure stages as the task has located failures in that setting. The samples are independent, with the stage probabilities $q^{\mathrm{P}}_k$ and $q^{\mathrm{S}}_k$ of the random-failure-stage value above and the stages in depth order. We keep the samples in which both settings have a single most common stage. Averaged over the tasks, the two most common stages then agree with probability 49\%. In 105 of the 188 tasks the common stage is the first stage in both settings, so much of the agreement comes from tasks that fail at their first stage. On the other 83 tasks the common stage is still the same in 57\% (95\% interval 46\% to 67\% over tasks). Under the same baseline, leaving out the samples in which both common stages are the first stage, the average over these 83 tasks is 9\%. Where the two common stages differ, the one of the \evalsetting{Standard} Setting lies earlier in the task in 25 tasks and later in 11. These counts describe where failures fall. They do not show that an agent follows the same plan in the two settings.

\paragraph{Pooled failure rates.} \Cref{tab:fa-pooled-rates} gives the rates behind \Cref{fig:fa-taxonomy-a}, for each setting and for both together: every attempt that reached a stage counts, whatever its suite. The stage categories are pooled over all their stage types, and the rows list the stage types found in at least three suites and reached at least 30 times in the two settings together.

\begin{table}[ht]
  \centering
  \footnotesize
  \setlength{\tabcolsep}{4pt}
  \caption{Failure rate (\%) of each stage category and of the stage types found in at least three suites, in the \evalsetting{Privileged} Setting, in the \evalsetting{Standard} Setting and in both together, with the stages at which attempts failed over the stages that attempts reached. The Suites column gives the number of suites with the stage type.}
  \label{tab:fa-pooled-rates}
  \begin{tabular}{@{}llrrrr@{}}
    \toprule
    Category & Stage type & Suites & \evalsetting{Privileged} & \evalsetting{Standard} & Both \\
    \midrule
    Mechanism & all & 18 & 14 (124/856) & 27 (220/825) & 20 (344/1,681) \\
     & Press & 6 & 2 (3/179) & 6 (7/114) & 3 (10/293) \\
     & Close & 9 & 5 (15/287) & 8 (25/305) & 7 (40/592) \\
     & Switch on/off & 4 & 11 (5/44) & 24 (8/33) & 17 (13/77) \\
     & Open & 12 & 29 (92/321) & 47 (164/350) & 38 (256/671) \\
    \midrule
    Pose & all & 16 & 16 (253/1,592) & 51 (389/763) & 27 (642/2,355) \\
     & Follow path & 6 & 14 (191/1,334) & 50 (267/534) & 25 (458/1,868) \\
     & Pick up & 10 & 24 (28/116) & 42 (51/121) & 33 (79/237) \\
     & Adjust pose & 9 & 24 (29/123) & 67 (65/97) & 43 (94/220) \\
    \midrule
    Dynamic & all & 3 & 17 (121/699) & 62 (266/430) & 34 (387/1,129) \\
     & Strike & 3 & 12 (50/418) & 56 (128/229) & 28 (178/647) \\
    \midrule
    Relation & all & 26 & 26 (615/2,356) & 45 (1,051/2,350) & 35 (1,666/4,706) \\
     & Place & 19 & 23 (150/651) & 32 (289/895) & 28 (439/1,546) \\
     & Thread/hang & 6 & 22 (69/311) & 49 (113/229) & 34 (182/540) \\
     & Put in & 20 & 28 (274/994) & 47 (426/905) & 37 (700/1,899) \\
     & Screw/unscrew & 3 & 22 (14/64) & 81 (26/32) & 42 (40/96) \\
     & Insert & 5 & 23 (16/69) & 64 (47/74) & 44 (63/143) \\
     & Stack & 4 & 25 (10/40) & 66 (25/38) & 45 (35/78) \\
    \midrule
    Competition & all & 2 & 18 (26/141) & 50 (67/133) & 34 (93/274) \\
    \midrule
    Body & all & 7 & 28 (143/504) & 67 (267/401) & 45 (410/905) \\
     & Traverse & 4 & 22 (43/194) & 58 (62/106) & 35 (105/300) \\
     & Navigate & 5 & 33 (8/24) & 68 (52/77) & 59 (60/101) \\
    \midrule
    Material & all & 7 & 44 (189/433) & 84 (447/535) & 66 (636/968) \\
     & Clean & 3 & 33 (28/84) & 82 (62/76) & 56 (90/160) \\
     & Pour & 3 & 25 (3/12) & 72 (113/156) & 69 (116/168) \\
    \bottomrule
  \end{tabular}
\end{table}

\begin{figure}[t]
  \centering
  \includegraphics[width=0.8\linewidth]{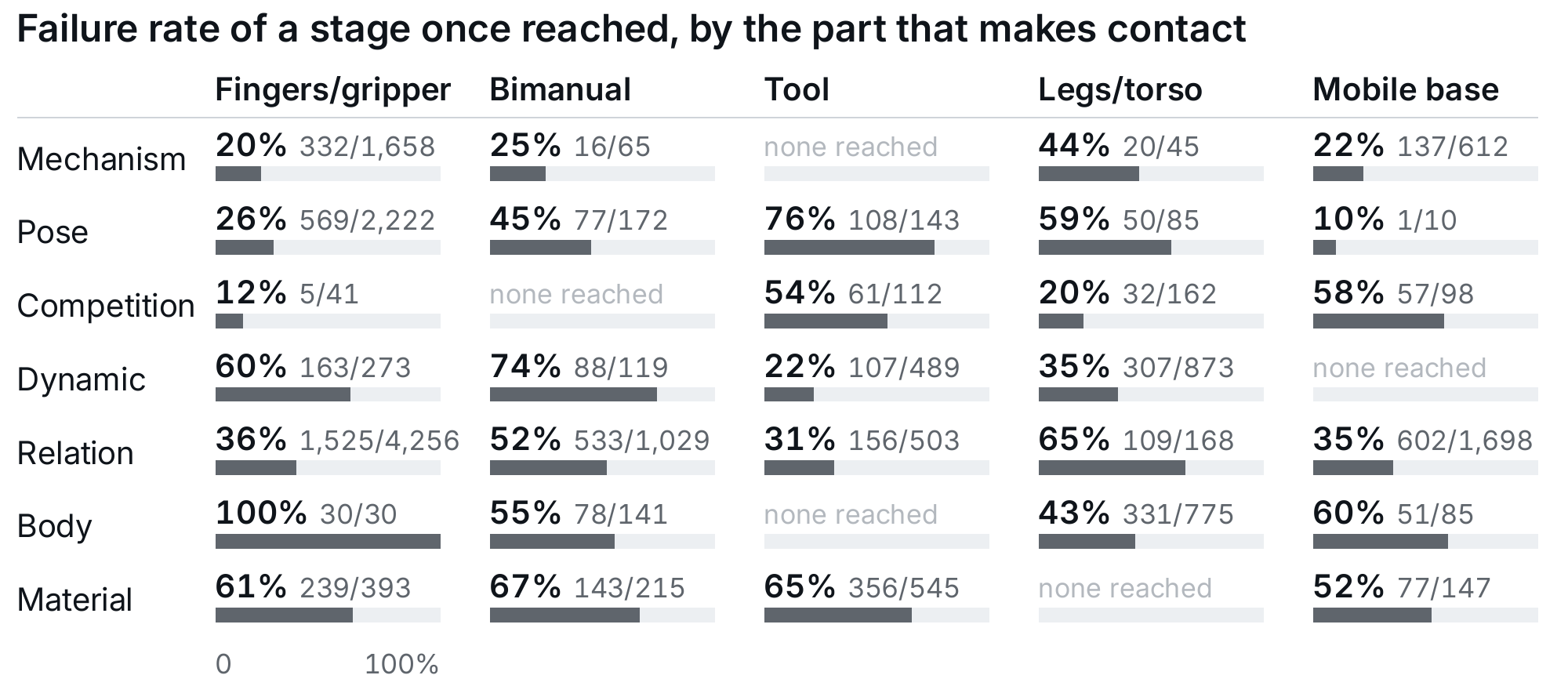}
  \caption{Failure rate of a stage once reached, by stage category and by the robot part that makes contact, with the \evalsetting{Privileged} and \evalsetting{Standard} settings pooled (\Cref{tab:fa-pooled-rates}). Each cell gives the rate, the stages failed over the stages reached, and a bar on a scale from 0 to 100\%. The rows are those of \Cref{fig:fa-taxonomy-a}, at the same heights.}
  \label{fig:fa-taxonomy-b}
\end{figure}

\paragraph{Interpreting the stage rates.} The rates of \Cref{fig:fa-taxonomy-a,fig:fa-taxonomy-b} show an association between kinds of stages and failure. They cannot be read as causal effects, for three reasons. First, each kind of stage occurs mostly in particular suites and with particular robots: the part that makes contact depends on the robot, so legs and torso belong to legged robots, and most material stages come from RoboPaint and Embodied Safety Exam. Second, stages differ in how much the robot must do in them. A relation stage usually covers grasping, carrying and releasing its object, because the task graphs list picking up as a stage of its own in only 11 of the 229 tasks with relation stages. A rate per stage therefore compares units of different size. Third, the taxonomy records what a stage changes but not how much contact it needs. Explaining the rates by contact therefore requires a judgment of what each kind of stage asks of the robot. Opening a door or a lid fails at 38\% of the stages that attempts reach. Closing one fails at 19\% when it does not already hold at the start, and at 7\% if the doors and fridges that are closed from the start count too. Relation stages fail at 36\% of the stages that attempts reach with one hand or gripper, at 52\% with two arms and at 65\% with the legs or torso. Rates with tool contact can be lower or higher than with direct contact: stages on moving objects fail less often when the robot acts through a cup or a racket, while pose stages fail more often when it moves an object through a tool. The same mode of contact can be easy or hard: stages done by pressing or wiping fail at 8\% of the stages that attempts reach when they press on a mechanism, such as a button, and at 69\% when they wipe or press on a material. The pose, relation and dynamic categories lie between these extremes, and their order differs between the two settings, so \Cref{fig:fa-taxonomy-a}, which counts both settings together, does not rank these three reliably.

\paragraph{What predicts the failure stage.} We predict whether an attempt that reaches a stage fails there from the setting plus one other attribute, such as the stage type or the contact. We use five-fold cross-validation over tasks, so that the tasks of the held-out fold are not seen in training, and report the relative drop in log loss compared with the setting alone. The stage type gives 7.4\%, contact as who $\times$ how 6.9\%, the stage category 5.6\%, the suite 3.9\%, how alone 3.8\%, the robot's body, grouped into eight kinds, 2.8\% and who alone 1.8\%. The agent gives 13\%. With the setting and the suite as the baseline, the stage type adds 8.2\%, who $\times$ how 7.4\%, the category 6.6\%, how 5.6\% and who 1.6\%. To test whether difficulty rankings hold across suites, we hold out one suite at a time, rank each pair of values by their failure rates on the other suites, where each value must be reached at least 20 times, and check the ranking inside the held-out suite, where it must be reached at least 10 times. The rankings agree in 55\% of 313 pairs for stage types, 53\% of 150 pairs for how, and 66\% of 76 pairs for stage categories.

\paragraph{Failure profiles of the agents.} \Cref{tab:fa-agents} gives the share of each agent's failed attempts in the \evalsetting{Standard} and \evalsetting{Realtime} settings with each label. A failed attempt can have two phenomena, so the columns do not sum to 100\%. The last two rows compare each agent with GPT-6 Astra on the tests with at least three stages that both failed, with ties excluded. Ties are common. Of the 145 tests that GPT-6 Astra and Claude Opus 5.5 both failed with readable progress, GPT-6 Astra reached more stages on 68 tests, Claude Opus 5.5 on 19, and the two tie on 58. Between the other two strongest agents, Claude Opus 5.5 reached more stages than Gemini 4 Argon in 51\% of the 68 such tests. In the \evalsetting{Standard} Setting, two of the three strongest agents that failed the same test failed at the same stage in 55\% of pairs pooled over tests, and two of the five weaker agents did so in 72\%. When they did, that stage came after the first stage in 51\% of the 277 such pairs of the strongest agents and in 27\% of the 1,700 such pairs of the weaker agents.

The shares of failed attempts depend on how far an agent gets: an agent that reaches more stages that need a hold has more chances to lose one. The two rows ``per stage reached'' of \Cref{tab:fa-agents} therefore divide by the stages that the attempts reached, using the attempts with readable stage progress. Of the reached stages that need contact, the weaker agents fail without contact at 9.1 to 18.5\%, and the strongest agents do so at 2.7 to 3.7\%. The pooled difference, 11 percentage points, has a 95\% interval from 9 to 13 points over tasks, and the weaker agents' rate is the higher one in all 19 suites with at least 20 such stages per group. Over the five pairs of GPT-6 Astra with a weaker agent, on the tests that both failed, 9 to 10\% of GPT-6 Astra's failures lack contact, compared with 19 to 25\% of the weaker agent's. Lost holds show a different pattern. They make up more of GPT-6 Astra's failed attempts than of any other agent's, but per stage reached that needs a grasp, GPT-6 Astra loses the hold in 5.6\% of cases, compared with 7.5\% for the other seven agents pooled, and the 95\% interval of the difference over tasks is $-0.2$ to 3.6 points. Per stage reached, GPT-6 Astra's observed rate is therefore lower than that of the other agents pooled, although the interval includes zero. Among single agents, only Gemini 4 Argon has a lower rate. Constraint violations, too, make up more of GPT-6 Astra's failures than of any other agent's. Per 100 attempts, Qwen3.8 Max has 4.5 and GPT-6 Astra 4.4, so per attempt the two are about equal.

\begin{table}[ht]
  \centering
  \footnotesize
  \setlength{\tabcolsep}{3pt}
  \caption{Failure profile of each agent, given as the share in percent of its failed attempts in the \evalsetting{Standard} and \evalsetting{Realtime} settings with each label. Zero progress, an attempt that met no stage of the task, counts only \evalsetting{Standard} attempts on tasks with at least two stages whose progress shows whether they met a stage. Of each agent's failed \evalsetting{Standard} attempts on such tasks, 3 to 33\% have no readable progress and are left out, the most for Qwen3.8 Max. The highest value of each row is in bold. The rows from ``Failed to make contact'' to ``Not attempted'' are phenomena, that row gives the position of the failure stage, and the two rows after it are root causes. The two rows ``per stage reached'' divide by stages instead of failed attempts: among the reached stages that need contact (or a grasp), the share at which the attempt failed without contact (or with a lost hold). The last two rows compare each agent with GPT-6 Astra on the tests with at least three stages that both failed, excluding ties: the share of these tests on which GPT-6 Astra reached more stages, and their number.}
  \label{tab:fa-agents}
  \begin{tabular}{@{}lrrrrrrrr@{}}
    \toprule
     & Astra & Opus & Argon & Kimi & G-3.8 Flash & Qwen & DeepSeek & GLM \\
    \midrule
    Failed attempts & 316 & 372 & 328 & 432 & 449 & 433 & 467 & 472 \\
    \midrule
    Zero progress, \evalsetting{Standard} & 27 & 36 & 32 & 44 & 47 & \textbf{64} & 59 & 59 \\
    Failed to make contact & 9.8 & 7.3 & 10.1 & 16.7 & 18.7 & 20.3 & 21.2 & \textbf{24.2} \\
    Lost hold & \textbf{17.4} & 12.6 & 10.1 & 12.5 & 10.5 & 7.2 & 9.4 & 10.8 \\
    Inaccurate pose & 20.9 & 20.4 & 23.5 & \textbf{25.2} & 21.2 & 11.1 & 12.8 & 14.2 \\
    Obstruction/collision & 17.1 & 15.1 & 8.2 & 6.5 & 10.0 & 11.5 & \textbf{18.8} & 12.9 \\
    Constraint violation & \textbf{6.6} & 4.3 & 4.3 & 3.0 & 1.8 & 4.6 & 3.6 & 3.4 \\
    Loss of balance & 12.7 & 17.2 & 9.8 & 13.0 & 12.7 & 19.9 & \textbf{20.1} & 18.9 \\
    Not attempted & 0.3 & 2.7 & \textbf{10.7} & 0.5 & 3.1 & 8.8 & 4.5 & 3.8 \\
    Failed while exploring & 3.2 & 7.3 & 4.3 & 1.6 & 3.8 & 8.1 & \textbf{13.7} & 7.6 \\
    False completion belief & 0.6 & 0.3 & 0.9 & 0.9 & \textbf{7.6} & 1.6 & 1.3 & 3.2 \\
    Feedback control or trajectory/IK & 4.7 & \textbf{14.0} & 3.7 & 10.2 & 4.7 & 9.7 & 12.0 & 8.1 \\
    \midrule
    No contact, per contact stage reached & 2.8 & 2.7 & 3.7 & 9.1 & 11.9 & 17.2 & 16.0 & \textbf{18.5} \\
    Lost hold, per grasp stage reached & 5.6 & 6.5 & 4.4 & 8.1 & 8.1 & 8.4 & 8.9 & \textbf{9.8} \\
    \midrule
    GPT-6 Astra reached more stages & -- & 78 & 70 & 92 & 95 & 100 & 99 & 98 \\
    Tests compared & -- & 87 & 61 & 95 & 92 & 79 & 120 & 111 \\
    \bottomrule
  \end{tabular}
\end{table}

\paragraph{Limitations of the labels.} The labels used here were written and reviewed by agents (\Cref{app:fa-procedure}), and GPT-6.1 Sol, which wrote most of them, is from the same family as GPT-6 Astra. Labels come from the replay video and the logs, without the simulator state, so a wrong estimate and a wrong motion often cannot be distinguished. Perception is named in 4\% of \evalsetting{Standard} failures, and 47\% of the analyzed failures have a root cause given at the category level, ``Control''. The root-cause layer therefore supports few conclusions, and the analyses of \Cref{sec:f-settings,sec:f-difficulty,sec:f-agents} rest mainly on failure stages, stage types and phenomena. Claude Sonnet 5.5 labeled all Kimi 3.1 Pro attempts and 629 of the analyzed attempts of the other seven agents. The two label writers leave the root cause empty at different rates, so the root-cause shares in \Cref{fig:fa-breakdown-robot,tab:fa-agents} should not be used for comparisons between agents. Each agent makes one attempt per test, so per-agent shares on a few dozen tests have wide uncertainty.

\subsection{Case Studies}
\label{app:fa-cases}

\looseness=-1 This section gives in full the attempts behind the examples of \Cref{sec:experiments}, each with its figure. Times are seconds on the replay video of the graded episode.

\paragraph{Carrying in groceries (\Cref{fig:fa-method}).} In this BEHAVIOR-1K task a mobile two-armed robot must carry a tomato and a carton of milk from a car into the fridge and then close the car and the fridge. The robot must open the garage door and the fridge before either food can go in, and close the fridge only after both are in, while closing the car does not depend on the other stages. Stage 5, the closed fridge, is met from the start and stops holding when the robot opens the fridge. GPT-6 Astra in the \evalsetting{Privileged} Setting reaches all six stages and passes. Claude Opus 5.5 in the \evalsetting{Standard} Setting opens the garage door and the fridge, closes and reopens the fridge twice, puts the tomato in and falls while carrying the milk through the door, so its failure stage is stage 4. DeepSeek V4.1 Flash in the \evalsetting{Standard} Setting lifts the tomato but drops it, so it fails at stage 3, and only the closed fridge is still met.

\begin{figure}[t]
  \centering
  \includegraphics[width=\linewidth]{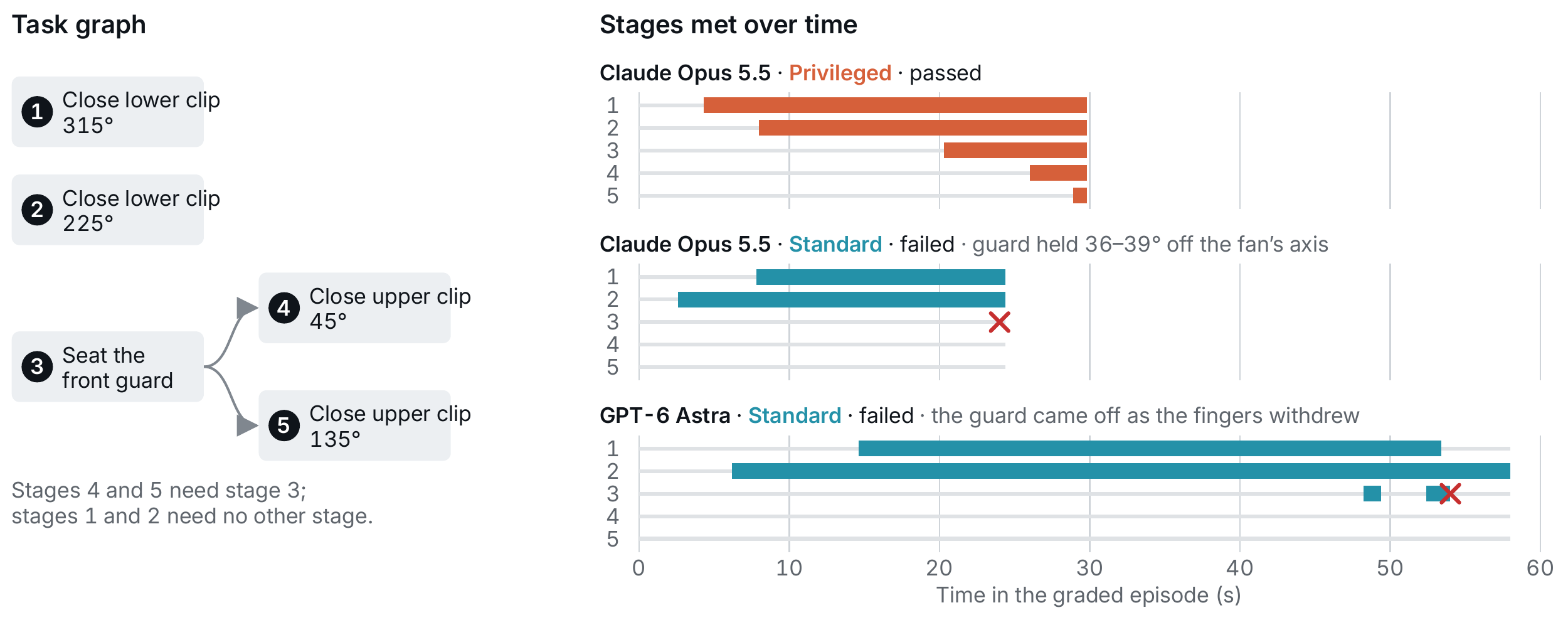}
  \caption{Task graph of a DexHandExam task, installing the front guard of a fan, and the stages that three attempts met over time. A bar shows when a stage's condition held, and a cross marks the failure stage.}
  \label{fig:case-fan}
\end{figure}

\begin{figure}[t]
  \centering
  \includegraphics[width=\linewidth]{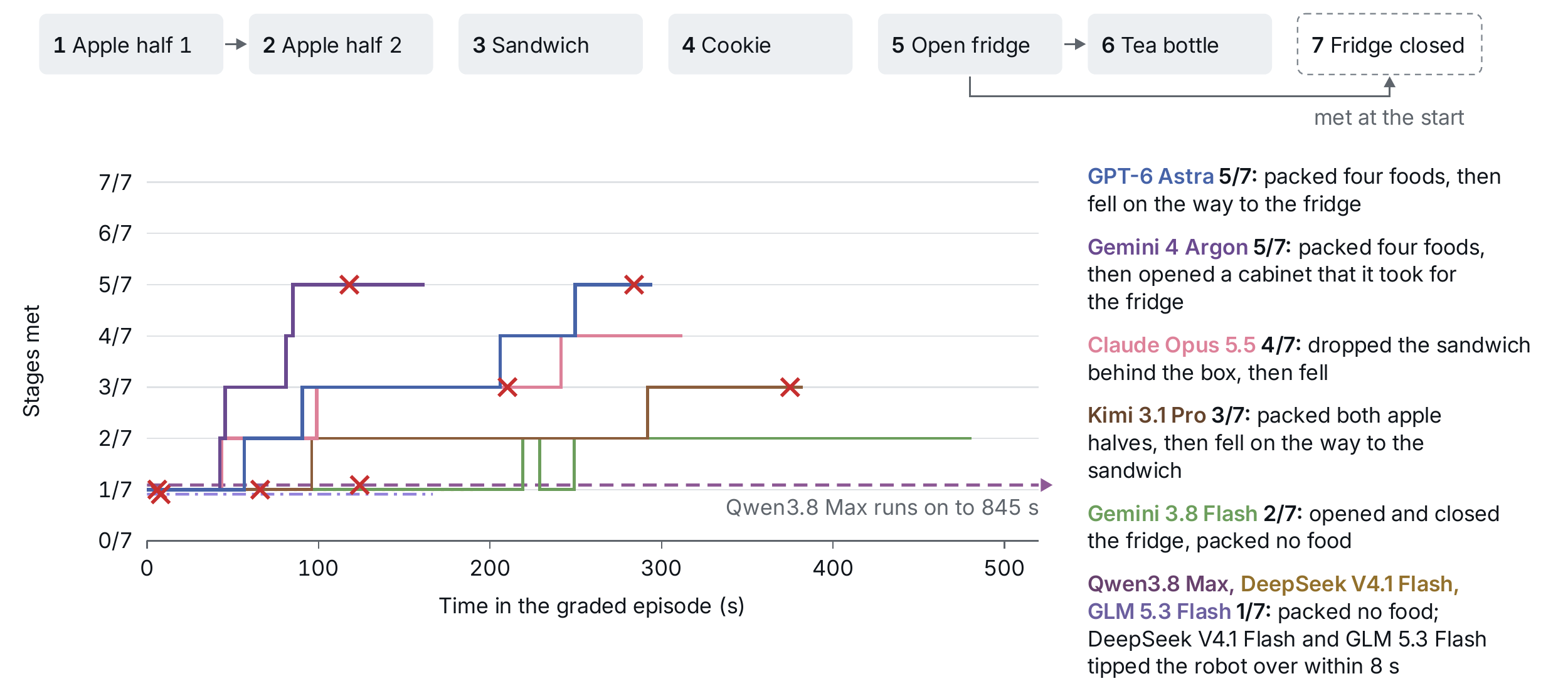}
  \caption{Progress maps of the eight agents on \emph{preparing a lunch box} (BEHAVIOR-1K, \evalsetting{Standard} Setting). Crosses mark the failure stages.}
  \label{fig:case-lunch}
\end{figure}

\paragraph{Installing a fan guard (\Cref{fig:case-fan}).} In this DexHandExam task a dexterous hand must seat the front guard of a fan on its rear guard and close four clips. The two upper clips need the seated guard, and the two lower clips do not. In the \evalsetting{Privileged} Setting, Claude Opus 5.5 closed the lower clips, seated the guard at 20 s and closed the upper clips. In the \evalsetting{Standard} Setting it closed the lower clips in the other order, but it held the guard 36 to 39 degrees off the fan's axis, whereas seating allows at most 5 degrees. When it pushed the guard on, the fan moved, and the simulator service ended the episode at 24 s. The labels attribute the failure to a wrong estimate of the guard's pose in the hand. In the same \evalsetting{Standard} test, GPT-6 Astra seated the guard twice, each time for less than 2 s. As the hand withdrew its fingers, the guard came off with it and reopened a lower clip, and the attempt ended when the guard dropped.

\paragraph{Catching a ball twice (\Cref{fig:case-ball}).} In this RoboDynamicExam task the robot must catch a thrown ball in a cup, pour it into a slot, catch it again after it rolls down and rebounds off a ramp, then pour it into a moving cart. In the \evalsetting{Standard} Setting, all eight agents failed, and seven of them caught the thrown ball and failed at the recatch stage. In four attempts, the episode ended because the cup touched the slot or the rebound ramp, which the task forbids. GPT-6 Astra, Gemini 4 Argon, Gemini 3.8 Flash and Qwen3.8 Max hit them while pouring the ball out or moving the empty cup downstream. The cup of Claude Opus 5.5 was still too high when the ball came off the ramp. Claude Opus 5.5 reported that the ball can be caught for only about 1.3 to 1.6 s after it leaves the cup, while straightening the cup after the pour alone takes about 1.4 s. In the runs of DeepSeek V4.1 Flash and GLM 5.3 Flash the ball never left the tilted cup. GLM 5.3 Flash then carried the ball to the cart, where it stayed in the cup again. Kimi 3.1 Pro did not move the arm, and its episode ended after about 1 s, so it failed at the first catch. In the \evalsetting{Privileged} Setting, GPT-6 Astra, Claude Opus 5.5 and Kimi 3.1 Pro completed all three stages, and Gemini 4 Argon, Gemini 3.8 Flash and Qwen3.8 Max submitted no answer.

\paragraph{Preparing a lunch box (\Cref{fig:case-lunch}).} In this BEHAVIOR-1K task a mobile two-armed robot must put two apple halves, a sandwich and a cookie from a cutting board, and a tea bottle from the closed fridge, into a lunch box, and leave the fridge closed. The closed fridge is met from the start. All eight agents failed the \evalsetting{Standard} test. GPT-6 Astra put both apple halves, the cookie and the sandwich into the box by 250 s. Driving to the fridge with empty hands, the robot tipped over at 284 s, and the episode ended with the fall. Gemini 4 Argon packed the same four foods by 85 s, then pulled open a tall dark cabinet that it had mistaken for the fridge, and it had not opened the fridge when the episode ended. The labels name a perception error. Claude Opus 5.5 packed an apple half and the cookie. At 210 s, the sandwich slipped from the gripper and landed behind the box, and placing the sandwich is its failure stage. After packing the second apple half at 242 s, the robot tipped over on its way to the fridge. Kimi 3.1 Pro packed the two apple halves by 292 s, and the robot tipped over at about 380 s as it went for the sandwich. Gemini 3.8 Flash opened and closed the fridge but never lifted an apple half off the board, and Qwen3.8 Max never grasped one in its 845 s episode. DeepSeek V4.1 Flash and GLM 5.3 Flash tipped the robot over within 8 s while exploring the kitchen and packed nothing. No agent passed this task in the \evalsetting{Privileged} Setting either, where the three strongest agents reached five or six of the seven stages.

\begin{figure}[tp]
  \centering
  \begin{subfigure}[t]{\linewidth}
    \includegraphics[width=\linewidth]{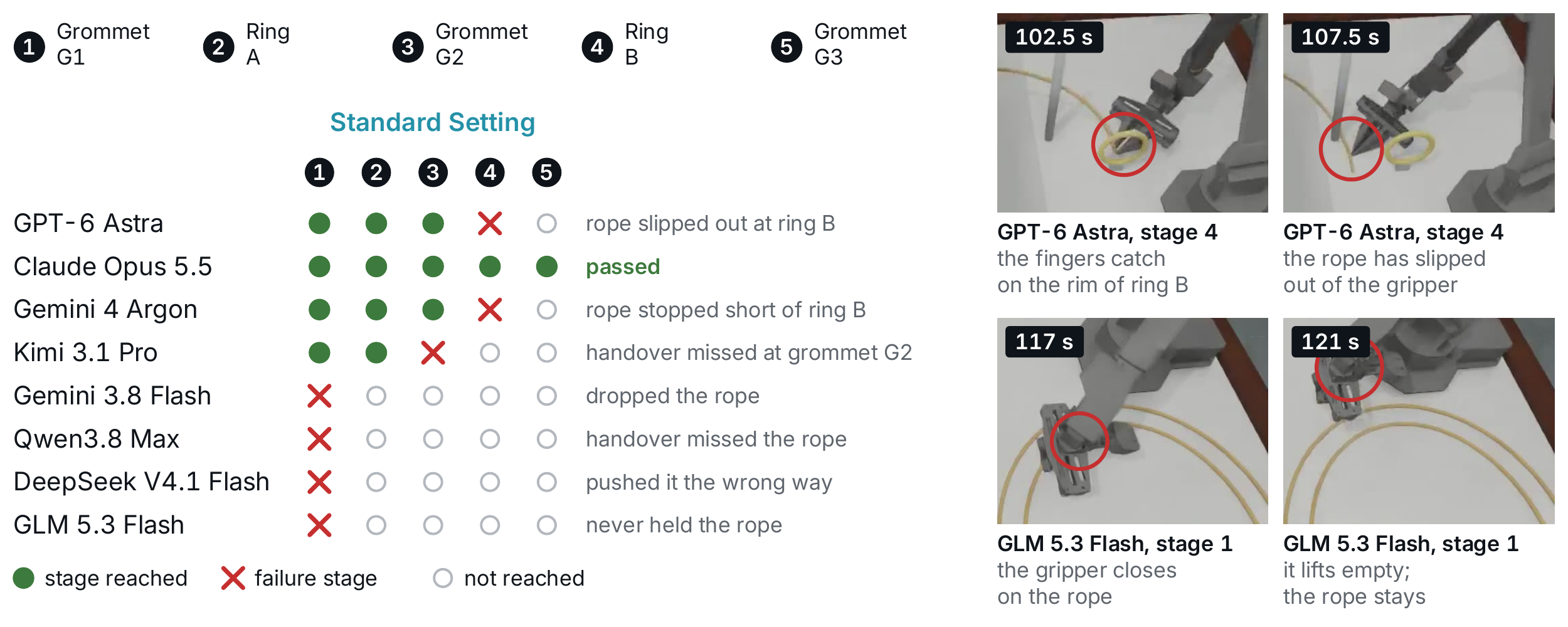}
    \caption{}
    \label{fig:fa-case-rope}
  \end{subfigure}

  \vspace{0.8em}
  \begin{subfigure}[t]{\linewidth}
    \includegraphics[width=\linewidth]{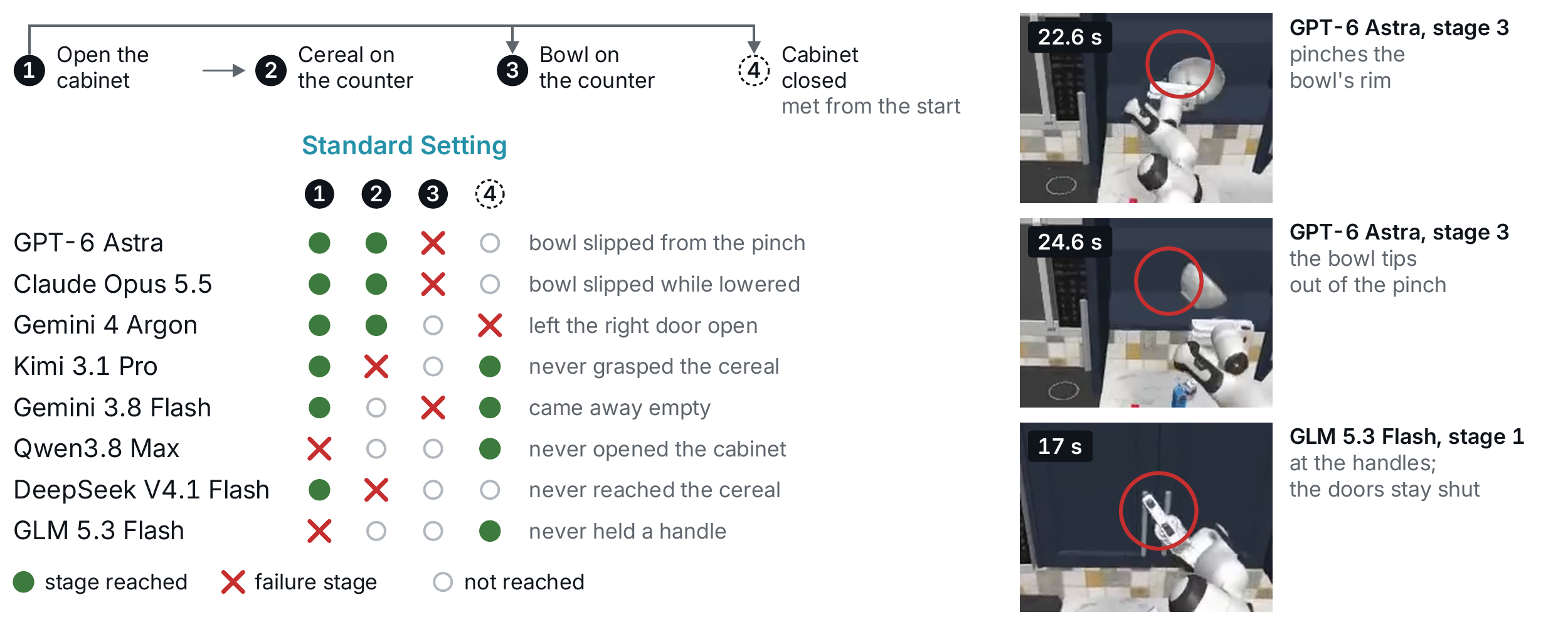}
    \caption{}
    \label{fig:fa-case-cereal}
  \end{subfigure}
  \caption{\evalsetting{Standard} Setting. (a) RoboARC, \emph{threading a rope}. (b) RoboCasa365, \emph{fetching cereal and a bowl}; the column of stage 4 shows the end of the episode.}
  \label{fig:fa-case-f8}
\end{figure}

\paragraph{Threading a rope through five passages (\Cref{sec:f-agents}, \Cref{fig:fa-case-rope}).} In this RoboARC task two arms with parallel grippers must thread a rope, in order, through a grommet in a wall, a ring, a second grommet, a second ring and a third grommet. Each passage is a stage. In the \evalsetting{Standard} Setting, Claude Opus 5.5 passed. GPT-6 Astra threaded the first three passages by 100 s. At the second ring, the fingers caught on the rim. At 106 s, the rope slipped out of the gripper, and the time limit ended the episode soon after. Gemini 4 Argon also threaded the first three passages, by about 55 s, but the rope never went through the second ring. Four of the five weaker agents did not get the rope through the first grommet, and two of them never got the hold they needed. GLM 5.3 Flash closed its gripper on the rope at 116 s and again later, and each time lifted the gripper without the rope. Qwen3.8 Max pushed the rope end into the grommet, but at 115 s the gripper on the other side closed without catching it, and the rope slid back. Gemini 3.8 Flash lifted the rope and dropped it at 16 s, and DeepSeek V4.1 Flash tried to thread the first grommet in the wrong direction. Kimi 3.1 Pro threaded the first grommet and the first ring by 395 s, and at the second grommet the gripper closed without catching the rope.

\paragraph{Fetching cereal and a bowl (\Cref{sec:f-agents}, \Cref{fig:fa-case-cereal}).} In this RoboCasa365 task a one-armed robot on a mobile base must open a cabinet, put a cereal box and a bowl from it on the counter, and close the cabinet. The closed cabinet is met from the start. In the \evalsetting{Standard} Setting, GPT-6 Astra and Claude Opus 5.5 opened the cabinet, put the cereal on the counter and failed at the bowl after holding it. GPT-6 Astra tried to pick up the bowl by its rim from 16 s. At 23 s, the bowl slipped out of the pinch and stayed on the shelf, and later pinches tipped it onto its side. Claude Opus 5.5 lifted the bowl out of the cabinet, and at 30 s, as the arm lowered it, the bowl slipped from the fingers and fell out of reach. Gemini 4 Argon put the cereal on the counter and carried the bowl out of the cabinet but did not put it on the counter, and it left the right door open. Four of the five weaker agents failed without the contact that their failure stage needed. Kimi 3.1 Pro opened the cabinet but never got hold of the cereal box. Gemini 3.8 Flash closed its gripper at the bowl at 2 s and came away empty. GLM 5.3 Flash never got hold of a door handle, and DeepSeek V4.1 Flash opened a door but never reached the cereal box. Qwen3.8 Max never opened the cabinet.

\begin{figure}[t]
  \centering
  \includegraphics[width=\linewidth]{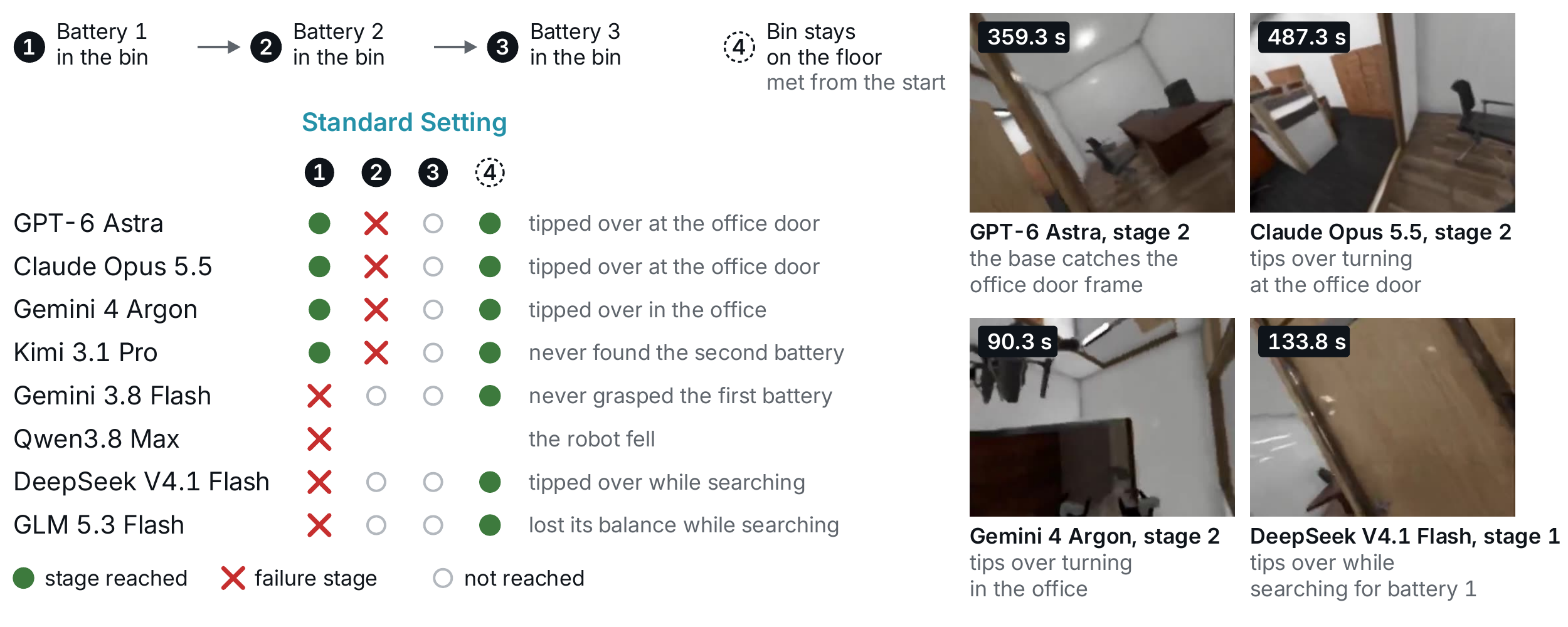}
  \caption{BEHAVIOR-1K, \emph{disposing of batteries}, \evalsetting{Standard} Setting: the task graph and the stages of each agent, where the wastebasket on the floor holds from the start, and head-camera frames as the robot tips over.}
  \label{fig:fa-case-batteries}
\end{figure}

\begin{figure}[tp]
  \centering
  \begin{subfigure}[t]{\linewidth}
    \includegraphics[width=\linewidth]{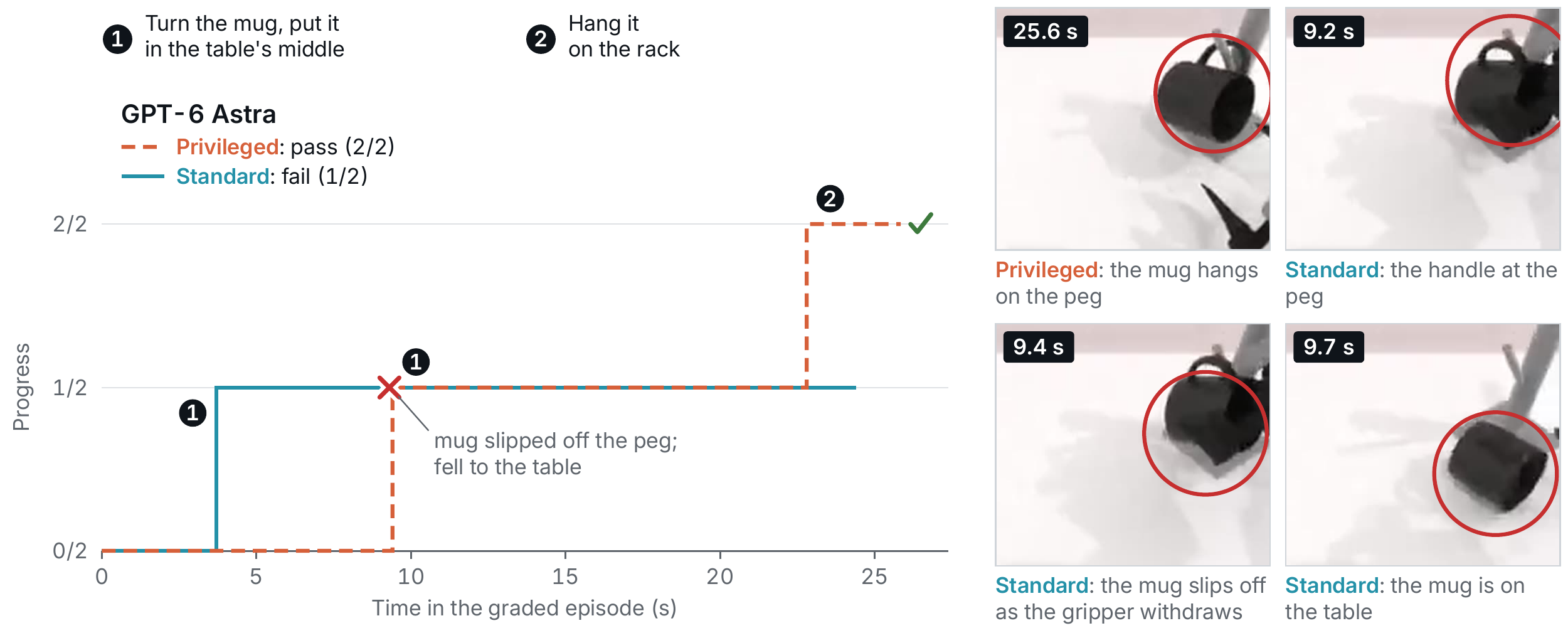}
    \caption{}
    \label{fig:fa-case-mug}
  \end{subfigure}

  \vspace{0.8em}
  \begin{subfigure}[t]{\linewidth}
    \includegraphics[width=\linewidth]{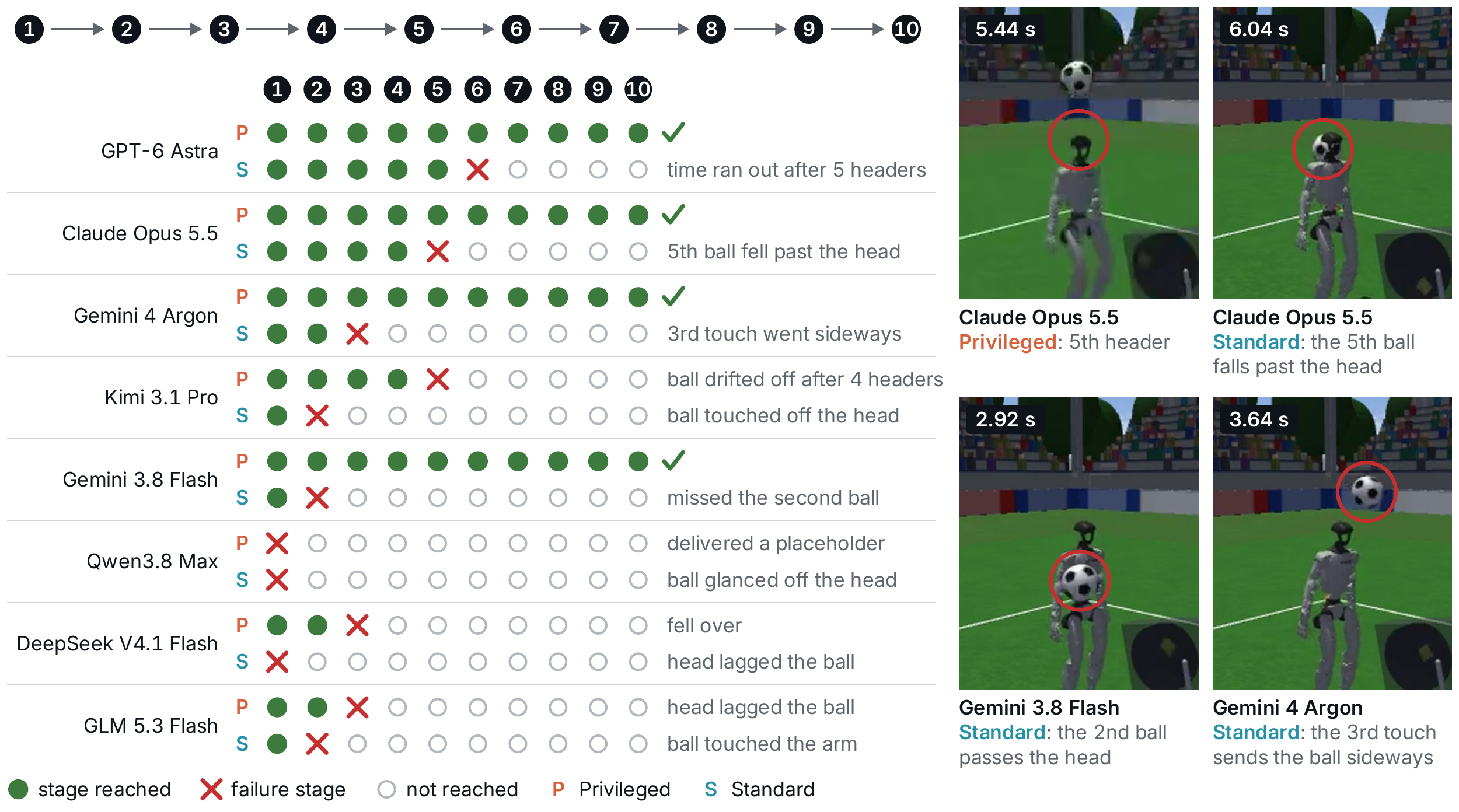}
    \caption{}
    \label{fig:fa-case-juggle}
  \end{subfigure}
  \caption{\evalsetting{Privileged} and \evalsetting{Standard} settings. (a) RoboTwin 2.0, \emph{hanging a mug}: progress maps of GPT-6 Astra. (b) Embodied Olympics - Fun, \emph{heading a ball ten times}: the headers of each agent in the \evalsetting{Privileged} (P) and \evalsetting{Standard} (S) settings.}
  \label{fig:fa-case-f7}
\end{figure}

\paragraph{Disposing of batteries (\Cref{sec:f-difficulty}, \Cref{fig:fa-case-batteries}).} In this BEHAVIOR-1K task a mobile two-armed robot must put three batteries into the wastebasket of a copy room, one from a cabinet in the copy room and two from a desk in an office along the corridor, and leave the wastebasket on the floor. The task graph has four stages, one for each battery and one for the wastebasket on the floor, which holds from the start. In the \evalsetting{Standard} Setting, the three strongest agents each put the cabinet battery into the wastebasket and then failed at the second battery, when the robot tipped over. GPT-6 Astra dropped the first battery into the wastebasket at 90 s. At 359 s, as it drove into the office, the base caught the door jamb and the robot tipped onto its side. It righted the robot by 542 s and picked up a second battery at 592 s, and it still held the battery when the episode ended. Claude Opus 5.5 dropped the first battery in at 102 s, and the robot tipped over at 486 s as it turned at the office door. Gemini 4 Argon dropped the first battery in at 23 s and tipped the robot over at 90 s while turning in the office. In both of these runs the simulator service ended the episode because the robot had fallen. Four of the five weaker agents failed at the first battery. Gemini 3.8 Flash never lifted it from the cabinet, DeepSeek V4.1 Flash tipped the robot over while searching for it, and GLM 5.3 Flash lost the robot's balance while searching. Qwen3.8 Max's robot also fell. Kimi 3.1 Pro put the cabinet battery into the wastebasket and then searched the office without finding another battery before the episode ended.

\paragraph{Hanging a mug (\Cref{sec:f-settings}, \Cref{fig:fa-case-mug}).} In this RoboTwin 2.0 task a two-armed robot must pick up a mug with the left arm, turn it and put it down in the middle of the table, and then hang it on a rack with the right arm. In the \evalsetting{Privileged} Setting, GPT-6 Astra put the mug down at 9 s and hung it at 23 s. In the \evalsetting{Standard} Setting it put the mug down at 4 s and carried it to the rack with the right arm. At 9 s, having reported that the peg was through the handle, it opened the gripper, and the mug slipped off as the gripper withdrew and fell to the table. Two more tries also failed, and the episode ended with the mug not hung. Claude Opus 5.5 passed both the \evalsetting{Privileged} and the \evalsetting{Standard} test.

\paragraph{Heading a ball ten times (\Cref{sec:f-settings}, \Cref{fig:fa-case-juggle}).} In this Embodied Olympics - Fun task a humanoid robot must head a ball from a ball machine ten times in a row, and each header is a stage. In the \evalsetting{Privileged} Setting, GPT-6 Astra, Claude Opus 5.5, Gemini 4 Argon and Gemini 3.8 Flash all made the ten headers. In the \evalsetting{Standard} Setting the agent sees the ball only in camera images, the task grants 5 resets, and only the last episode counts. There, the graded episodes of the same four agents reached five, four, two and one headers. Claude Opus 5.5 first watched the lob twice while standing still. It then tracked the ball in the venue camera, fitted its flight, and timed a crouch and a push up of the legs to the ball's arrival. From its third episode on, this approach made the first header every time, and it spent the remaining resets aiming the headers with the waist and ankles. In its graded episode the fifth ball fell past the head and hit the ground at 6.2 s. GPT-6 Astra made its first header in its fourth episode, and its graded episode had reached five headers when the agent's time ran out. Gemini 4 Argon's third touch, at 3.6 s, sent the ball sideways. Gemini 3.8 Flash made one header, and in its graded episode the second ball passed the head and hit the ground at 3.1 s.

%% file: appendix_sections/j_real/j_real.tex
\begin{figure}[tb]
  \centering
  \begin{subfigure}[b]{0.68\linewidth}
    \centering
    \includegraphics[width=\linewidth]{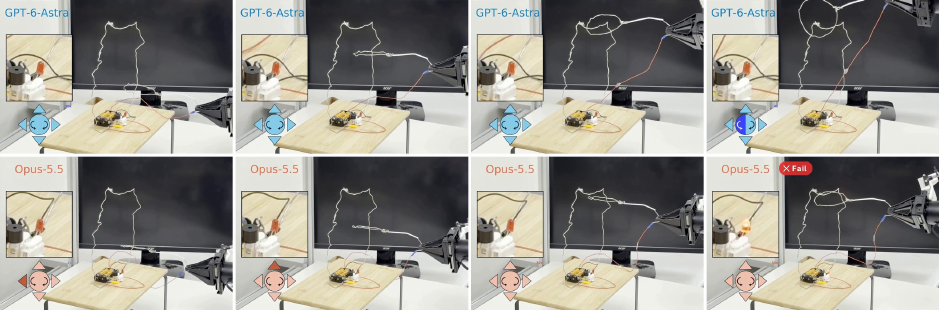}
    \caption{}
    \label{fig:real-frames}
  \end{subfigure}\hfill
  \begin{subfigure}[b]{0.30\linewidth}
    \centering
    \includegraphics[width=\linewidth]{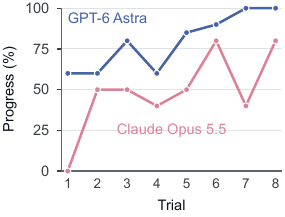}
    \caption{}
    \label{fig:real-progress}
  \end{subfigure}
  \caption{Buzz wire game on a physical robot arm. (a) Four frames in time order of one trial of GPT-6 Astra and one of Claude Opus 5.5, with a close-up of the light and the buzzer and the command being executed. In the last frame of Claude Opus 5.5's trial, the loop has touched the wire and the light is on. (b) The plot gives the progress score of each agent in each of its eight trials.}
  \label{fig:real}
\end{figure}

Because the \evalsetting{Standard} Setting gives the agent only the robot's sensor readings and its control interface (\Cref{sec:settings}), an interface of the same kind can connect a coding agent to a physical robot. As a first test, we connected two coding agents to a physical robot arm, and each agent played a buzz wire game for eight trials in a row (\Cref{fig:real}). This demonstration is not part of the exam, and its results are not included in any success rate in this paper.

\paragraph{Task.} In a buzz wire game, a player guides a metal loop along a bent wire from one end to the other without touching the wire. The loop and the wire belong to one electric circuit, so a contact closes the circuit and turns on a light and a buzzer. In our setup, both ends of the bent wire are fixed to a wooden board, and the robot holds the handle of the loop in its gripper (\Cref{fig:real-frames}). A contact counts as a failure, as in the buzz wire tasks of RoboARC (\Cref{app:tasks-new}), in which a simulated Franka Panda arm guides a metal loop along a wire.

\paragraph{Commands and video.} The agent controls the robot with discrete commands. Each command moves the gripper up, down, left or right or rotates it in either direction. In each panel of \Cref{fig:real-frames}, the pad at the lower left highlights the command being executed, and the inset shows a close-up of the light and the buzzer, so a contact is visible in the video.

\paragraph{Agents and trials.} The two agents use GPT-6 Astra and Claude Opus 5.5, the two models with the highest \evalsetting{Standard} success rates in \Cref{fig:leaderboard}. Each agent ran one sequence of eight trials on the same wire.

\paragraph{Progress score.} Each trial receives a progress score, the percentage of the wire that the loop covers before the trial ends. A score of 100\% means that the loop passed along the whole wire without a contact.

\paragraph{Results.} \textbf{Both agents score higher in the last four trials than in the first four, and GPT-6 Astra scores higher than Claude Opus 5.5 in every trial} (\Cref{fig:real-progress}). Averaged over the first four trials, the progress is 65.0\% for GPT-6 Astra and 35.0\% for Claude Opus 5.5. Averaged over the last four trials, the progress is 93.8\% and 62.5\%. GPT-6 Astra scores at least 80\% in five of its eight trials, and Claude Opus 5.5 does so in two. GPT-6 Astra starts at 60\% and passes along the whole wire in trials 7 and 8. Its progress falls once, from 80\% in trial 3 to 60\% in trial 4. Claude Opus 5.5 scores 0\% in trial 1 and between 40\% and 80\% in trials 2 to 8, and it does not pass along the whole wire in any trial. Its progress varies more from trial to trial. It rises to 80\% in trial 6, falls to 40\% in trial 7 and returns to 80\% in trial 8. In the trial shown in \Cref{fig:real-frames}, Claude Opus 5.5 touches the wire, and the light in its close-up turns on. The two agents rank in the same order as in the \evalsetting{Standard} Setting in simulation, where GPT-6 Astra solves 33.6\% of the \evalsetting{Standard} tests and Claude Opus 5.5 solves 25.0\% (\Cref{fig:leaderboard}).

\paragraph{Scope.} The demonstration covers one task, one robot and two agents, with one sequence of eight trials per agent. It shows that a coding agent can control a physical robot through an interface of this kind, and that the progress of both agents rose over the eight trials. With one sequence per agent, the demonstration cannot separate this rise from variation between runs. It also does not measure how well success rates in simulation predict results on hardware (\Cref{app:limitations}). A systematic evaluation on physical robots, with a grading program and several runs per agent, remains future work.

\FloatBarrier

%% file: appendix_sections/i_limitations/i_limitations.tex
Each result holds for one model in one harness. Each agent also makes a single attempt per test, so the results cannot separate an agent's ability from the variation between runs. The core set excludes tasks that we judged easy and tasks that GPT-6 Luna solved in a screening run, so its success rates are lower than those of the full exam by an amount that differs between suites. Because GPT-6 Luna belongs to the same family as GPT-6 Astra, the screening may also bias the core set against tasks that this family solves. Many per-suite rates are based on few tests. In a suite with five or six tests, a single test changes the rate by 17 to 20 percentage points, and a single \evalsetting{Realtime} test changes an agent's success rate over all \evalsetting{Realtime} tests by about two points. The stage-level and control-approach labels come from agents. The model that wrote the stage-level labels, GPT-6.1 Sol, belongs to the same family as GPT-6 Astra. This shared family could bias the label-based comparisons between agents. Of the 1,156 tasks, 392 have no reference solution, so we have not shown that each of them can be solved in every setting. Time limits differ between suites, the tasks of Embodied Olympics - Fun run with open network access, and results on tasks adapted from public benchmarks may partly reflect memorization of published solutions (\Cref{app:protocol}). Finally, all results outside \Cref{app:real} are in simulation. The demonstration in \Cref{app:real} covers one task on a physical robot, so how well the simulation results predict results on hardware remains open.

%% file: appendix_sections/k_related/k_related.tex
This appendix extends the review in \Cref{sec:related}. We first review benchmarks for coding agents and benchmarks for robot learning, and then work that uses language models for robots. We then review recent work that runs general-purpose agents on robots, together with the harnesses and benchmarks built for these agents.

\paragraph{Benchmarks for coding agents.} \looseness=-1 Language models and the agents built on them are evaluated mostly on digital tasks. These tasks include writing programs from a problem description~\citep{chen2021evaluating,jain2024livecodebench}, resolving issues in real repositories~\citep{jimenez2023swebench,yang2024swebenchmm,miserendino2025swelancer}, completing long software tasks and tasks in a terminal~\citep{kwa2025measuring,merrill2026terminalbench}, engineering algorithms for hard optimization problems~\citep{imajuku2025alebench}, machine learning engineering and research~\citep{chan2024mlebench,wijk2024rebench,starace2025paperbench}, writing efficient GPU kernels~\citep{ouyang2025kernelbench}, cybersecurity~\citep{zhang2024cybench}, and operating websites, desktop applications and other tools~\citep{zhou2023webarena,xie2024osworld,liu2023agentbench,mialon2023gaia,yao2024taubench}. Most of these agents interleave reasoning with tool calls~\citep{yao2023react} and revise their work from feedback~\citep{shinn2023reflexion}. SWE-agent~\citep{yang2024sweagent} shows that the interface between the model and the computer affects how often an agent succeeds, and OpenHands~\citep{wang2024openhands} provides an open platform for building and evaluating such agents. Other work measures how often agents exploit the tests that grade them~\citep{zhong2025impossiblebench} and proposes practices for building rigorous agentic benchmarks~\citep{zhu2025establishing}. The \efe{} keeps each coding agent's harness unchanged and extends these evaluations to robotics tasks. In its \evalsetting{Standard} and \evalsetting{Realtime} settings, the agent must interpret sensor observations and often cannot undo an action, and in the \evalsetting{Realtime} Setting the environment keeps running while the agent reasons. In every setting, a program that the agent cannot change grades each attempt.

\paragraph{Benchmarks for robot learning.} Most robotics benchmarks focus on visuomotor policy learning and evaluate learned policies, from imitation-learning policies~\citep{chi2023diffusion,zhao2023aloha} to generalist policies and vision-language-action models~\citep{brohan2022rt1,brohan2023rt2,team2024octo,kim2024openvla,kim2025oft,black20240,pertsch2025fast,intelligence2025pi05,intelligence2025pistar06,nvidia2025gr00t,team2025gemini} trained on large robot datasets~\citep{oxe2024,khazatsky2024droid}. Recent world action models build policies on pretrained video generation models such as Cosmos~\citep{nvidia2025cosmos} and predict future observations together with actions~\citep{kim2026cosmospolicy,ye2026world,lu2026wamsurvey}, extending earlier work that extracts actions from generated videos~\citep{du2023unipi} or learns a joint representation of video and actions~\citep{li2025uva}. These benchmarks are built on physics simulators, control suites and robot learning frameworks~\citep{todorov2012mujoco,tassa2018deepmind,makoviychuk2021isaacgym,mittal2023orbit,nvidia2025isaaclab,zhu2020robosuite,gu2023maniskill2,tao2024maniskill3,szot2021habitat,puig2023habitat3,geng2025roboverse}. They cover tabletop and bimanual manipulation~\citep{james2020rlbench,yu2020metaworld,mees2022calvin,liu2023libero,jiang2023vima,pumacay2024colosseum,yang2026robolab,mclean2025metaworld,zhang2024vlabench,chen2025robotwin,chen2026robodojo,zhang2026liberomax,yamlab2026,nvidia2025gr00t}, household activities~\citep{shridhar2020alfred,srivastava2022behavior,li2024behavior1k,nasiriany2024robocasa,nasiriany2026robocasa365,kim2026molmospaces}, humanoid and legged locomotion and whole-body control~\citep{sferrazza2024humanoidbench,alhafez2023locomujoco,zhang2026bigym,zakka2025mujoco}, dexterous and tool manipulation~\citep{chen2022bidexhands,kedia2026simtoolreal,wang2026dexjoco}, musculoskeletal control~\citep{caggiano2022myosuite}, physical reasoning and creative problem solving~\citep{huang2026kinder,lin2026robowits}, aerial manipulation~\citep{wang2026ambench}, laboratory automation~\citep{lan2025autobio,li2025labutopia} and safety~\citep{peng2026maniguard}. Policies trained on real data are also evaluated in simulated counterparts of real setups~\citep{li2024simplerenv}, on real robots at scale~\citep{zhou2025autoeval,atreya2025roboarena} and in learned world models~\citep{quevedo2025worldgym,guo2025ctrlworld,gemini2025veo}, which can also generate training data for policies~\citep{yang2023unisim,jang2025dreamgen}. Most of these benchmarks do not need to restrict what the evaluated system can access, because a learned policy receives only a fixed set of observations. A coding agent can run any program on its computer and could read the simulator's files, so the exam adds rules about what the agent may access (\Cref{sec:settings}). The exam adapts 24 of these benchmarks and two others, EmbodiedSWE~\citep{shen2026embodiedswe} and ABC Sim~\citep{abcsim2026}, and keeps their scenes and success rules, so that coding agents can attempt their tasks under the exam's rules (\Cref{app:tasks-adapted}). Its \evalsetting{Standard} Setting gives the agent observations of the kind that a learned policy receives, so coding agents and learned policies can be compared with the same observations in future work.

\paragraph{Language models for robot planning and control.} \looseness=-1 Language models have been used to plan tasks for robots, write robot programs and design training signals. SayCan~\citep{ahn2022saycan}, Inner Monologue~\citep{huang2022inner}, ProgPrompt~\citep{singh2023progprompt}, Text2Motion~\citep{lin2023text2motion} and SayPlan~\citep{rana2023sayplan} sequence learned or scripted skills into plans, PaLM-E~\citep{driess2023palme} and Gemini Robotics-ER~\citep{team2025gemini,team2025gemini15} train multimodal models for embodied reasoning, and embodied chain-of-thought reasoning~\citep{zawalski2024ecot} trains a vision-language-action model to reason about plans, sub-tasks, motions and object locations before it acts. AutoRT~\citep{ahn2024autort} uses foundation models to propose and assign tasks to a fleet of robots, and SOAR~\citep{zhou2024soar} uses them to propose tasks and practice them autonomously. Code as Policies~\citep{liang2022code}, ChatGPT for Robotics~\citep{vemprala2023chatgpt}, Instruct2Act~\citep{huang2023instruct2act} and RoboTool~\citep{xu2023creative} turn instructions into programs that call perception and control primitives, and Code-as-Monitor~\citep{zhou2024codeasmonitor} applies the same idea to failure detection, with a model that writes code to check spatio-temporal constraints during execution. VoxPoser~\citep{huang2023voxposer}, ReKep~\citep{huang2024rekep} and MOKA~\citep{liu2024moka} have a model specify value maps, keypoint constraints or keypoints selected from candidates marked on the image, which a planner or an optimizer then turns into motions. Language to Rewards~\citep{yu2023language}, Eureka~\citep{ma2024eureka}, Text2Reward~\citep{xie2024text2reward}, DrEureka~\citep{ma2024dreureka} and RDA~\citep{lee2026rda} have a model write and refine reward functions, which a controller or a reinforcement learning algorithm then optimizes. RoboGen~\citep{wang2024robogen} and GenSim~\citep{wang2024gensim} generate simulated tasks and training data, and Voyager~\citep{wang2023voyager} builds a library of code skills in an open-ended game. Most of these systems pair a model with an interface designed for it, such as a set of primitives, a planner or a training pipeline. Primitives and planners contribute the kind of classical engineering that \citet{goldberg2025gofe} argues robotics still needs, because robot data are far scarcer than the internet data used to train vision-language models. The \efe{} builds on this line of work and evaluates unchanged general-purpose coding agents on the whole loop of reading an interface, choosing a control approach, carrying it out and revising it.

\paragraph{Embodied benchmarks for language models.} \looseness=-1 ALFWorld~\citep{shridhar2021alfworld} pairs text-based household tasks with their embodied versions and has become a common testbed for language-model agents~\citep{yao2023react,shinn2023reflexion}. Embodied Agent Interface~\citep{li2024embodied}, EmbodiedBench~\citep{yang2025embodiedbench} and VisualAgentBench~\citep{liu2024visualagentbench} evaluate language models as decision makers in simulated households and other embodied environments, and MineDojo~\citep{fan2022minedojo} provides open-ended Minecraft tasks for embodied agents. Embodied Agent Interface divides embodied decision making into goal interpretation, subgoal decomposition, action sequencing and transition modeling, and EmbodiedBench evaluates both high-level tasks and low-level navigation and manipulation. In these benchmarks the model chooses from an action space that the benchmark defines, which ranges from high-level skills such as picking up an object to discretized end-effector poses. The exam gives the agent the robot's control interface, so the agent also decides how its actions are produced.

\paragraph{Agentic robots.} A growing body of recent work runs general-purpose agents on robots, many with GPT-6 Astra~\citep{openai2026gpt6astra} or Claude Fable~\citep{anthropic2026fable5} as the model. Recent surveys organize this work by model, data, environment and harness~\citep{qin2026agentic} and compare agents that use robots with agents that use computers~\citep{chen2026mmea}. This work differs in two respects: the control approach that the agent uses, and the harness that connects the model to the robot. This paragraph groups the work by the three control approaches of \Cref{fig:approaches}, and the next paragraph reviews harnesses.

\emph{Direct control.} In this approach, the agent observes the scene and outputs each action itself. Several studies use GPT-6 Astra as the policy in this way. They evaluate it on RoboDojo, RoboTwin~2.0 and other benchmarks~\citep{zhang2026unexpected,zhou2026gpt6,sensenova2026astra}, on memory-dependent manipulation tasks~\citep{jedlicki2026memory}, on contact-rich tasks with tactile feedback~\citep{wang2026touch}, and with demonstrations and its own interaction history in context~\citep{liu2026roboicl,cheng2026incontext}.

\emph{Code to control.} In this approach, the agent writes programs that produce the robot's actions. Several studies give a general-purpose coding agent the robot's programming interface and add few tools for the task. FAEA~\citep{tsui2026demonstration} runs the unmodified Claude Agent SDK on simulated manipulation benchmarks with access to the simulator state, and Agent as Policy~\citep{jia2026agent} has a general-purpose agent interpret camera images, write programs and send motion commands on real robots. Claude Code (Opus~5) and Codex (GPT-5.6 Sol and GPT-6 Astra) write task and motion planning programs that generalize to unseen problem instances~\citep{merler2026coding}. With Claude Fable, a coding agent writes a controller that solves Push-T without demonstrations~\citep{xie2026pusht} and writes force-feedback programs for contact-rich tasks~\citep{tan2026forca}. In Anthropic's Project Fetch, human teams programmed a quadruped robot with and without Claude~\citep{anthropic2025fetch}, and Claude Opus~4.7, working without human help, was later about 20 times faster than the fastest human team at all the tasks that the teams had completed~\citep{ilie2026fetch2}. Other systems add procedures that help the agent write better programs. They search over multi-file policy repositories that compose perception, planning and control primitives~\citep{elmaaroufi2026rho}, build libraries of programs from execution feedback~\citep{lu2026aspire,kuroki2026learning} or of verified skills from a few demonstrations~\citep{nie2026teach}, reuse body descriptions, past experience and skills derived from the agent's own routines~\citep{he2026astra}, write a shared code library that makes every decision without a model at test time~\citep{hu2026etm}, infer a program from a single human demonstration~\citep{liu2026rapid}, let the agent check and retry primitives within Python cells~\citep{si2026fewer}, let the agent develop its tools in simulation before it controls a real robot~\citep{hu2026simex}, or have one agent revise, on a real robot, the tools that another agent calls~\citep{gao2026selfevolving}. RAX~\citep{huang2026rax} has a coding agent explore online without hand-designed primitives, given at most a few goal images or demonstrations, and turn the successful action sequences into programs for single-arm, bimanual, dexterous, mobile and humanoid systems, both in simulation and on real robots.

\emph{Train a policy.} In this approach, the agent trains a policy and then uses it to control the robot. ENPIRE~\citep{xiao2026enpire} gives coding agents a real-world loop of resetting the scene, running a policy, verifying the outcome and refining the policy. With this loop, frontier coding agents train policies that reach a 99\% success rate on dexterous tasks such as organizing a pin box. HARBOR~\citep{li2026harbor} has a team of agents set up simulated environments, design rewards and tune reinforcement learning. The reward design methods above automate one step of this approach.

\looseness=-1 Most of these studies fix the control approach in advance, and Agent as Policy and ENPIRE let the agent choose between two of the approaches. The exam supports all three approaches and leaves the choice to the agent, so it also measures which approach agents choose (\Cref{sec:f-control}).

\paragraph{Harnesses for agentic robots.} The harness is the program between the model and the robot. Harnesses for robots differ in the action interface they provide, in whether the agent can call learned policies, and in the support they add for tracking state and checking outcomes.

\emph{Action interfaces.} \looseness=-1 The success rates of coding agents depend on the level of abstraction of the interface~\citep{fu2026capx}. Show-Harness~\citep{chen2026showharness} gives vision-language models a small set of semantic action units, which robot-specific interpreters convert into robot actions, and VIA~\citep{hu2026via} gives off-the-shelf agents a visual interface with tools to probe pixels and set end-effector targets. HexaShell~\citep{gao2026hexashell} shares 14 general tools for motion, perception and state across all tasks and robots, because code refined for one benchmark tends to overfit it. OpenRUA~\citep{chu2026openrua} gives an unmodified coding agent only terminal access to the robot's ROS~2 interface, its documentation and basic tools. The uni-robot-agent harness~\citep{yuan2026uniroboagent} supports further training of open-weight models.

\emph{Learned policies as tools.} Hi Robot~\citep{shi2025hirobot} trains a high-level vision-language model that sends language commands to a vision-language-action policy. RoboClaw~\citep{li2026roboclaw} and Harness VLA~\citep{zhang2026harnessvla} let the agent call vision-language-action policies as tools, Recursive Harness Distillation~\citep{kim2026rhd} distills a strong agent's interventions in a vision-language-action policy into a playbook for a lighter agent, and Gemini Robotics-ER~1.5 orchestrates the Gemini Robotics~1.5 vision-language-action model~\citep{team2025gemini15}. The Physical Agency orchestrator~\citep{galanti2026physical} plans subgoals, calls vision-language-action policies and parametrized skills, checks the outcomes and recovers from failures. \citet{yao2026ebench} evaluate GPT-6 Astra on EBench both as the policy and as an agent that can also call a vision-language-action or world action model policy, and compare both with the policies alone.

\emph{State tracking and verification.} Other harnesses add a persistent representation of the scene and a check of each action's outcome~\citep{wang2026thea}, calibrated 3D measurements and a preview of each motion~\citep{wang2026auriga}, sub-agents for perception, verification and memory~\citep{fang2026emerge}, or a vision-language model that monitors a policy and corrects it after a confirmed failure~\citep{li2026spotter}.

The \efe{} runs each harness unmodified with the models of its own provider, and through an adapter with other providers' models (\Cref{app:ablation}), so different harnesses can be compared on the same tasks.

\paragraph{Benchmarks for agentic robots.} Several benchmarks evaluate agents that control robots. CaP-X~\citep{fu2026capx} evaluates coding agents that compose perception and control primitives at several levels of abstraction and finds that their success rates depend on the abstractions that designers provide. RobotWorld~\citep{yang2026robotworld} evaluates agents on 84 tasks in manipulation, mobile manipulation, locomotion, driving and aerial control with executable success checks. EmbodiedSWE~\citep{shen2026embodiedswe} evaluates coding agents on long-horizon dexterous tasks and turns successful agent solutions into demonstrations for training vision-language-action policies. RoboQuest~\citep{liu2026roboquest} evaluates whether agents explore to find the information a task requires, BiGym~2.0~\citep{zhang2026bigym} compares learned policies with programs that coding agents develop on humanoid household tasks, and \citet{xu2026safeharness} evaluate whether coding agents avoid an obstacle they are told not to touch and propose obstacle-aware harnesses for route planning and contact. Earlier safety studies jailbreak robots that language models control into harmful actions~\citep{robey2024jailbreaking}, test whether models recognize undesirable situations in images and text~\citep{sermanet2025asimov}, and test whether agents that plan with high-level actions reject hazardous household tasks~\citep{yin2024safeagentbench}. In the exam's Embodied Safety Exam, the agent controls the robot itself, and the verifier decides success from what physically happens in a physics simulation. Of the agentic-robot studies above, many report results on RoboDojo, several on RoboTwin~2.0 and one on KinDER~\citep{merler2026coding}. The exam adapts all three suites. The \efe{} covers 47 robots of nine types, which include humanoids, quadrupeds, dexterous hands, aerial robots and a wheeled car. It also separates what an agent can do with full access to the simulator from what it can do through the robot's sensors alone.